\documentclass[10pt,journal,compsoc]{IEEEtran}

\usepackage{amsmath}
\usepackage{color}
\usepackage{colortbl}
\usepackage[pagebackref=true,breaklinks=true,colorlinks,bookmarks=false]{hyperref}
\usepackage{amssymb}
\usepackage{longtable}
\usepackage{xcolor}
\usepackage{siunitx}
\usepackage{multirow}

\ifCLASSOPTIONcompsoc
  \usepackage[nocompress]{cite}
\else
  \usepackage{cite}
\fi

\ifCLASSINFOpdf
  \usepackage[pdftex]{graphicx}
\else
\fi

\ifCLASSOPTIONcompsoc
 \usepackage[caption=false,font=footnotesize,labelfont=sf,textfont=sf]{subfig}
\else
 \usepackage[caption=false,font=footnotesize]{subfig}
\fi

\usepackage{romannum}

\begin{document}
\pagenumbering{arabic}
\title{Deep High-Resolution Representation Learning for Visual Recognition}

\author{
Jingdong Wang,
Ke Sun,
Tianheng Cheng,
Borui Jiang,
Chaorui Deng,
Yang Zhao,
Dong Liu,
Yadong Mu,
Mingkui Tan,
Xinggang Wang,
Wenyu Liu,
and Bin Xiao
\IEEEcompsocitemizethanks{\IEEEcompsocthanksitem J. Wang is with Microsoft Research,
Beijing, P.R. China.\protect\\
E-mail: jingdw@microsoft.com
}%
}

\markboth{IEEE Transactions on Pattern Analysis and Machine Intelligence, March~2020}%
{Wang \MakeLowercase{\textit{et al.}}: Deep High-Resolution Representation Learning for Visual Recognition}

\IEEEtitleabstractindextext{%
\begin{abstract}
High-resolution representations are essential for position-sensitive
vision problems, such as human pose estimation, semantic segmentation,
and object detection.
Existing state-of-the-art frameworks 
first encode the input image
as a low-resolution representation
through a subnetwork that is formed by connecting high-to-low resolution convolutions \emph{in series} (e.g., ResNet, VGGNet),
and then
recover the high-resolution representation
from the encoded low-resolution representation.
Instead, our proposed network, named as High-Resolution Network (HRNet),
maintains high-resolution representations through the whole process. 
There are two key characteristics: 
(i) Connect the high-to-low resolution convolution streams \emph{in parallel};
(ii) Repeatedly exchange the information across resolutions.
The benefit is that the resulting representation is semantically richer
and spatially more precise. 
We show the superiority of the proposed HRNet
in a wide range of applications, 
including human pose estimation, 
semantic segmentation, and object detection, 
suggesting that the HRNet is a stronger backbone for computer vision problems. 
All the codes are available at~{\url{https://github.com/HRNet}}.
\end{abstract}

}

\maketitle

\IEEEdisplaynontitleabstractindextext

\IEEEpeerreviewmaketitle

\ifCLASSOPTIONcompsoc
\IEEEraisesectionheading{\section{Introduction}\label{sec:introduction}}
\else
\section{Introduction}
\label{sec:introduction}
\fi

\IEEEPARstart{D}{eep} convolutional neural networks (DCNNs)
have achieved state-of-the-art results in many computer vision tasks,
such as image classification,
object detection,
semantic segmentation,
human pose estimation, and so on.
The strength
is that DCNNs are able to learn richer representations
than conventional hand-crafted representations.

\begin{figure*}[t]
    \centering
    \includegraphics[width = 0.99\textwidth]{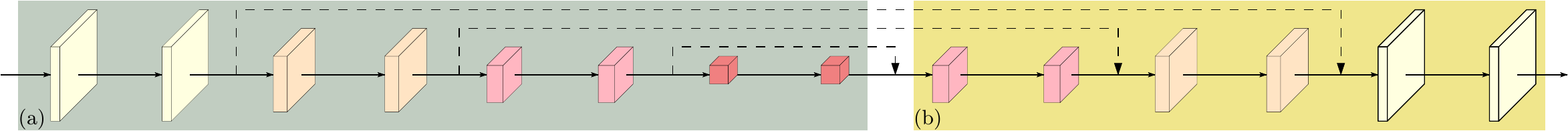}
    \vspace{-.2cm}
    \caption{The structure of recovering high resolution from low resolution.
    (a) A low-resolution representation learning subnetwork (such as VGGNet~\cite{SimonyanZ14a}, ResNet~\cite{HeZRS16}),
    which is formed by connecting high-to-low convolutions in series.
    (b) A high-resolution representation recovering subnetwork,
    which is formed by connecting low-to-high convolutions in series.
    Representative examples include SegNet~\cite{BadrinarayananK17}, DeconvNet~\cite{NohHH15}, U-Net~\cite{RonebergerFB15} and Hourglass~\cite{NewellYD16}, encoder-decoder~\cite{PengFWM16}, and SimpleBaseline~\cite{XiaoWW18}.}
    \label{fig:ClassificationNetwork}
    \vspace{-3mm}
\end{figure*}

Most recently-developed classification networks,
including AlexNet~\cite{KrizhevskySH12}, VGGNet~\cite{SimonyanZ14a}, GoogleNet~\cite{SzegedyLJSRAEVR15}, ResNet~\cite{HeZRS16}, 
etc.,
follow the design rule of LeNet-$5$~\cite{Lecun98}.
The rule is depicted in Figure~\ref{fig:ClassificationNetwork} (a):
gradually reduce the spatial size of the feature maps,
connect the convolutions from high resolution
to low resolution in series,
and lead to a~\emph{low-resolution representation},
which is further processed for classification.

\emph{High-resolution representations}
are needed for position-sensitive tasks,
e.g., semantic segmentation, human pose estimation,
and object detection.
The previous state-of-the-art methods 
adopt the high-resolution recovery process
to raise the representation resolution from 
the low-resolution representation outputted by a classification or classification-like network
as depicted in Figure~\ref{fig:ClassificationNetwork} (b),
e.g., Hourglass~\cite{NewellYD16},
SegNet~\cite{BadrinarayananK17}, 
DeconvNet~\cite{NohHH15},
U-Net~\cite{RonebergerFB15},
SimpleBaseline~\cite{XiaoWW18},
and encoder-decoder~\cite{PengFWM16}.
In addition, dilated convolutions
are used to remove some down-sample layers
and thus yield medium-resolution representations~\cite{ChenPKMY18,ZhaoSQWJ17}.

\begin{figure*}[t]
\footnotesize
    \centering
    \includegraphics[width = 0.99\textwidth]{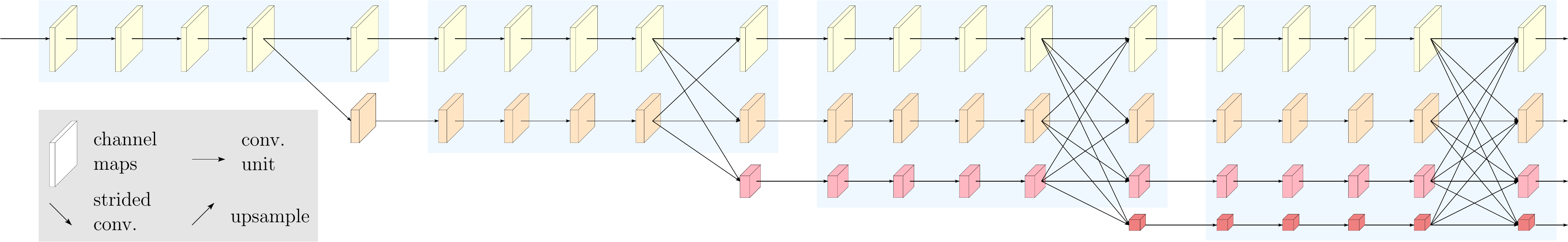}
    \vspace{-.2cm}
    \caption{An example 
    of a high-resolution network.
    {Only the main body is illustrated, and 
    the stem (two stride-$2$ $3\times3$ convolutions) is not included.}
    There are four stages. 
    The $1$st stage consists of high-resolution convolutions.
    The $2$nd ($3$rd, $4$th) stage
    repeats two-resolution
    (three-resolution, four-resolution)
    blocks. The detail is given in Section~\ref{sec:HRNet}.}
    \label{fig:HRNet}
    \vspace{-3mm}
\end{figure*}

We present a novel architecture,
namely High-Resolution Net (HRNet),
which is able to \emph{maintain high-resolution representations} 
through the whole process.
We start from a high-resolution convolution stream,
gradually add high-to-low resolution convolution streams one by one,
and connect the multi-resolution streams in parallel.
The resulting network consists of several ($4$ in this paper) stages as 
depicted in Figure~\ref{fig:HRNet},
and
the $n$th stage contains $n$ streams
corresponding to $n$ resolutions.
We conduct repeated multi-resolution fusions
by exchanging the information
across the parallel streams
over and over.

The high-resolution representations 
learned from HRNet are not only semantically strong 
but also spatially precise.
This comes from two aspects.
(\romannum{1})
Our approach connects 
high-to-low resolution convolution streams
in parallel 
rather than in series.
Thus, our approach is able to {maintain} the high resolution
instead of recovering high resolution from low resolution,
and accordingly the learned representation is potentially spatially more precise.
(\romannum{2})
Most existing fusion schemes
aggregate high-resolution low-level 
and high-level representations
obtained by upsampling low-resolution representations.
Instead, we repeat multi-resolution fusions
to boost the high-resolution representations 
with the help of the low-resolution representations,
and vice versa.
As a result, all the high-to-low resolution representations
are semantically strong. 

We present two versions of HRNet.
The first one, named as HRNetV$1$,
only outputs the high-resolution representation
computed from the high-resolution convolution stream.
We apply it to human pose estimation
by following the heatmap estimation framework.
We empirically demonstrate the superior 
pose estimation performance 
on the COCO keypoint detection dataset~\cite{LinMBHPRDZ14}.

The other one, named as HRNetV$2$,
combines the representations 
from all the high-to-low resolution parallel streams.
We apply it
to semantic segmentation
through estimating segmentation maps
from the combined high-resolution representation.
The proposed approach achieves
state-of-the-art results
on PASCAL-Context, Cityscapes, and LIP
with similar model sizes and lower computation complexity.
We observe similar performance for HRNetV$1$
and HRNetV$2$ over COCO pose estimation,
and the superiority of HRNetV$2$ to HRNet$1$
in semantic segmentation.

In addition, we construct a multi-level representation,
named as HRNetV$2$p,
from the high-resolution representation output from HRNetV$2$,
and apply it to 
state-of-the-art detection frameworks,
including Faster R-CNN, Cascade R-CNN \cite{CaiV18},
FCOS~\cite{TianSCH19}, and CenterNet~\cite{DuanBXQHT19},
and state-of-the-art joint detection and instance segmentation frameworks,
including Mask R-CNN \cite{HeGDG17}, 
Cascade Mask R-CNN,
and Hybrid Task Cascade~\cite{ChenPWXLSFLSOLL19}.
The results show that our method gets detection performance improvement and in particular dramatic improvement
for small objects.

\section{Related Work}
We review closely-related representation learning techniques developed 
mainly for human pose estimation~\cite{HuR16},
semantic segmentation
and object detection,
from three aspects:
low-resolution representation learning,
high-resolution representation recovering, 
and high-resolution representation maintaining. 
Besides,
we mention about some works
related to multi-scale fusion.

\noindent\textbf{Learning low-resolution representations.}
The fully-convolutional network approaches~\cite{LongSD15,SermanetEZMFL13}
compute low-resolution representations
by removing the fully-connected layers in a classification network,
and estimate their coarse segmentation maps.
The estimated segmentation maps are improved
by combining the fine segmentation score maps
estimated from intermediate low-level medium-resolution representations~\cite{LongSD15},
or iterating the processes~\cite{KowalskiNT17}.
Similar techniques have also been applied 
to edge detection, e.g., holistic edge detection~\cite{XieT15}.

The fully convolutional network
is extended,
by replacing a few (typically two) strided convolutions 
and the associated convolutions with dilated convolutions,
to the dilation version,
leading to medium-resolution representations~\cite{ZhaoSQWJ17,ChenPKMY18, YuKF17,ChenPKMY14, LiPYZDS18}.
The representations are further augmented
to multi-scale contextual representations~\cite{ZhaoSQWJ17,ChenPKMY18,ChenYWXY16} through feature pyramids
for segmenting objects at multiple scales.

\vspace{.1cm}
\noindent\textbf{Recovering high-resolution representations.}
An upsample process
can be used
to gradually recover the high-resolution representations
from the low-resolution representations.
The upsample subnetwork could be a symmetric version
of the downsample process (e.g., VGGNet),
with skipping connection over some mirrored layers 
to transform the pooling indices, 
e.g., SegNet~\cite{BadrinarayananK17} and DeconvNet~\cite{NohHH15},
or copying the feature maps, e.g., U-Net~\cite{RonebergerFB15} and Hourglass~\cite{NewellYD16, ChuYOMYW17, YangLOLW17, KeCQL18, YangLZ17, BulatT17, DengTZZ17, BulatT17a, TangYW18},
encoder-decoder~\cite{PengFWM16}, and so on.
An extension of U-Net,
full-resolution residual network~\cite{PohlenHML17},
introduces an extra full-resolution stream
that carries information at the full image resolution,
to replace the skip connections,
and each unit in the downsample and upsample subnetworks
receives information from and sends information to
the full-resolution stream.

The asymmetric upsample process is also widely studied.
RefineNet~\cite{LinMSR17} improves the combination 
of upsampled representations and
the representations of the same resolution
copied from the downsample process.
Other works include: 
light upsample process~\cite{BulatT16, ChenWPZYS17,XiaoWW18, LinDGHHB17},
possibly with dilated convolutions used in the backbone~\cite{InsafutdinovPAA16, PishchulinITAAG16, LifshitzFU16};
light downsample and heavy upsample processes~\cite{ValleBVB18}, 
recombinator networks~\cite{HonariYVP16}; 
improving skip connections with more or complicated convolutional units~\cite{PengZYLS17, ZhangZPXS18, IslamRBW17}, as well as sending information from low-resolution skip connections to high-resolution skip connections~\cite{ZhouSTL18}
or exchanging information between them~\cite{GuoDXZ18};
studying the details of the upsample process~\cite{WojnaUGSCFF17};
combining multi-scale pyramid representations~\cite{ChenZPSA18, XiaoLZJS18};
stacking multiple DeconvNets/U-Nets/Hourglass~\cite{FuLWL17, Wu0YWC018}
with dense connections~\cite{TangPGWZM18}.

\vspace{.1cm}
\noindent\textbf{Maintaining high-resolution representations.}
Our work is closely related to several works
that can also generate high-resolution representations,
e.g., convolutional neural fabrics~\cite{SaxenaV16}, interlinked CNNs~\cite{ZhouHZ15},
GridNet~\cite{FourureEFMT017}, 
and multi-scale DenseNet~\cite{HuangCLWMW17}.

The two early works, convolutional neural fabrics~\cite{SaxenaV16} 
and interlinked CNNs~\cite{ZhouHZ15},
lack careful design 
on when to start low-resolution parallel streams,
and how and where to exchange information across parallel streams,
and do not use batch normalization and residual connections,
thus not showing satisfactory performance.
GridNet~\cite{FourureEFMT017} 
is like a combination of multiple U-Nets
and includes two symmetric information exchange stages:
the first stage passes information 
only 
from high resolution
to low resolution,
and the second stage passes information
only from low resolution to high resolution.
This limits its segmentation quality.
Multi-scale DenseNet~\cite{HuangCLWMW17} is not able to learn 
strong high-resolution representations
as there is no information received from low-resolution representations.

\vspace{.1cm}
\noindent\textbf{Multi-scale fusion.}
Multi-scale fusion\footnote{In this paper,
Multi-scale fusion
and multi-resolution fusion are interchangeable,
but in other contexts, they may not be interchangeable.}
is widely studied~\cite{CaiFFV16, ChenPKMY18, ChenWPZYS17, FourureEFMT017, XieT15, ZhaoSQWJ17, KanazawaSJ14, XuXZYZ14, SamyAESE18, SaxenaV16, HuangCLWMW17, ZhouHZ15}.
The straightforward way is 
to feed multi-resolution images separately into 
multiple networks and aggregate the output response maps~\cite{TompsonGJLB15}.
Hourglass~\cite{NewellYD16}, U-Net~\cite{RonebergerFB15}, and SegNet~\cite{BadrinarayananK17} combine low-level features in the high-to-low downsample process into 
the same-resolution high-level features in 
the low-to-high upsample process progressively through skip connections.
PSPNet~\cite{ZhaoSQWJ17} and DeepLabV2/3~\cite{ChenPKMY18}
fuse the pyramid features obtained by pyramid pooling module and atrous spatial pyramid pooling.
Our multi-scale (resolution) fusion module resembles the two pooling modules.
The differences include: (1) Our fusion outputs four-resolution representations
other than only one,
and (2) our fusion modules are repeated several times
which is inspired by deep fusion~\cite{WangWZZ16, XieW0LHQ18, SunLLW18, ZhangQ0W17, ZhaoLMLZZTW18}.

\vspace{.1cm}
\noindent\textbf{Our approach.}
Our network connects high-to-low convolution streams in parallel.
It maintains high-resolution representations
through the whole process,
and generates reliable high-resolution representations
with strong position sensitivity
through repeatedly fusing the representations
from multi-resolution streams.

This paper represents a very substantial extension of our
previous conference paper~\cite{SunXLW19}
with an additional material added from our unpublished technical report~\cite{SunZJCXLMWLW19}
as well as more object detection results
under recently-developed start-of-the-art 
object detection and instance segmentation frameworks.
The main technical novelties compared with~\cite{SunXLW19} lie in threefold.
(1) We extend the network (named as HRNetV$1$) proposed in~\cite{SunXLW19},
to two versions: HRNetV$2$ and HRNetV$2$p,
which explore all the four-resolution representations.
(2) We build the connection between multi-resolution fusion
and regular convolution,
which provides an evidence for
the necessity of 
exploring all the four-resolution representations
in HRNetV$2$ and HRNetV$2$p.
(3) We show the superiority of
HRNetV$2$ and HRNetV$2$p 
over HRNetV$1$
and 
present the applications of
HRNetV$2$ and HRNetV$2$p 
in a broad range of vision problems, 
including semantic segmentation
and object detection.

\section{High-Resolution Networks}
\label{sec:HRNet}
We input the image into a stem,
which consists of two stride-$2$ $3\times 3$ convolutions decreasing
the resolution to $\frac{1}{4}$,
and subsequently the main body
that outputs the representation with the same resolution ($\frac{1}{4}$).
The main body, illustrated in Figure~\ref{fig:HRNet}
and detailed below,
consists of several components:
parallel multi-resolution convolutions,
repeated multi-resolution fusions, 
and representation head
that is shown in Figure~\ref{fig:highresolutionneck}.

\subsection{Parallel Multi-Resolution Convolutions}
We start from a high-resolution convolution stream as the first stage,
gradually add high-to-low resolution streams one by one,
forming new stages,
and connect the multi-resolution streams in parallel.
As a result,
the resolutions for the parallel streams of a later stage  
consists of the resolutions from the previous stage,
and an extra lower one.

An example network structure
illustrated in Figure~\ref{fig:HRNet}, containing $4$ parallel streams, 
is logically as follows,
\begin{equation}
\begin{array}{llll}
\mathcal{N}_{11}  & \rightarrow ~~ \mathcal{N}_{21} &\rightarrow ~~ \mathcal{N}_{31} & \rightarrow ~~ \mathcal{N}_{41} \\
& \searrow ~~ \mathcal{N}_{22} &\rightarrow ~~ \mathcal{N}_{32} & \rightarrow ~~ \mathcal{N}_{42} \\
&   &\searrow  ~~ \mathcal{N}_{33} & \rightarrow ~~ \mathcal{N}_{43}  \\
&   & & \searrow ~~ \mathcal{N}_{44},
\end{array}
\end{equation}
where
$\mathcal{N}_{sr}$ is a sub-stream
in the $s$th stage
and 
$r$ is the resolution index.
The resolution index of the first stream
is $r=1$.
The resolution of index $r$ is $\frac{1}{2^{r-1}}$ of the resolution of the first stream.

\begin{figure}[t]
\small
\centering
\includegraphics[width=1.0\linewidth]{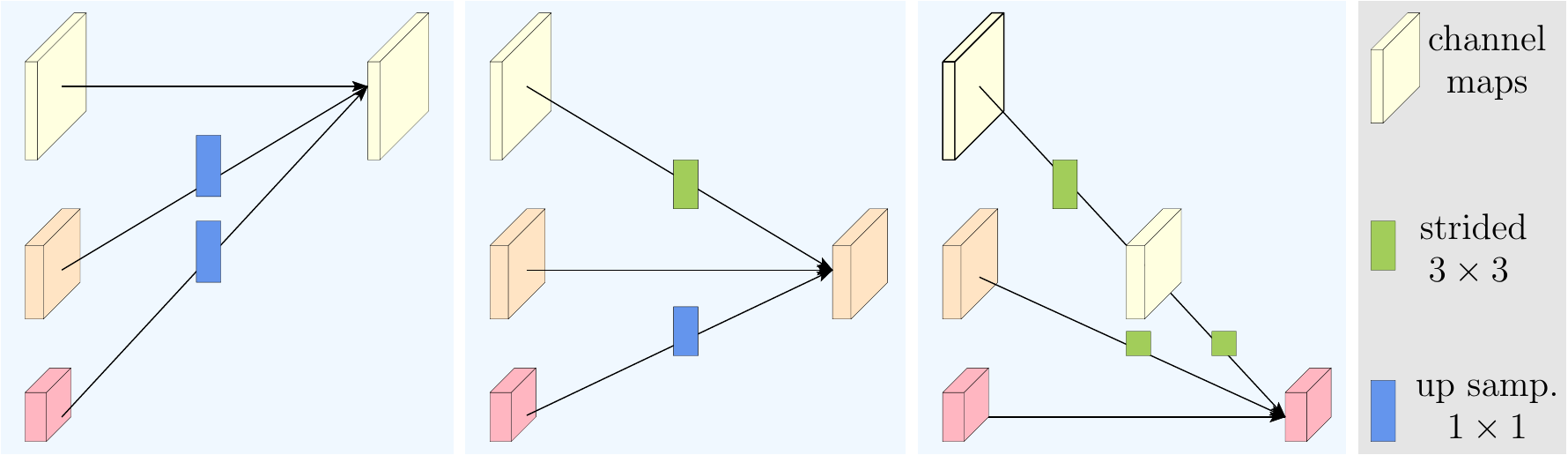}\\
\vspace{-0.2cm}
\caption{Illustrating
how 
the fusion module
aggregates the information
for high, medium and low resolutions
from left to right,
respectively.
Right legend: strided $3\times 3$ = stride-$2$ $3\times 3$ convolution,
up samp. $1\times 1$ = bilinear upsampling followed by a $1\times 1$ convolution.}
\label{fig:fusionmodule}
\vspace{-3mm}
\end{figure}

\subsection{Repeated Multi-Resolution Fusions}
The goal of the fusion module 
is to exchange the information 
across multi-resolution representations.
It is repeated
several times (e.g., every $4$ residual units).

Let us look at an example of fusing $3$-resolution representations, which is illustrated in Figure~\ref{fig:fusionmodule}.
Fusing $2$ representations and $4$ representations can be
easily derived.
The input consists of three representations:
$\{\mathbf{R}_r^i, r=1, 2, 3\}$,
with $r$ is the resolution index,
and the associated output representations
are $\{\mathbf{R}_r^o, r=1, 2, 3\}$. 
Each output representation 
is the sum of the transformed representations of the three inputs:
$\mathbf{R}_r^o = f_{1r}(\mathbf{R}_1^i)
+ f_{2r}(\mathbf{R}_2^i) + f_{3r}(\mathbf{R}_3^i)$.
The fusion across stages 
(from stage $3$ to stage $4$)
has an extra output:
$\mathbf{R}_4^o = f_{14}(\mathbf{R}_1^i)
+ f_{24}(\mathbf{R}_2^i) + f_{34}(\mathbf{R}_3^i)$.

The choice of the transform function $f_{xr}(\cdot)$
is dependent on the input resolution index $x$
and the output resolution index $r$.
If $x=r$, $f_{xr}(\mathbf{R}) = \mathbf{R}$.
If $x<r$, $f_{xr}(\mathbf{R})$ downsamples
the input representation $\mathbf{R}$
through $(r-s)$ stride-$2$ $3\times 3$ convolutions.
For instance,
one stride-$2$ $3 \times 3$ convolution
for $2\times$ downsampling,
and two consecutive 
stride-$2$ $3 \times 3$ convolutions
for $4\times$ downsampling.
If $x>r$, $f_{xr}(\mathbf{R})$ upsamples
the input representation $\mathbf{R}$
through the bilinear upsampling 
followed by a $1\times 1$ convolution
for aligning the number of channels.
The functions are depicted in Figure~\ref{fig:fusionmodule}.

\begin{figure*}[t]
\footnotesize
    \centering
    \subfloat[]{\includegraphics[scale=.7,angle=90]{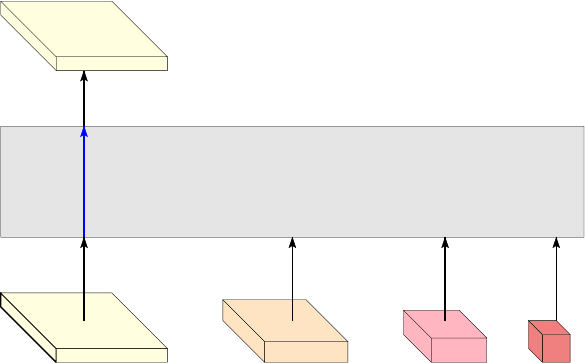}}~~~~~~~~~~~~~~~~~~~~
    \subfloat[]{\includegraphics[scale=.7,angle=90]{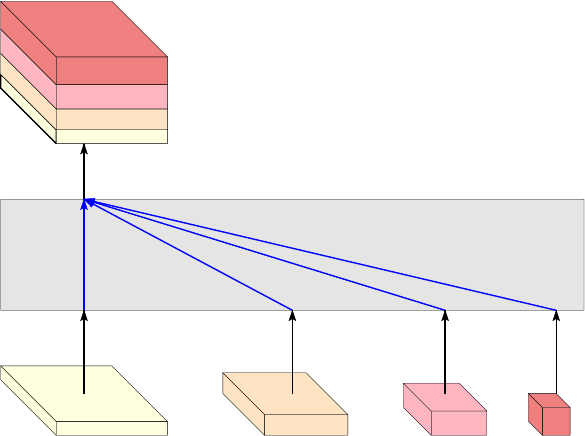}}~~~~~~~~~~~~~~~~~~~~
    \subfloat[]{\includegraphics[scale=.7,angle=90]{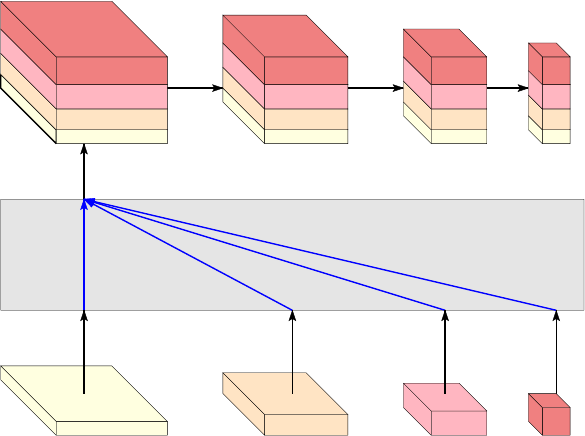}}
	\vspace{-.2cm}
    \caption{
    (a) HRNetV$1$: only output 
    the representation
    from the high-resolution convolution stream.
    (b) HRNetV$2$: Concatenate the (upsampled) representations
    that are from all the resolutions
    (the subsequent $1\times 1$ convolution is not shown for clarity).
    (c) HRNetV$2$p: form a feature pyramid
    from the representation by HRNetV$2$.
    The four-resolution representations at the bottom in each sub-figure are outputted from the network in Figure~\ref{fig:HRNet}, 
    and the gray box indicates how the output representation is obtained from the input four-resolution representations.}
    \label{fig:highresolutionneck}
    \vspace{-3mm}
\end{figure*}

\subsection{Representation Head}
We have three kinds of representation heads
that are illustrated in Figure~\ref{fig:highresolutionneck},
and call them as
HRNetV$1$, HRNetV$2$, and HRNetV$1$p,
respectively.

\vspace{.1cm}
\noindent\textbf{HRNetV$1$.}
The output is 
the representation only from the high-resolution stream.
Other three representations are ignored. 
This is illustrated in Figure~\ref{fig:highresolutionneck} (a).

\vspace{.1cm}
\noindent\textbf{HRNetV$2$.}
We rescale the low-resolution representations through bilinear upsampling without changing the number of channels
to the high resolution,
and concatenate the four representations,
followed by a $1\times 1$ convolution to mix the four representations.
This is illustrated in Figure~\ref{fig:highresolutionneck} (b).

\vspace{.1cm}
\noindent\textbf{HRNetV$2$p.}
We construct multi-level representations
by downsampling the high-resolution representation
output from HRNetV$2$
to multiple levels.
This is depicted in Figure~\ref{fig:highresolutionneck} (c).

In this paper,
we will show the results
of applying HRNetV$1$ to human pose estimation,
HRNetV$2$ to semantic segmentation,
and HRNetV$2$p to object detection.

\begin{figure}[t]
    \centering
    \footnotesize
   \subfloat[]{\includegraphics[height=0.33\linewidth]{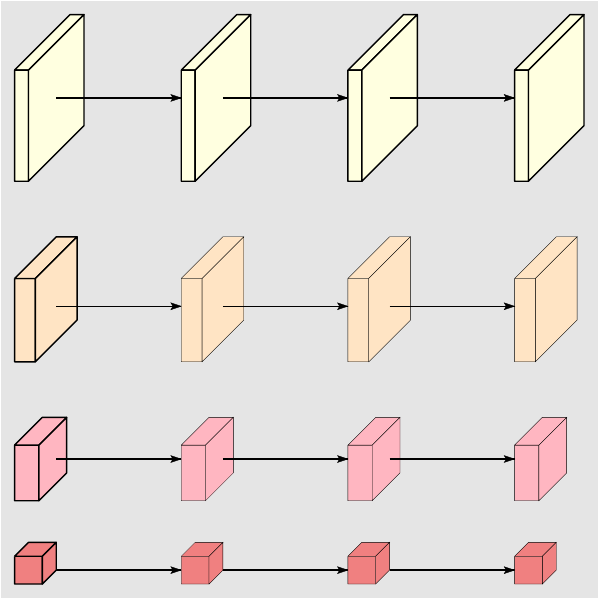}}~~
    \subfloat[]{\includegraphics[height=0.33\linewidth]{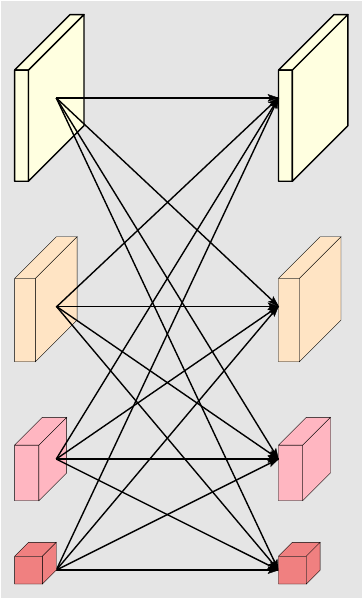}}~~
    \subfloat[]{\includegraphics[height=0.33\linewidth]{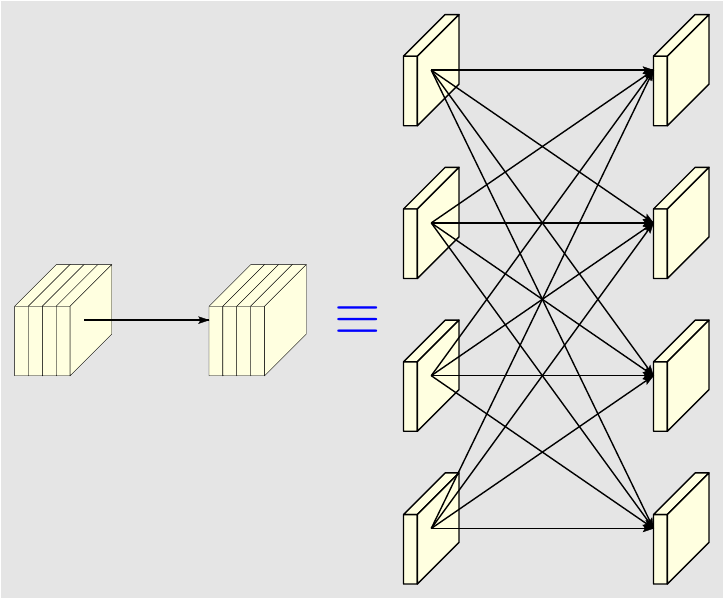}}
	\vspace{-.2cm}
    \caption{
    (a) Multi-resolution parallel convolution,
    (b) multi-resolution fusion.
    (c) A normal convolution (left) is equivalent to
    fully-connected multi-branch convolutions (right).}
    \label{fig:multiresolutionblock}
    \vspace{-3mm}
\end{figure}

\subsection{Instantiation}
The main body 
contains four stages
with four parallel convolution streams.
The resolutions are $1/4$, $1/8$, $1/16$, and $1/32$. 
The first stage contains $4$ residual units 
where each unit
is formed by a bottleneck with the width $64$,
and is followed by one $3 \times3$ convolution
changing the width of feature maps
to $C$.
The $2$nd, $3$rd, $4$th stages
contain $1$, $4$, $3$ modularized blocks, respectively.
Each branch in multi-resolution parallel convolution of 
the modularized block 
contains $4$ residual units. 
Each unit contains two $3\times3$ convolutions for each resolution,
where each convolution is followed by batch normalization and the nonlinear activation ReLU.
The widths (numbers of channels) of the convolutions
of the four resolutions 
are $C$, $2C$, $4C$, and $8C$, respectively.
An example is depicted in Figure~\ref{fig:HRNet}.

\subsection{Analysis}
We analyze the modularized block 
that is divided into
two components: multi-resolution parallel convolutions
(Figure~\ref{fig:multiresolutionblock} (a)),
and multi-resolution fusion (Figure~\ref{fig:multiresolutionblock} (b)).
The multi-resolution parallel convolution resembles the group convolution. It divides the input channels 
into several subsets of channels 
and performs a regular convolution 
over each subset 
over different spatial resolutions separately,
while in the group convolution,
the resolutions are the same.
This connection implies that
the multi-resolution parallel convolution
enjoys some benefit of the group convolution. 

The multi-resolution fusion unit 
resembles the multi-branch full-connection form
of the regular convolution, illustrated in Figure~\ref{fig:multiresolutionblock} (c).
A regular convolution 
can be divided as multiple small convolutions
as explained in~\cite{ZhangQ0W17}.
The input channels are divided into several subsets,
and the output channels are also divided into several subsets.
The input and output subsets are connected 
in a fully-connected fashion,
and each connection is a regular convolution.
Each subset of output channels
is a summation of the outputs of the convolutions 
over each subset of input channels. 
The differences lie in that our multi-resolution fusion
needs to handle the resolution change.
The connection between multi-resolution fusion
and regular convolution provides
an evidence
for exploring all the four-resolution representations
done in HRNetV$2$ and HRNetV$2$p.

\begin{figure*}[t]
	\centering
	\includegraphics[height = 0.162\textwidth]{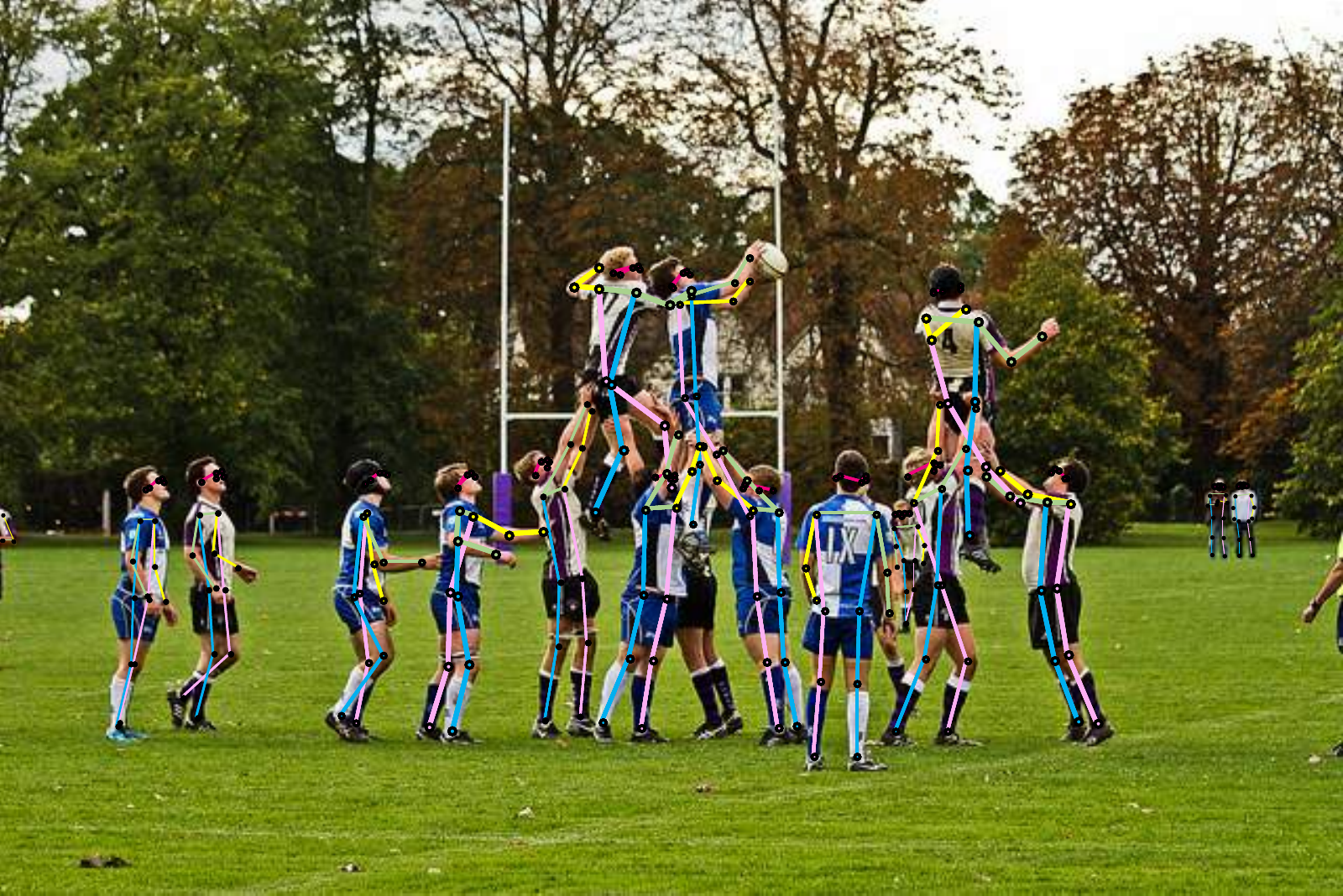}
	\includegraphics[height = 0.162\textwidth]{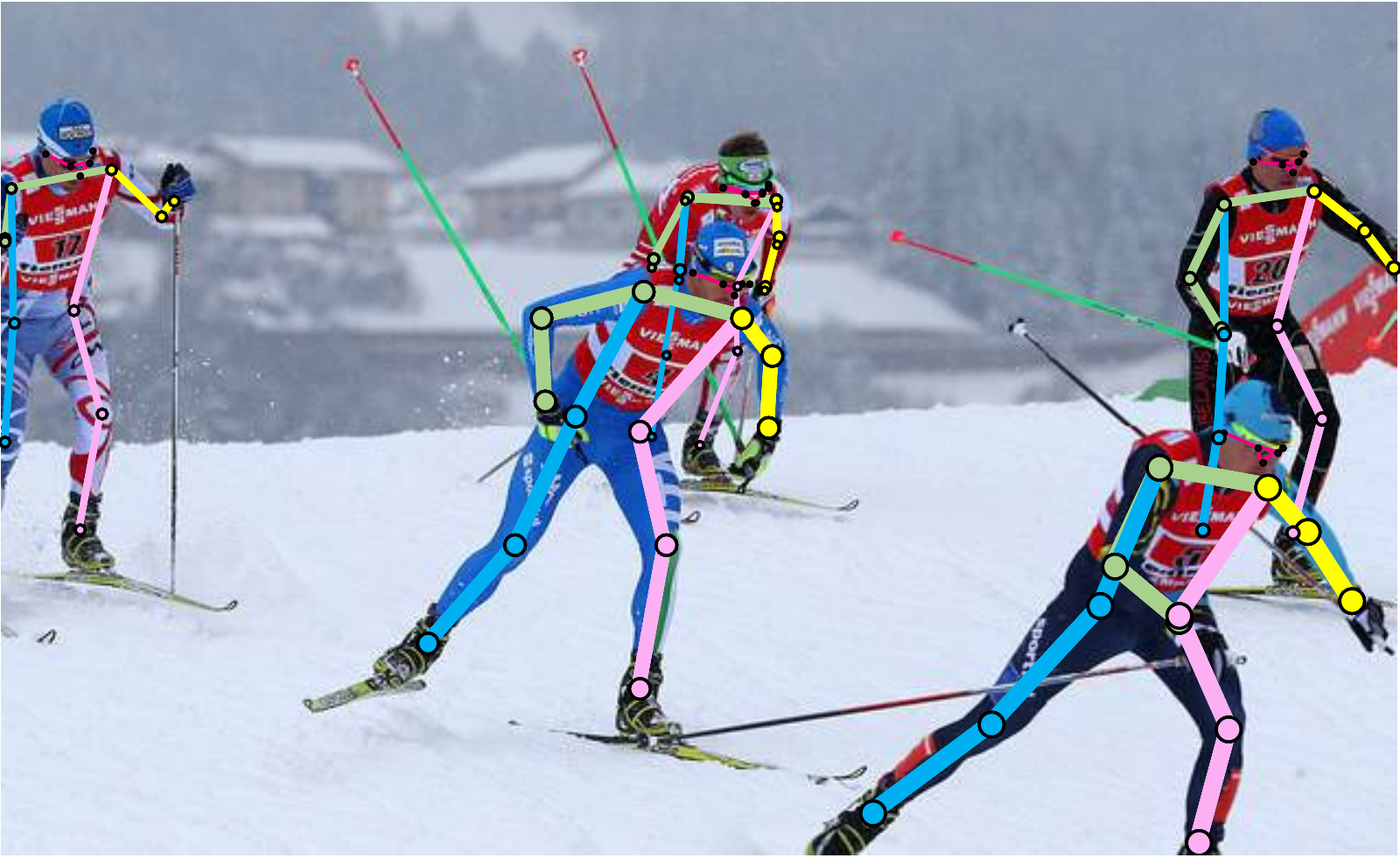}
	\includegraphics[height = 0.162\textwidth]{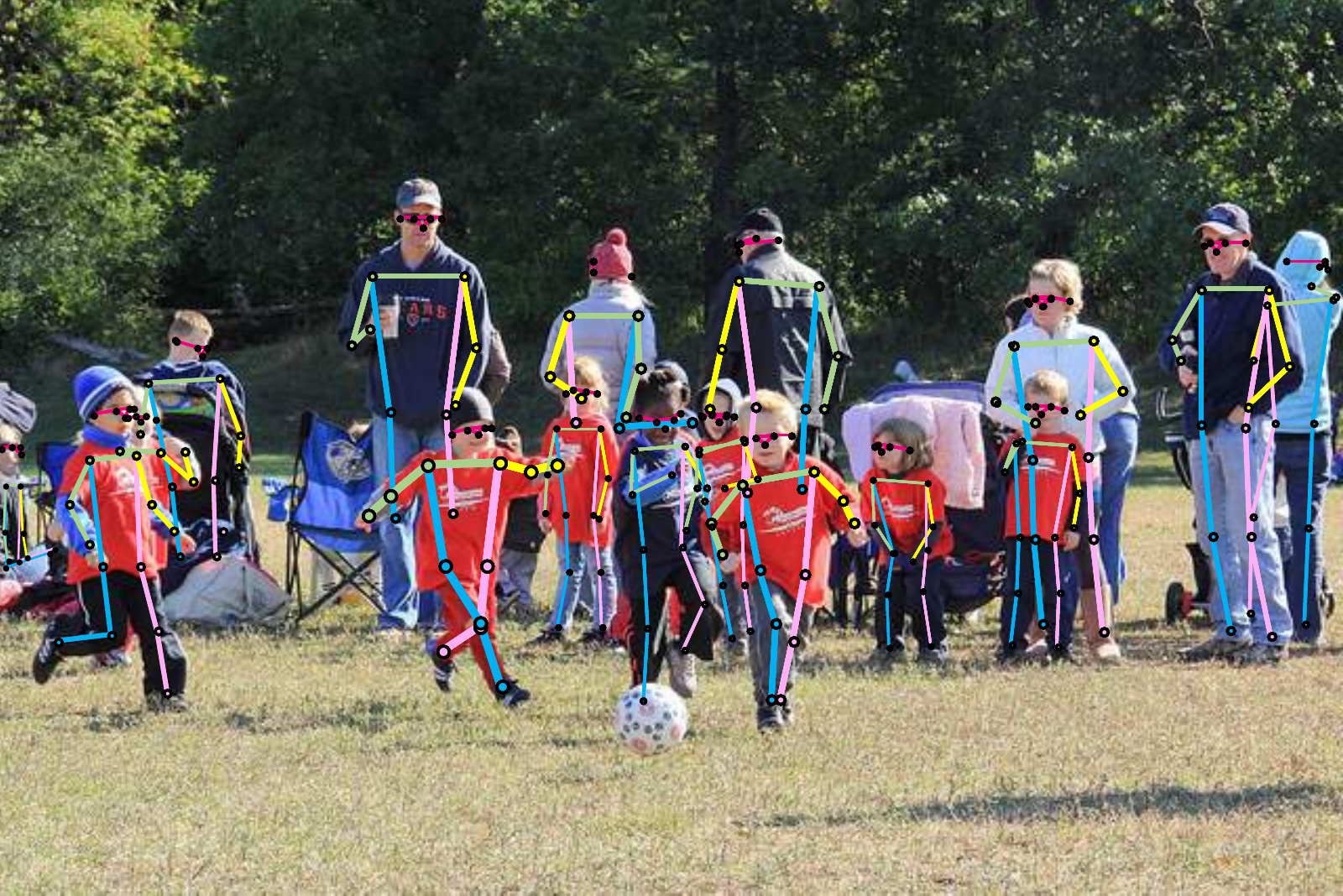}
	\includegraphics[height = 0.162\textwidth]{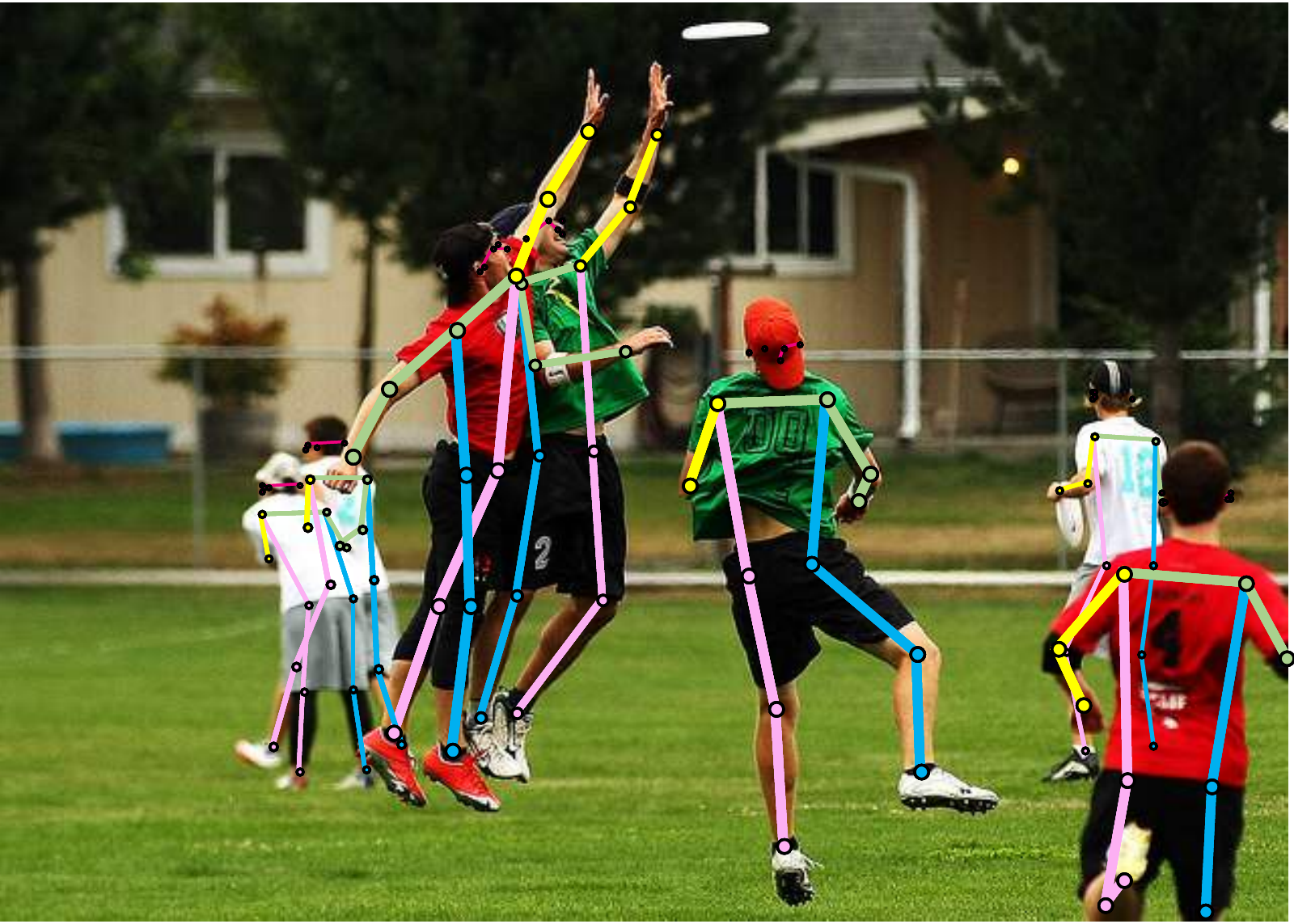}\\
	\vspace{-3mm}
	\caption{Qualitative COCO human pose estimation results
	{over representative images with various human size, different poses,
	or clutter background}.}
	\label{fig:COCOPoseEstimation}
	\vspace{-3mm}
\end{figure*}

\renewcommand{\arraystretch}{1.1}
\begin{table*}[t]
\footnotesize
\centering
\caption{Comparisons on COCO \texttt{val}. 
{
Under the input size $256 \times 192$, 
our approach with a small model HRNetV$1$-W$32$, trained from scratch,
performs better than previous state-of-the-art methods.
Under the input size $384 \times 288$,
our approach with a small model HRNetV$1$-W$32$
achieves a higher AP score
than SimpleBaseline with a large model.
In particular,
the improvement of our approach 
for $\operatorname{AP}^{75}$,
a strict evaluation scheme,
is more significant
than $\operatorname{AP}^{50}$,
a loose evaluation scheme.
}
Pretrain = pretrain the backbone on ImageNet. 
OHKM = online hard keypoints mining~\cite{ChenWPZYS17}.
\#Params and FLOPs are calculated 
for the pose estimation network, and 
those for human detection and keypoint grouping are not included.}
\label{table:coco_val}
\vspace{-0.3cm}
\begin{tabular}{l|l|c|r|r|c|cccccc}
\hline
Method & Backbone &  Pretrain & Input size & \#Params & GFLOPs & 
$\operatorname{AP}$ & $\operatorname{AP}^{50}$ & $\operatorname{AP}^{75}$ & $\operatorname{AP}^{M}$ & $\operatorname{AP}^{L}$ & $\operatorname{AR}$ \\
\hline
$8$-stage Hourglass~\cite{NewellYD16} & $8$-stage Hourglass & N &  $256 \times 192$ & $25.1$M & $14.3$&
$66.9$&$-$&$-$&$-$&$-$&$-$\\ 
CPN~\cite{ChenWPZYS17}& ResNet-50 & Y & $256 \times 192$ & $27.0$M & $6.20$&
$68.6$&$-$&$-$&$-$&$-$&$-$\\ 
CPN + OHKM~\cite{ChenWPZYS17} & ResNet-50 & Y & $256 \times 192$ & $27.0$M & $6.20$&
$69.4$&$-$&$-$&$-$&$-$&$-$\\ 
SimpleBaseline~\cite{XiaoWW18} & ResNet-50 & Y & $256\times192$  &$34.0$M &$8.90$
&${70.4}$ & ${88.6}$&${78.3}$&${67.1}$&${77.2}$&${76.3}$\\
SimpleBaseline~\cite{XiaoWW18} & ResNet-101 & Y & $256\times192$  &$53.0$M &$12.4$
&${71.4}$ & ${89.3}$&${79.3}$&${68.1}$&${78.1}$&${77.1}$\\
SimpleBaseline~\cite{XiaoWW18} & ResNet-152  & Y & $256\times192$ &$68.6$M &$15.7$
&${72.0}$ & ${89.3}$&${79.8}$&${68.7}$&${78.9}$&${77.8}$\\
\hline
HRNetV$1$ & HRNetV$1$-W$32$ & N & $256\times 192$&  $28.5$M & $7.10$ &
$73.4$&$89.5$&$80.7$&$70.2$&$80.1$&$78.9$  \\
HRNetV$1$ & HRNetV$1$-W$32$& Y & $256\times 192$&  $28.5$M & $7.10$ &
$74.4$&$90.5$&$81.9$&$70.8$&$81.0$&$79.8$  \\
HRNetV$1$ & HRNetV$1$-W$48$& Y &  $256\times 192$&  $63.6$M &$14.6$ &
$75.1$&$90.6$&$82.2$&$71.5$&$81.8$&$80.4$  \\
\hline
SimpleBaseline~\cite{XiaoWW18} & ResNet-152  & Y &  $384\times288$     &$68.6$M &$35.6$
&${74.3}$ & ${89.6}$&${81.1}$&${70.5}$&${79.7}$&${79.7}$\\
HRNetV$1$ &HRNetV$1$-W$32$ & Y &  $384\times 288$&  $28.5$M &$16.0$ & $75.8$&$90.6$&${82.7}$&$71.9$&$82.8$&$81.0$  \\
HRNetV$1$ & HRNetV$1$-W$48$& Y &  $384\times 288$&  $63.6$M &$32.9$ & $\mathbf{76.3}$&$\mathbf{90.8}$&$\mathbf{82.9}$&$\mathbf{72.3}$&$\mathbf{83.4}$&$\mathbf{81.2}$  \\
\hline
\end{tabular}
\vspace{-1mm}
\end{table*}

	\begin{table*}[t]
		\caption{Comparisons on COCO \texttt{test-dev}.   
		{The observations are similar to
		the results on COCO \texttt{val}.
		}
		}
	\centering
			\label{table:coco_test_dev}
			\footnotesize
			\vspace{-0.3cm}
            \begin{tabular}{l|l|c|c|c|lllllc}
				\hline
				Method &Backbone& Input size & \#Params & GFLOPs&
				$\operatorname{AP}$ & $\operatorname{AP}^{50}$ & $\operatorname{AP}^{75}$ & $\operatorname{AP}^{M}$ & $\operatorname{AP}^{L}$ & $\operatorname{AR}$\\
				\hline
				\multicolumn{11}{c}{Bottom-up: keypoint detection and grouping}\\
				\hline
				OpenPose~\cite{CaoSWS17} &$\hfil-$& $-$ &$-$& $-$  
				&$61.8$ & $84.9$&$67.5$&$57.1$&$68.2$&$66.5$\\
				Associative Embedding~\cite{NewellHD17} & $\hfil-$ & $-$ &$-$& $-$
				&$65.5$ & $86.8$&$72.3$&$60.6$&$72.6$&$70.2$\\
				PersonLab~\cite{PapandreouZCGTK18} & $\hfil-$ & $-$ &$-$& $-$
				&$68.7$ & $89.0$&$75.4$&$64.1$&$75.5$&$75.4$\\
				MultiPoseNet~\cite{KocabasKA18} & $\hfil-$ & $-$ &$-$& $-$
				&$69.6$ & $86.3$&$76.6$&$65.0$&$76.3$&$73.5$\\
				\hline 
				\multicolumn{11}{c}{Top-down: human detection and single-person keypoint detection}\\
				\hline
				Mask-RCNN~\cite{HeGDG17} & ResNet-50-FPN& $-$ &$-$& $-$
				& $63.1$ & $87.3$&$68.7$&$57.8$&$71.4$&$-$\\
				G-RMI~\cite{PapandreouZKTTB17} & ResNet-101 & $353\times257$ &$42.6$M& $57.0$
				&$64.9$ & $85.5$&$71.3$&$62.3$&$70.0$&$69.7$\\
				Integral Pose Regression~\cite{SunXWLW18} & ResNet-101 & $256\times256$ &$45.0$M& $11.0$
				&$67.8$ & $88.2$&$74.8$&$63.9$&$74.0$&$-$\\
				G-RMI + extra data~\cite{PapandreouZKTTB17} & ResNet-101 & $353\times257$ &$42.6$M& $57.0$
				&$68.5$ & $87.1$&$75.5$&$65.8$&$73.3$&$73.3$\\
				CPN~\cite{ChenWPZYS17} & ResNet-Inception& $384\times288$ &$-$& $-$
				& $72.1$ & $91.4$&$80.0$&$68.7$&$77.2$&$78.5$\\
				RMPE~\cite{FangXTL17} & PyraNet~\cite{YangLOLW17} & $320\times256$ &$28.1$M& $26.7$
				&$72.3$ & $89.2$&$79.1$&$68.0$&$78.6$&$-$\\
				CFN~\cite{HuangGT17} & $\hfil-$ & $-$ &$-$& $-$
				& $72.6$ & $86.1$&$69.7$&$78.3$&$64.1$&$-$\\
				CPN (ensemble)~\cite{ChenWPZYS17} & ResNet-Inception& $384\times288$ &$-$& $-$
				&$73.0$ & $91.7$&$80.9$&$69.5$&$78.1$&$ 79.0$\\
				SimpleBaseline~\cite{XiaoWW18} & ResNet-152&$384\times288$  &$68.6$M& $35.6$
				&${73.7}$ & ${91.9}$&${81.1}$&${70.3}$&${80.0}$&${79.0}$\\
				\hline
				HRNetV$1$ & HRNetV$1$-W$32$& $384\times 288$ &$28.5$M&$16.0$ &$74.9$&$92.5$&$82.8$&$71.3$&$80.9$&$80.1$\\
				HRNetV$1$ & HRNetV$1$-W$48$& $384\times 288$ &$63.6$M& $32.9$
				& ${75.5}$&${92.5}$&${83.3}$&${71.9}$&${81.5}$&${80.5}$\\
				\hline
				HRNetV$1$ + extra data & HRNetV$1$-W$48$& $384\times 288$ &$63.6$M& $32.9$
				& $\mathbf{77.0}$&$\mathbf{92.7}$&$\mathbf{84.5}$&$\mathbf{73.4}$&$\mathbf{83.1}$&$\mathbf{82.0}$\\
				\hline
			\end{tabular}
			\vspace{-1mm}
	\end{table*}
	
\section{Human Pose Estimation}
\label{sec:poseexperiments}

Human pose estimation, a.k.a. keypoint detection,
aims to detect the locations of $K$ keypoints or parts
(e.g., elbow, wrist, etc) from an image $\mathbf{I}$
of size $W \times H \times 3$.
We follow the state-of-the-art framework
and transform this problem
to estimating $K$ heatmaps of size $\frac{W}{4} \times \frac{H}{4}$,
$\{\mathbf{H}_1, \mathbf{H}_2, \dots, \mathbf{H}_K\}$, 
where each heatmap $\mathbf{H}_k$ indicates 
the location confidence of the $k$th keypoint.

We regress the heatmaps
over the high-resolution representations
output by HRNetV$1$.
We empirically observe
that the performance is almost the same
for HRNetV$1$
and HRNetV$2$,
and thus we choose HRNetV$1$
as its computation complexity is a little lower.
The loss function, defined as 
the mean squared error,
is applied for comparing
the predicted heatmaps
and the groundtruth heatmaps.
The groundtruth heatmaps are generated 
by applying $2$D Gaussian 
with standard deviation of $2$ pixel
centered on the groundtruth location
of each keypoint. 
Some example results are given
in Figure~\ref{fig:COCOPoseEstimation}.

\vspace{.1cm}
\noindent\textbf{Dataset.}
The COCO dataset~\cite{LinMBHPRDZ14} contains over 
$200,000$ images and $250,000$ person instances labeled with $17$ keypoints. 
We train our model on the COCO \texttt{train2017} set, 
including $57K$ images 
and $150K$ person instances. 
We evaluate our approach on the \texttt{val2017} and \texttt{test-dev2017} sets, 
containing $5000$ images and $20K$ images, respectively.

\vspace{.1cm}
\noindent\textbf{Evaluation metric.}
The standard evaluation metric
is
based on Object Keypoint Similarity (OKS):
$\operatorname{OKS} = \frac{\sum_{i}\exp(-d_i^2/2s^2k_i^2)\delta(v_i > 0)}{\sum_i \delta(v_i > 0)}.$
Here $d_i$ is the Euclidean distance between 
the detected keypoint and the corresponding ground truth,
$v_i$ is the visibility flag of the ground truth,
$s$ is the object scale, and 
$k_i$ is a per-keypoint constant that controls falloff.
We report standard average precision and recall scores\footnote{\url{http://cocodataset.org/\#keypoints-eval}}:
$\operatorname{AP}^{50}$ ($\operatorname{AP}$ at $\operatorname{OKS} = 0.50$),
$\operatorname{AP}^{75}$,
$\operatorname{AP}$ 
(the mean of $\operatorname{AP}$ scores at $10$ 
$\operatorname{OKS}$
positions,
$0.50, 0.55, \dots,0.90, 0.95$);
$\operatorname{AP}^M$  for medium objects,
$\operatorname{AP}^L$  for large objects,
and $\operatorname{AR}$ (the mean of $\operatorname{AR}$ scores at $10$ $\operatorname{OKS}$
positions,
$0.50, 0.55, \dots,0.90, 0.95$).

\vspace{.1cm}
\noindent\textbf{Training.}
We extend the human detection box in height or width 
to a fixed aspect ratio:
$\operatorname{height}: \operatorname{width} = 4 : 3$,
and then crop the box from the image,
which is resized to a fixed size, $256 \times 192$ or $384 \times 288$.
The data augmentation scheme includes
random rotation ($[\ang{-45}, \ang{45}] $),
random scale ($[0.65, 1.35]$), and flipping. 
Following ~\cite{wang2018mscoco}, half body data augmentation is also involved.

We use the Adam optimizer~\cite{KingmaB14}.
The learning schedule
follows the setting~\cite{XiaoWW18}.
The base learning rate is set as $1\mathrm{e}{-3}$,
and is dropped to $1\mathrm{e}{-4}$ and $1\mathrm{e}{-5}$ 
at the $170$th and $200$th epochs, respectively.
The training process is terminated within $210$ epochs.
{The models
are trained on
$4$ V$100$ GPUs
and it takes around $60$ ($80$) hours for HRNet-W$32$
(HRNet-W$48$).}

\vspace{.1cm}
\noindent\textbf{Testing.}
The two-stage top-down paradigm similar as~\cite{PapandreouZKTTB17,ChenWPZYS17,XiaoWW18}
is used:
detect the person instance using a person detector,
and then predict 
detection keypoints. 

We use the same person detectors provided 
by SimpleBaseline\footnote{\url{https://github.com/Microsoft/human-pose-estimation.pytorch}} for both the \texttt{val}
and \texttt{test-dev} sets. 
Following~\cite{XiaoWW18,NewellYD16,ChenWPZYS17},
we compute the heatmap by averaging the heatmaps of the original and flipped images.
Each keypoint location is predicted
by adjusting the highest heatvalue location with a quarter offset
in the direction from the highest response
to the second highest response.

\vspace{.1cm}
\noindent\textbf{Results on the \texttt{val} set.}
We report the results of our method and other state-of--the-art methods
in Table~\ref{table:coco_val}.
The network - HRNetV$1$-W$32$, trained from scratch
with the input size $256 \times 192$,
achieves an AP score $73.4$,
outperforming other methods 
with the same input size.
(\romannum{1}) Compared to Hourglass~\cite{NewellYD16},
our network improves AP by $6.5$ points, 
and the GFLOP of our network is much lower and less than half,
while the numbers of parameters are similar 
and ours is slightly larger.
(\romannum{2}) Compared to CPN~\cite{ChenWPZYS17} w/o and w/ OHKM, our network, with slightly larger model size and slightly higher complexity, achieves $4.8$ and $4.0$ points gain, respectively.
(\romannum{3}) Compared to the previous best-performed method SimpleBaseline~\cite{XiaoWW18},
our HRNetV$1$-W$32$
obtains significant improvements: $3.0$ points gain for the backbone ResNet-$50$ with a similar model size and GFLOPs, 
and $1.4$ points gain for the backbone ResNet-$152$ whose model size (\#Params) and GFLOPs
are twice as many as ours.

Our network can benefit from 
(\romannum{1}) training
from the model pretrained on the ImageNet:
The gain is $1.0$ points for HRNetV$1$-W$32$;
(\romannum{2}) increasing the capacity by increasing the width: HRNetV$1$-W$48$ gets $0.7$ and $0.5$ 
points gain for the input sizes $256\times192$ and $384\times288$, respectively.

Considering the input size $384 \times 288$,
our HRNetV$1$-W$32$ and HRNetV$1$-W$48$,
get the $75.8$ and $76.3$ AP, which have $1.4$ and $1.2$ improvements
compared to the input size $256 \times 192$.
In comparison to SimpleBaseline~\cite{XiaoWW18} that uses ResNet-$152$ as the backbone,
our HRNetV$1$-W$32$ and HRNetV$1$-W$48$ attain $1.5$ and $2.0$ points gain in terms of AP
at $45\%$ and $92.4\%$ computational cost, respectively.

\vspace{.1cm}
\noindent\textbf{Results on the \texttt{test-dev} set.}
Table~\ref{table:coco_test_dev} reports the pose estimation performances of our approach
and the existing state-of-the-art approaches.
Our approach is significantly better than 
bottom-up approaches.
On the other hand,
our small network, HRNetV$1$-W$32$, 
achieves an AP of $74.9$.
It outperforms all the other top-down approaches,
and is more efficient in terms of model size (\#Params) and computation complexity (GFLOPs).
Our big model, HRNetV$1$-W$48$, achieves the highest AP score $75.5$.
Compared to SimpleBaseline~\cite{XiaoWW18}
with the same input size, 
our small and big networks receive $1.2$ and $1.8$ improvements, respectively.
With the additional data from AI Challenger~\cite{wu2017ai} for training, our single big network can obtain an AP of $77.0$.

\begin{figure*}[t]
	\centering
	\includegraphics[height = 0.12\textwidth]{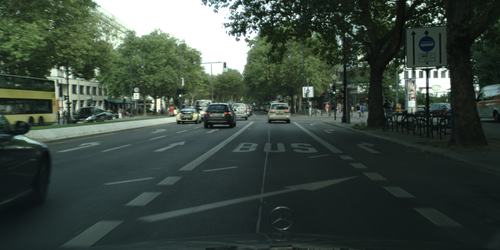}
	\includegraphics[height = 0.12\textwidth]{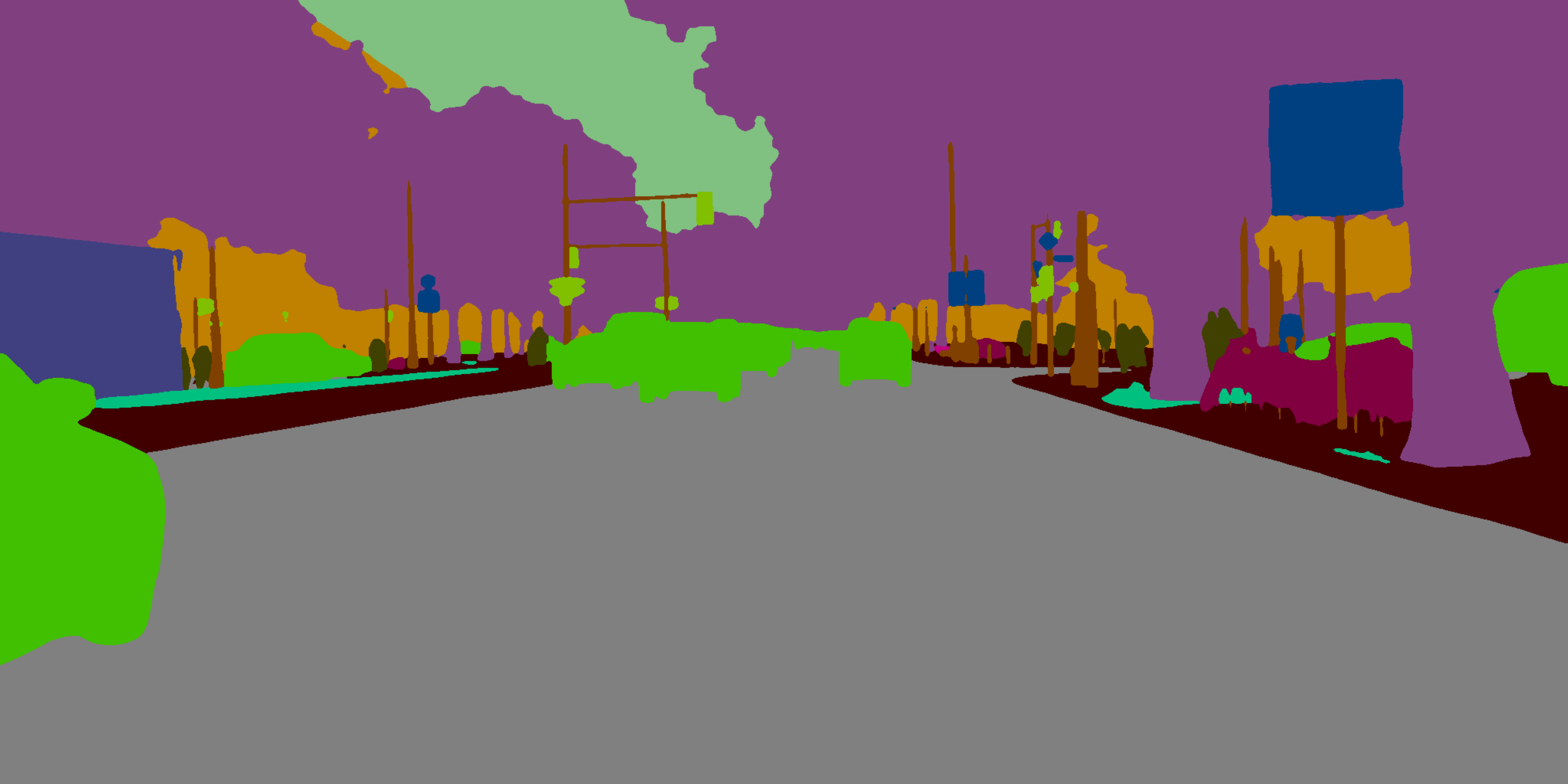}
	\includegraphics[height = 0.12\textwidth]{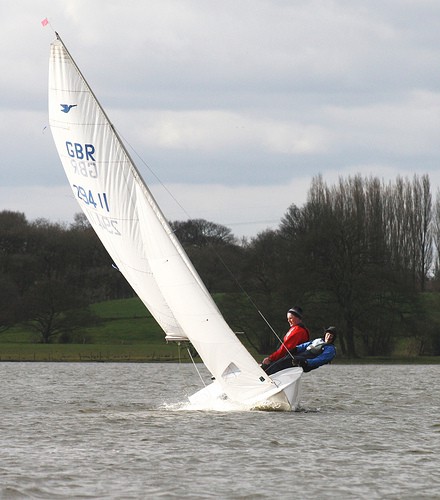}
	\includegraphics[height = 0.12\textwidth]{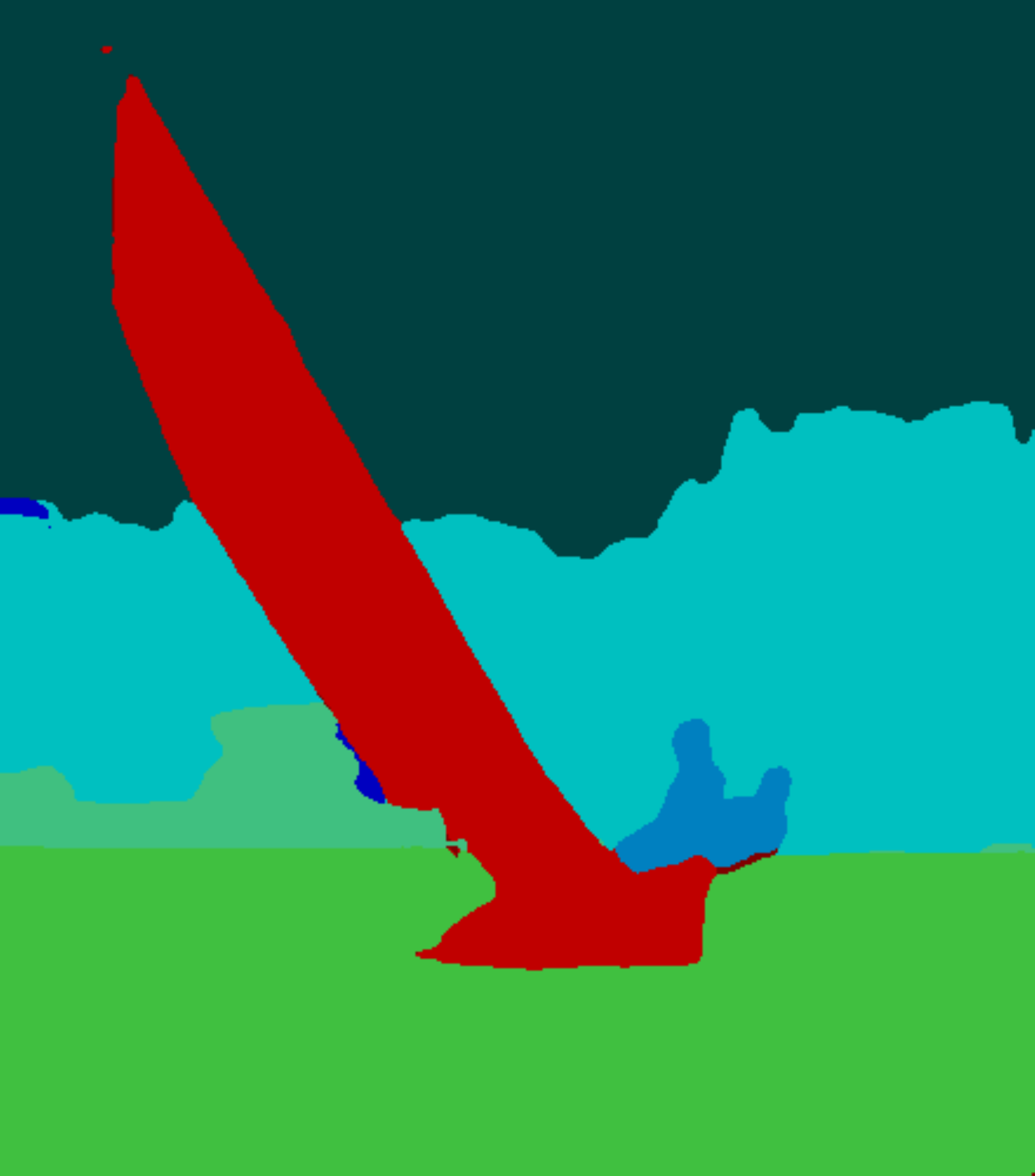}
	\includegraphics[height = 0.12\textwidth]{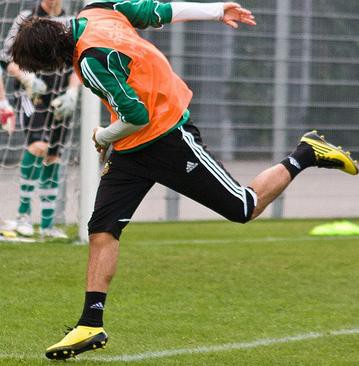}
	\includegraphics[height = 0.12\textwidth]{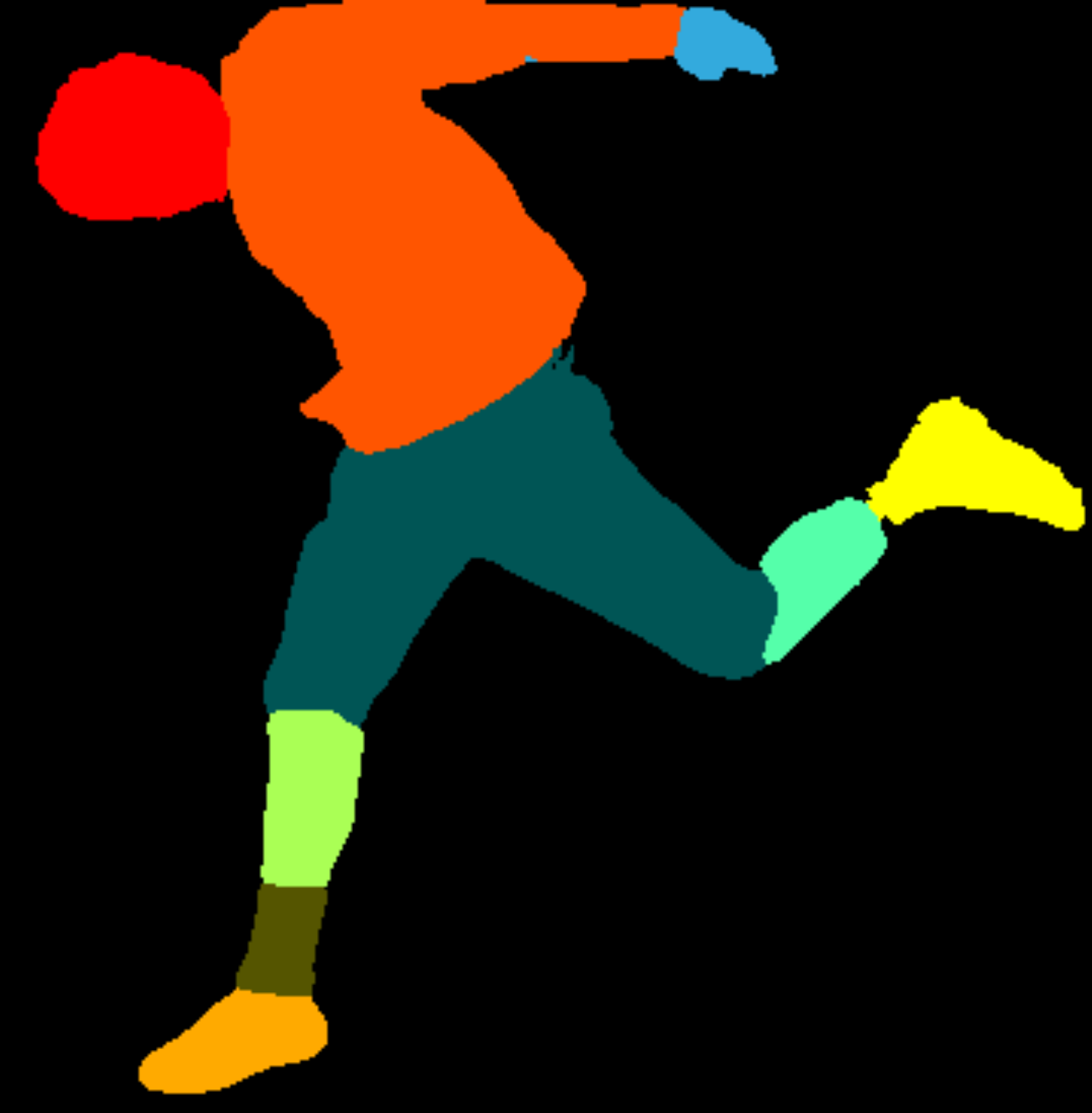}
	\vspace{-3mm}
	\caption{Qualitative segmentation examples
	{from Cityscapes (left two), PASCAL-Context (middle two), and LIP (right two).}}
	\label{fig:qualitativesegmentationresults}
	\vspace{-.2cm}
\end{figure*}

\setlength{\tabcolsep}{5.5pt}
	\begin{table}[t]
		\scriptsize
		\centering
		\caption{Semantic segmentation results on 
		Cityscapes \texttt{val}
	    (single scale and no flipping). 
		The GFLOPs is calculated on the input size $1024 \times 2048$.
		{The small model HRNetV$2$-W$40$
		with the smallest GFLOPs
		performs better than 
		two representative contextual methods (Deeplab and PSPNet).
		Our approach combined with 
		the recently-developed object contextual
		(OCR) representation scheme~\cite{YuanCW18} 
		gets further improvement.}
		D-ResNet-$101$ = Dilated-ResNet-$101$.
		}
		\label{tab:cityscapevalresults}
		\vspace{-3mm}
		\begin{tabular}{l|lrr|c}
			\hline%
			 & backbone & \#param. & GFLOPs & mIoU\\
			 \hline
			 
			 \hline
			UNet++~\cite{ZhouSTL18} & ResNet-$101$ & $59.5$M & $748.5$ & $75.5$ \\
			Dilated-ResNet~\cite{HeZRS16} & D-ResNet-$101$ & $52.1$M & $1661.6$ & $75.7$ \\
			DeepLabv3~\cite{ChenPSA17} & D-ResNet-$101$ & $58.0$M & $1778.7$ & $78.5$ \\
			DeepLabv3+~\cite{ChenZPSA18} & D-Xception-$71$ & $43.5$M & $1444.6$ & $79.6$ \\
			PSPNet~\cite{ZhaoSQWJ17} & D-ResNet-$101$ & $65.9$M & $2017.6$ & $79.7$ \\
			\hline
			HRNetV$2$ &  HRNetV$2$-W$40$ & $45.2$M & $493.2$ & $80.2$ \\
			HRNetV$2$ & HRNetV$2$-W$48$ & $65.9$M & $696.2$ & $81.1$ \\
			HRNetV$2$ + OCR~\cite{YuanCW18}& HRNetV$2$-W$48$ & $70.3$M & $1206.3$ & $\mathbf{81.6}$ \\
			\hline
		\end{tabular}
		\vspace{-.2cm}
	\end{table}
	
	\renewcommand{\arraystretch}{1.3}
	\setlength{\tabcolsep}{3pt}
	\begin{table}[t]
		\scriptsize
		\centering 
		\caption{Semantic segmentation results on Cityscapes \texttt{test}. 
		{We use HRNetV$2$-W$48$,
		whose parameter complexity and computation complexity are comparable to dilated-ResNet-$101$ based networks,
		for comparison. 
		Our results are superior
		in terms of the four evaluation metrics.
		The result
		from the combination with OCR~\cite{YuanCW18}
		is further improved.} D-ResNet-$101$ = Dilated-ResNet-$101$.}
		\label{tab:cityscaperesults}
		\vspace{-3mm}
		\begin{tabular}{l|l|cccc}
			\hline%
			  & backbone & mIoU  & iIoU cla. & IoU cat. & iIoU cat.\\
			\hline
			
			\hline
			\multicolumn{3}{l}{\emph {Model learned on the \texttt{train} set}}\\
			\hline
			PSPNet~\cite{ZhaoSQWJ17} & D-ResNet-$101$ & $78.4$ & $56.7$ & $90.6$  & $78.6$ \\
			PSANet~\cite{ZhaoZLSLLJ18} & D-ResNet-$101$ & $78.6$ & - & - & - \\
			PAN~\cite{LiXAW18} & D-ResNet-$101$ & $78.6$ & - & - & - \\
			AAF~\cite{KeHLY18} & D-ResNet-$101$ & $79.1$ & - & - & -\\
			\hline
			HRNetV$2$ & HRNetV$2$-W$48$ & $\mathbf{80.4}$ & $\mathbf{59.2}$ & $\mathbf{91.5}$ & $\mathbf{80.8}$\\
			\hline
			
			\hline
			\multicolumn{3}{l}{
			\emph {Model learned on the \texttt{train+val} set}}\\
			\hline
			GridNet~\cite{FourureEFMT017} & - & $69.5$ & $44.1$ & $87.9$ & $71.1$\\
			LRR-4x~\cite{GhiasiF16} & - & $69.7$ & $48.0$ & $88.2$ & $74.7$\\
			DeepLab~\cite{ChenPKMY18} & D-ResNet-$101$ & $70.4$ & $42.6$ & $86.4$ & $67.7$\\
			LC~\cite{LiLLLT17}& - & $71.1$ & - & - & - \\
			Piecewise~\cite{LinSHR16}& VGG-$16$ & $71.6$ & $51.7$ & $87.3$ & $74.1$\\
			FRRN~\cite{PohlenHML17}& - & $71.8$ & $45.5$ & $88.9$ & $75.1$\\
			RefineNet~\cite{LinMSR17}& ResNet-$101$ & $73.6$ & $47.2$ & $87.9$ & $70.6$\\
			PEARL~\cite{JinLXSLYCDLJFY17} & D-ResNet-$101$ & $75.4$ & $51.6$ & $89.2$ & $75.1$ \\
			DSSPN~\cite{LiangZX18} & D-ResNet-$101$ & $76.6$ & $56.2$ & $89.6$ & $77.8$\\
			LKM~\cite{PengZYLS17}& ResNet-$152$ & $76.9$ & - & - & - \\
			DUC-HDC~\cite{WangCYLHHC18}& - & $77.6$ & $53.6$ & $90.1$ & $75.2$\\
			SAC~\cite{ZhangTZLY17} & D-ResNet-$101$ & $78.1$ & - & - & - \\
			DepthSeg~\cite{KongF18} & D-ResNet-$101$ & $78.2$& - & - & - \\
			ResNet38~\cite{WuSH16e} & WResNet-38 &$78.4$ &$59.1$ &$90.9$ &$78.1$ \\
			BiSeNet~\cite{YuWPGYS182} & ResNet-$101$ & $78.9$ & - & - & - \\
			DFN~\cite{YuWPGYS18} & ResNet-$101$ & $79.3$ & - & - & - \\
			PSANet~\cite{ZhaoZLSLLJ18} & D-ResNet-$101$ & $80.1$ & - & - & - \\
			PADNet~\cite{OWS18} & D-ResNet-$101$ & $80.3$ & $58.8$ & $90.8$ & $78.5$\\
			CFNet~\cite{ZhangZWX} & D-ResNet-$101$ & $79.6$ & - & - & -\\
			Auto-DeepLab~\cite{liu2019auto} & - & $80.4$ & - & - & -\\
			DenseASPP~\cite{ZhaoSQWJ17} & WDenseNet-$161$ & $80.6$ & $59.1$ & $90.9$ & $78.1$ \\
			SVCNet~\cite{DingJSLW19} & ResNet-$101$ & $81.0$ & - & - & -\\
			ANN~\cite{ZhuXBHB2019} & D-ResNet-$101$ & $81.3$ & - & - & -\\
			CCNet~\cite{HuangWHHWL2019} & D-ResNet-$101$ & $81.4$ & - & - & -\\
            DANet~\cite{FuLYL19} & D-ResNet-$101$ & $81.5$ & - & - & - \\
			\hline
			HRNetV$2$ &  HRNetV$2$-W$48$  & $81.6$ & $\mathbf{61.8}$ & $\mathbf{92.1}$ & $\mathbf{82.2}$ \\
			HRNetV$2$ + OCR~\cite{YuanCW18} & HRNetV$2$-W$48$ &  $\mathbf{82.5}$ & $61.7$ &$\mathbf{92.1}$ & $81.6$\\
			\hline
		\end{tabular}
		\vspace{-2mm}
	\end{table}
	
	\renewcommand{\arraystretch}{1.3}
	\setlength{\tabcolsep}{3.0pt}
	\begin{table}[t]
	\scriptsize
	\centering
	\caption{Semantic segmentation results on PASCAL-Context. The methods are
	evaluated on $59$ classes and $60$ classes.
	{Our approach performs
	the best for $60$ classes,
	and performs worse for $59$ classes
	than APCN~\cite{HeDZWQ}
	that developed a strong contextual method.
	Our approach, combined with OCR~\cite{YuanCW18},
	achieves significant gain, and
	performs the best.
	}
	D-ResNet-$101$ = Dilated-ResNet-$101$.}
	\label{tab:pasctxresults}
	\vspace{-.3cm}
	\begin{tabular}{l@{~~~~}|@{~~~~}l@{~~~~}|@{~~~~}c@{~~~~}c}
		\hline%
		  & backbone & mIoU ($59$) & mIoU ($60$) \\
		\hline
		
		\hline
		FCN-$8$s~\cite{ShelhamerLD17} & VGG-$16$ & - & $35.1$ \\
		BoxSup~\cite{DaiHS15} & - & - & $40.5$ \\
		HO\_CRF~\cite{ArnabJ0T16} & - & - & $41.3$ \\
		Piecewise~\cite{LinSHR16} & VGG-$16$ & - &$43.3$ \\
		DeepLab-v$2$~\cite{ChenPKMY18} & D-ResNet-$101$ & -& $45.7$ \\
		RefineNet~\cite{LinMSR17} & ResNet-$152$ & - & $47.3$ \\
		UNet++~\cite{ZhouSTL18} & ResNet-$101$ & $47.7$ & - \\
		PSPNet~\cite{ZhaoSQWJ17} & D-ResNet-$101$ & $47.8$ & - \\
		Ding et al.~\cite{DingJSL018} & ResNet-$101$ & $51.6$ & - \\
		EncNet~\cite{0005DSZWTA18} & D-ResNet-$101$ & $52.6$ & - \\
		DANet~\cite{FuLYL19} & D-ResNet-$101$ & $52.6$ & - \\
		ANN~\cite{ZhuXBHB2019} & D-ResNet-$101$ & $52.8$ & - \\
		SVCNet~\cite{DingJSLW19} & ResNet-$101$ & $53.2$ & - \\
		CFNet~\cite{ZhangZWX} & D-ResNet-$101$ & ${54.0}$ & - \\
		\hline
		APCN~\cite{HeDZWQ} & D-ResNet-$101$ & 
		$55.6$
		& - \\
		\hline
		HRNetV$2$ & HRNetV$2$-W$48$ &  ${54.0}$ & $48.3$ \\
		HRNetV$2$ + OCR~\cite{YuanCW18} & HRNetV$2$-W$48$ &  $\mathbf{56.2}$ & $\mathbf{50.1}$ \\
		\hline
	\end{tabular} 
	\vspace{-.2cm}
	\end{table}

\renewcommand{\arraystretch}{1.3}
	\setlength{\tabcolsep}{2.8pt}
	\begin{table}[t]
	\scriptsize
	\centering
	\caption{Semantic segmentation results on LIP. Our method doesn't exploit 
	any extra information, e.g., pose or edge. 
	{
	The overall performance of our approach is the best, and the OCR scheme~\cite{YuanCW18}
	further improves the segmentation quality.}
	D-ResNet-$101$ = Dilated-ResNet-$101$.}
	\label{tab:lipresults}
	\vspace{-.3cm}
	\begin{tabular}{l|lc|ccc}
		\hline
		 & backbone & extra. & pixel acc. & avg. acc. & mIoU \\
		\hline
		
		\hline
		Attention+SSL~\cite{GongLSL17} & VGG$16$ & Pose & $84.36$ & $54.94$ & $44.73$ \\
		DeepLabV$3$+~\cite{ChenZPSA18} & D-ResNet-$101$ & - & $84.09$ & $55.62$ & $44.80$ \\
		MMAN~\cite{LuoZZGYY18} & D-ResNet-$101$ & - & - & - & $46.81$ \\
		SS-NAN~\cite{ZhaoLNZCWFY17} & ResNet-$101$ & Pose & $87.59$ & $56.03$ & $47.92$ \\
		MuLA~\cite{NieFY18} & Hourglass & Pose & $\mathbf{88.50}$ & $60.50$ & $49.30$ \\
		JPPNet~\cite{XL18} & D-ResNet-$101$ & Pose & $86.39$ & $62.32$ & $51.37$ \\
		CE2P~\cite{TL18}  & D-ResNet-$101$ & Edge & $87.37$ & $63.20$ & $53.10$ \\
		\hline
		HRNetV$2$ & HRNetV$2$-W$48$ & N & $88.21$ & $67.43$ & $55.90$ \\	
		HRNetV$2$ + OCR~\cite{YuanCW18} & HRNetV$2$-W$48$ & N & $88.24$ & $\mathbf{67.84}$ & $\mathbf{56.48}$\\
		\hline
	\end{tabular}
	 \vspace{-2mm}
	\end{table}

\section{Semantic Segmentation}
Semantic segmentation
is a problem of assigning
a class label to each pixel.
Some example results 
by our approach
are given in Figure~\ref{fig:qualitativesegmentationresults}.
We feed the input image 
to the HRNetV$2$ (Figure~\ref{fig:highresolutionneck} (b))
and then pass the resulting $15C$-dimensional representation
at each position 
to a linear classifier 
with the softmax loss
to predict the segmentation maps.
The segmentation maps are upsampled ($4$ times) 
to the input size by bilinear upsampling
for both training and testing.
We report the results over 
two scene parsing datasets, PASCAL-Context~\cite{MottaghiCLCLFUY14} and
Cityscapes~\cite{CordtsORREBFRS16},
and a human parsing dataset, LIP~\cite{GongLSL17}. The mean of class-wise intersection over union (mIoU) 
is adopted as the evaluation metric.

\begin{figure*}[t]
	\centering
	\includegraphics[height = 0.11\textwidth]{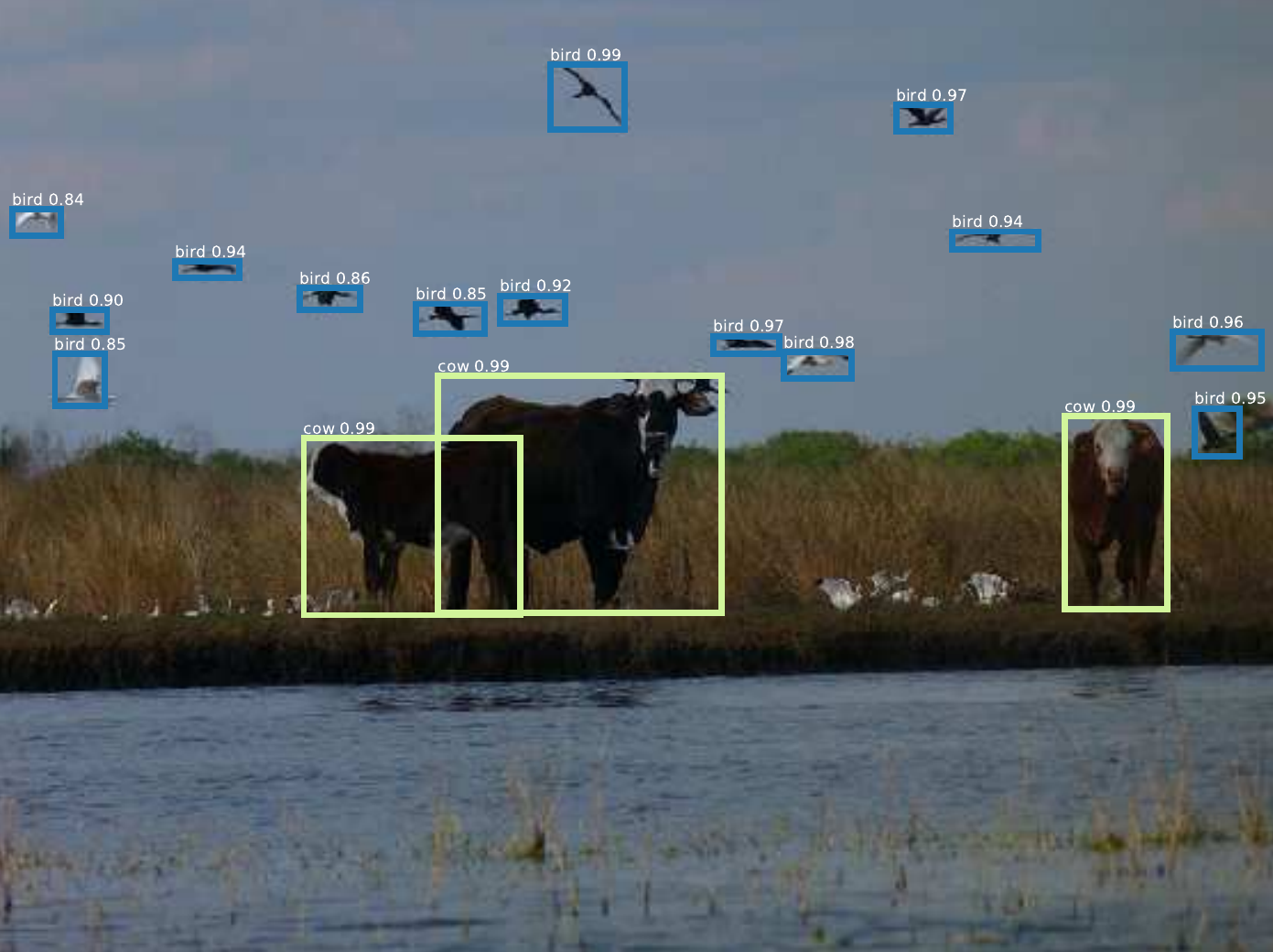}
	\includegraphics[height = 0.11\textwidth]{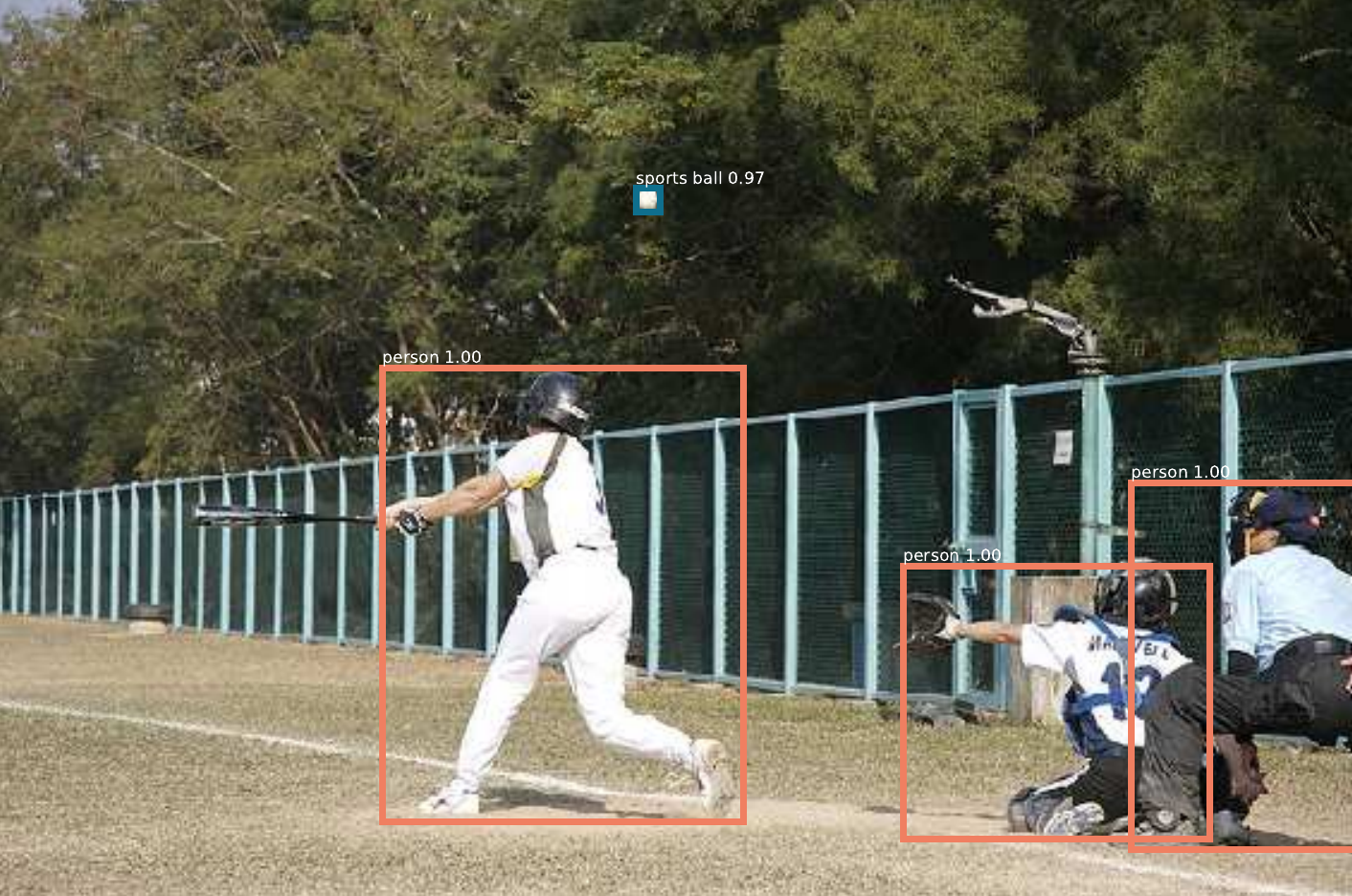}
	\includegraphics[height = 0.11\textwidth]{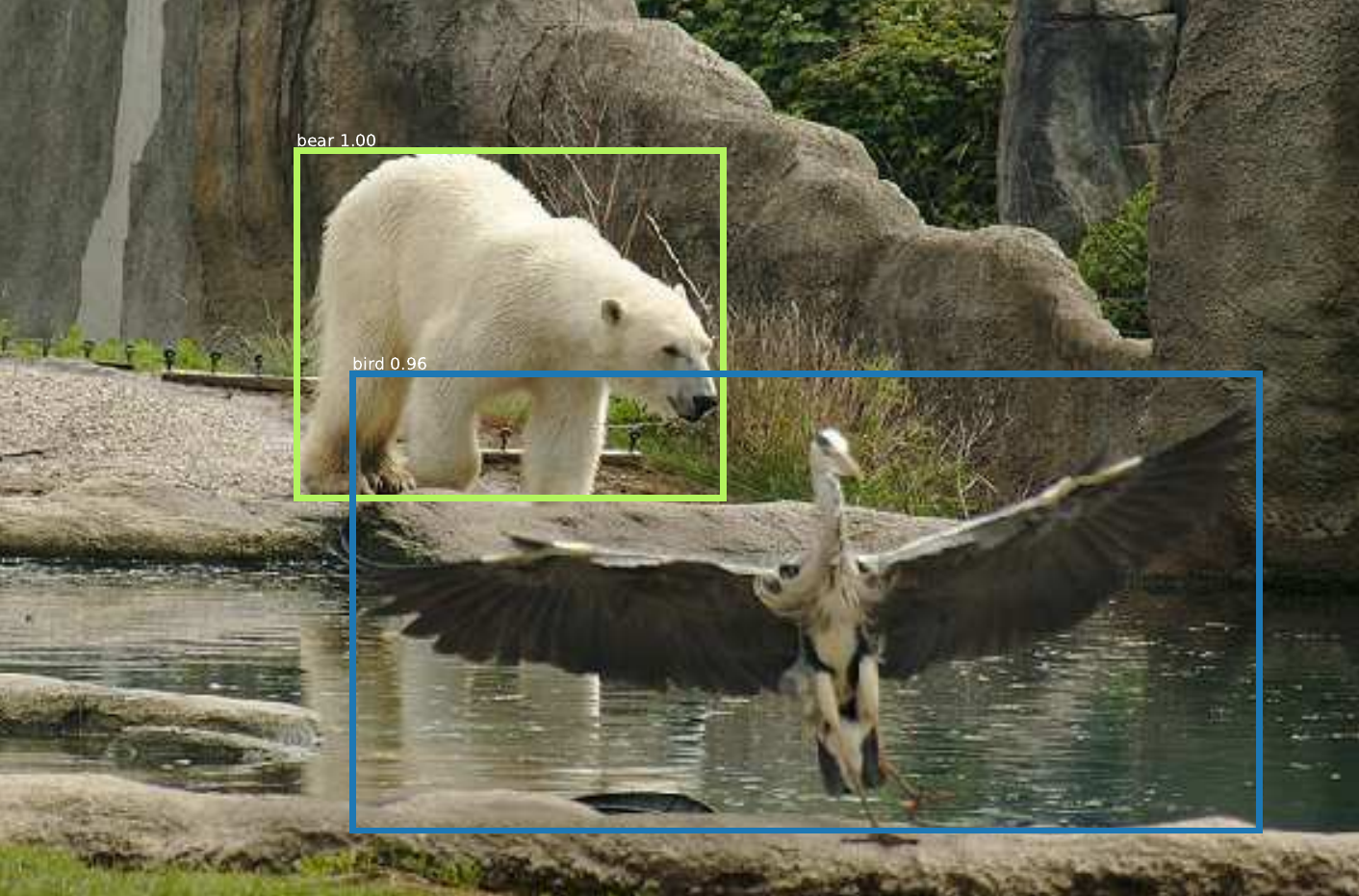}
	\includegraphics[height = 0.11\textwidth]{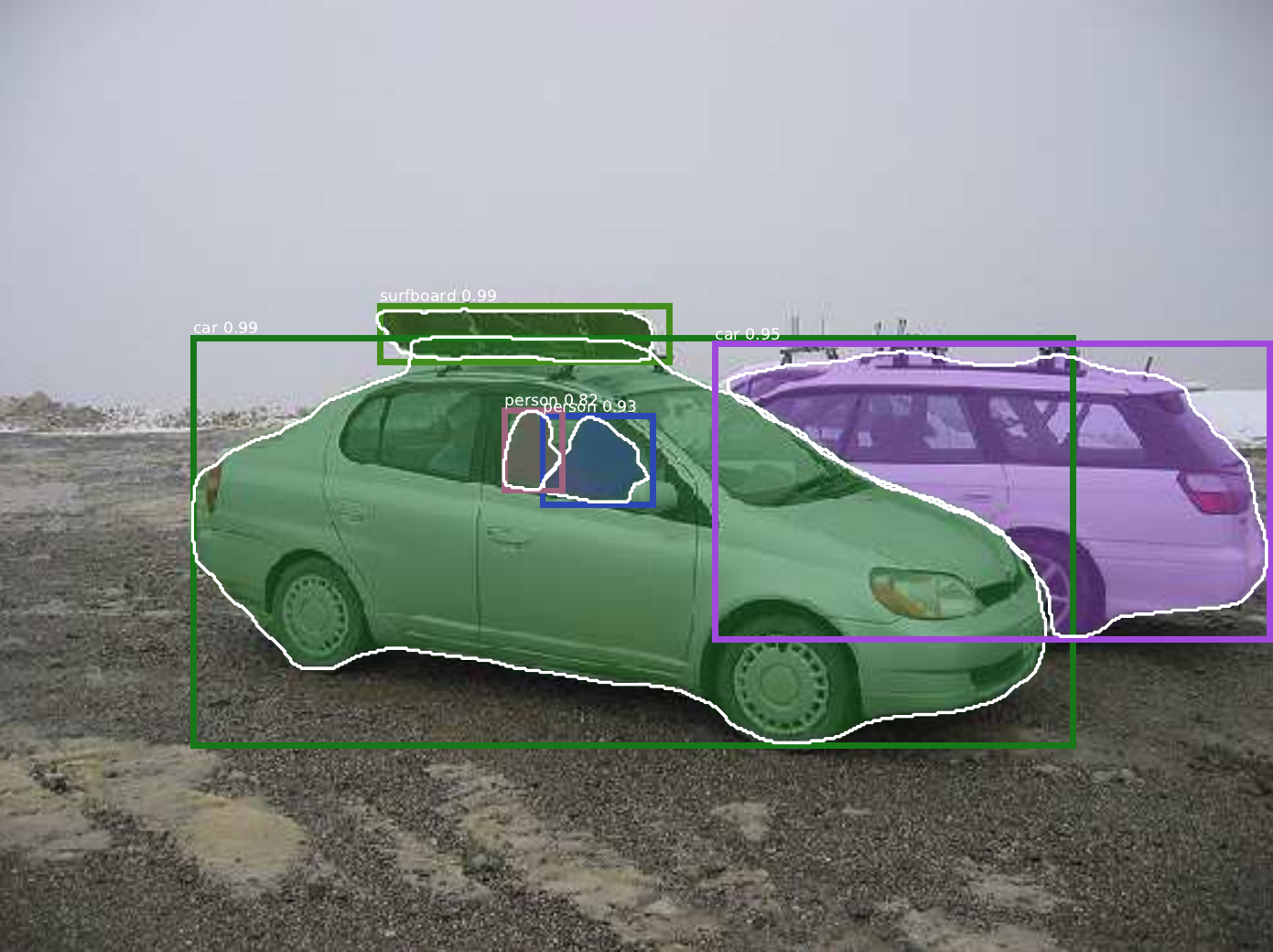}
	\includegraphics[height = 0.11\textwidth]{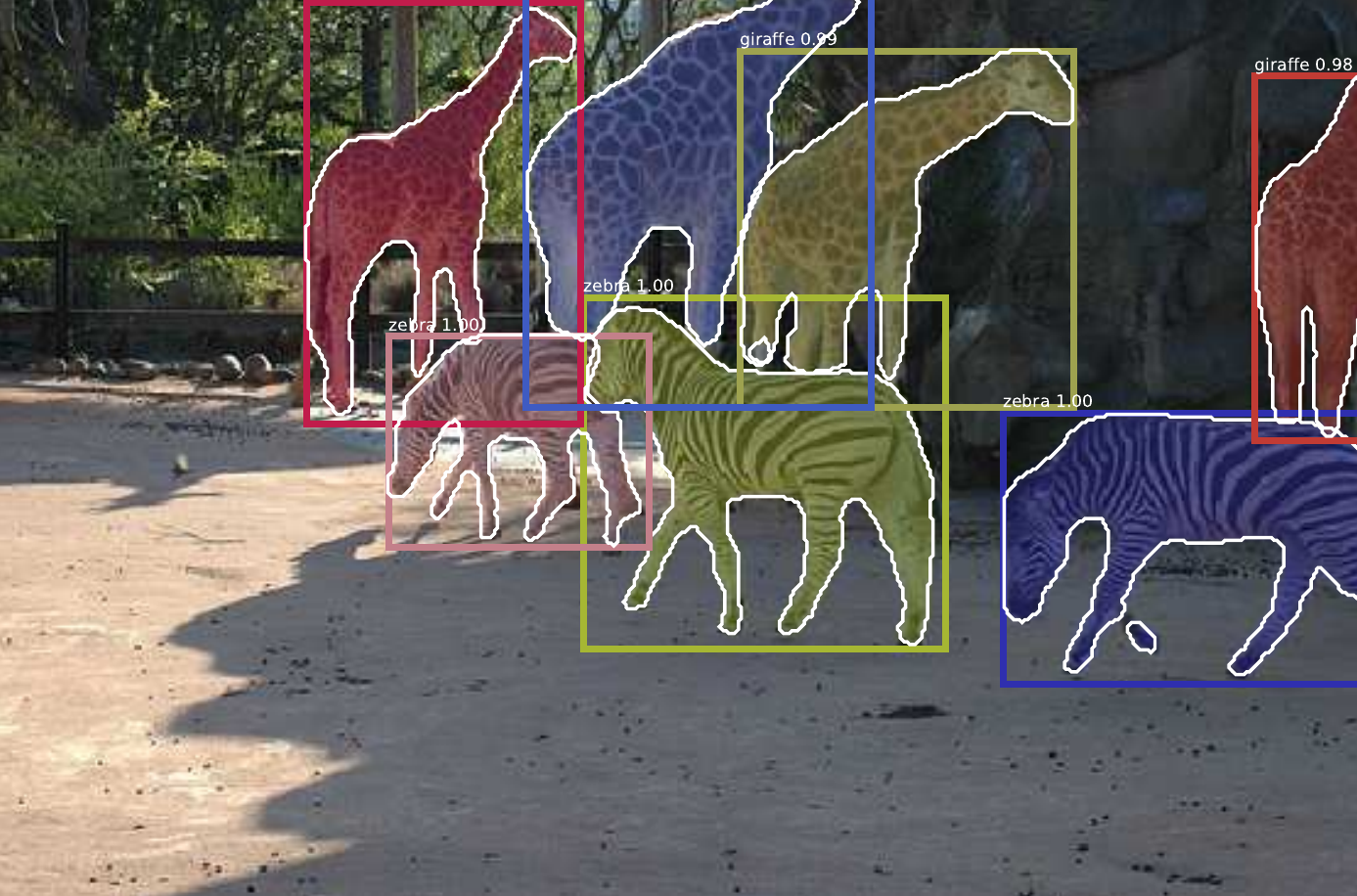}
	\includegraphics[height = 0.11\textwidth]{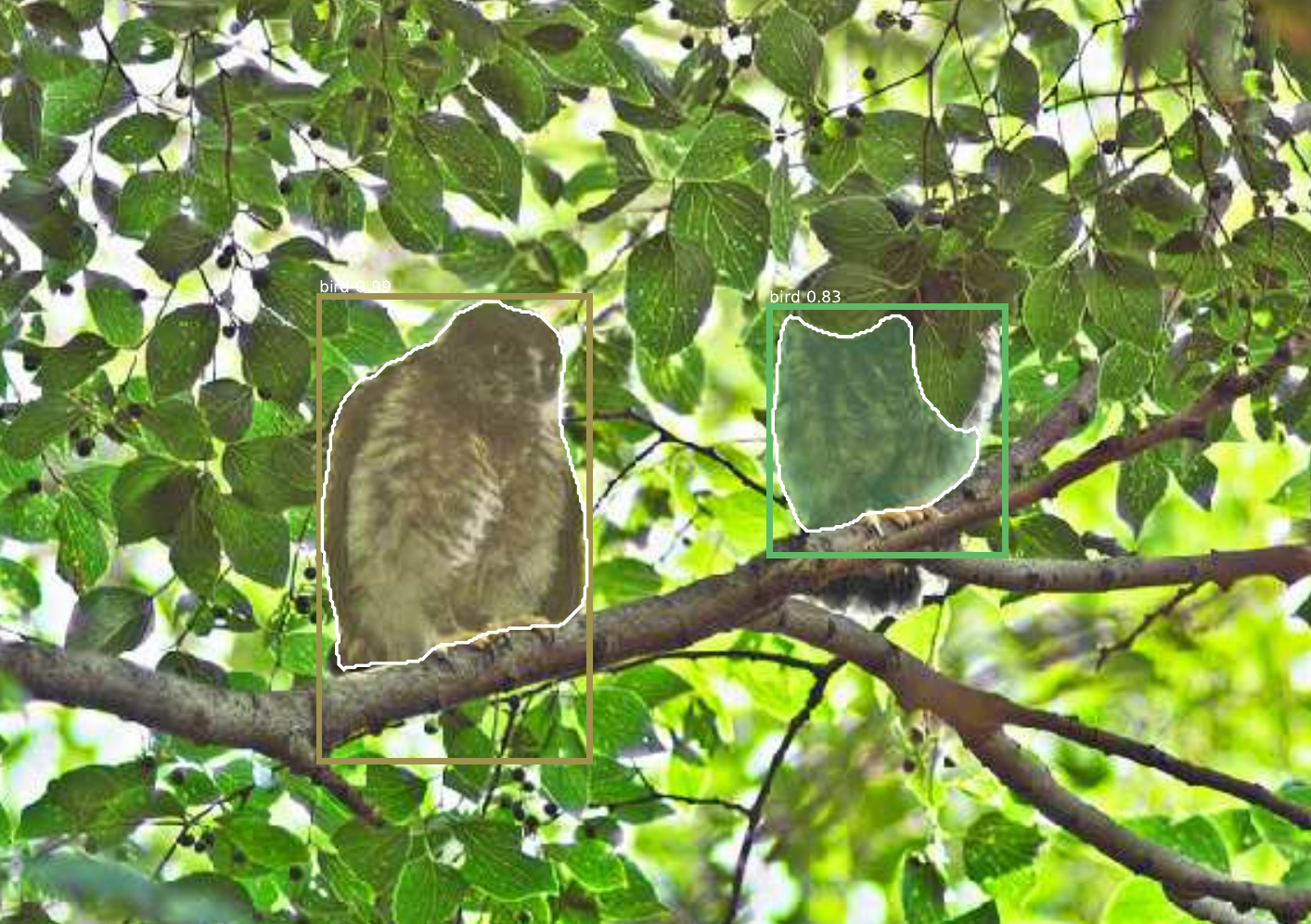}
	\vspace{-.3cm}
	\caption{Qualitative examples for COCO object detection (left three) and instance segmentation (right three).}
	\label{fig:qualitativedetectionresults}
	\vspace{-.2cm}
\end{figure*}

\renewcommand{\arraystretch}{1.3}
	\begin{table*}[t]
	    \centering\setlength{\tabcolsep}{2pt}
	    \scriptsize
	    \caption{GFLOPs and \#parameters
	    for
	    COCO object detection. 
	    The numbers are obtained
	    with the input size $800 \times 1200$
	    and
	    if applicable
	    $512$ proposals fed into R-CNN
	    except the numbers for CenterNet 
	    are obtained with the input size $511\times511$.
	    R-$x$ = ResNet-$x$-FPN,  
	    X-$101$ = ResNeXt-$101$-$64$$\times$$4$d,
	    H-$x$ = HRNetV2p-W$x$,
	    and HG-$52$ = Hourglass-$52$.
	    }
	    \label{tab:det_GLOPS_Parameter_comparision}
	    \vspace{-.3cm}
	    \begin{tabular}{l|rr|rr|rr|rr|rr|rr|rr|rr|rr|rr}
	        \hline %
	        & \multicolumn{6}{c|}{Faster R-CNN~\cite{HeGDG17}} & \multicolumn{6}{c|}{Cascade R-CNN~\cite{CaiV19}} & \multicolumn{4}{c|}{FCOS~\cite{TianSCH19}} & \multicolumn{4}{c}{CenterNet~\cite{DuanBXQHT19}} \\
	        \hline
	        ~ & R-$50$ & H-$18$& R-$101$ &H-$32$ & X-$101$ & H-$48$ & R-$50$ & H-$18$& R-$101$ &H-$32$ & X-$101$ & H-$48$ & R-$50$ & H-$18$& R-$101$ &H-$32$ & HG-$52$ & H-$48$ & HG-$104$ & H-$64$ \\
	        \hline
	        
	        \hline
	        \#param. (M) & $39.8$ & $26.2$ & $57.8$ & $45.0$ &  $94.9$ & $79.4$ & $69.4$ & $55.1$ & $88.4$ & $74.9$ & $127.3$ & $111.0$ & $32.0$ & $17.5$ & $51.0$ & $37.3$  & $104.8$ & $73.6$ & $210.1$ & $127.7$\\
	        GFLOPs & $172.3$ & $159.1$ & $239.4$ & $245.3$ & $381.8$ &  $399.1$ & $226.2$ & $ 207.8$ & $298.7$ & $ 300.8$ & $448.3$ & $466.5$ & $190.0$ & $180.3$ & $261.2$ & $273.3$ & $227.0$ & $217.1$ & $388.4$ & $318.5$ \\
	        \hline 
	        
	        & \multicolumn{6}{c|}{Cascade Mask R-CNN~\cite{CaiV19}} & \multicolumn{6}{c|}{Hybrid Task Cascade~\cite{ChenPWXLSFLSOLL19}}  & \multicolumn{4}{c|}{Mask R-CNN~\cite{HeGDG17}} \\
	        \cline{1-17}
	        ~ & R-$50$ & H-$18$& R-$101$ &H-$32$ & X-$101$ & H-$48$ & R-$50$ & H-$18$& R-$101$ &H-$32$ & X-$101$ & H-$48$ & R-$50$ & H-$18$& R-$101$ &H-$32$\\
	        \cline{1-17}
	        
	        \#param. (M) & $77.3$ & $63.1$ & $96.3$ & $82.9$ & $135.2$ & $118.9$ & $80.3$ & $66.1$ & $99.3$ & $85.9 $ &  $138.2$ & $121.9$ & $44.4$ & $30.1$ & $63.4$ & $49.9$ \\
            GFLOPs & $431.7$ & $413.1$ & $504.1$ & $506.2$ & $653.7$ & $671.9$ & $476.9$ & $458.3$ & $549.2$ & $551.4$ & $698.9$ & $717.0$ & $266.5$ & $247.9$ & $338.8$ & $341.0$ \\
	        \cline{1-17}
	    \end{tabular} 
	    \vspace{-.2cm}
	\end{table*}

\renewcommand{\arraystretch}{1.1}
	\begin{table}[t]
	\setlength{\tabcolsep}{4.4pt}
	\scriptsize
	\centering
	\caption{Object detection results on COCO \texttt{val}
	in the Faster R-CNN and Cascade R-CNN frameworks.
	LS = learning schedule.
	{$1\times$ = $12e$,
	$2\times$ = $24e$.
	Our approach performs better than
	ResNet and ResNeXt.
	Our approach gets more significant improvement
	for $2\times$ than $1\times$
	and for small objects (AP$_S$)
	than medium (AP$_M$) and large objects
	(AP$_L$).}}
	\label{tab:object_detection_fpn}
	\vspace{-.3cm}
	\begin{tabular}{l|c|ccc|ccc}
		\hline %
		backbone & LS & AP & AP$_{50}$ & AP$_{75}$ & AP$_S$ & AP$_M$ & AP$_L$\\
		\hline
		\multicolumn{8}{c}{Faster R-CNN~\cite{LinDGHHB17}}\\
        \hline
	  	ResNet-$50$-FPN  & $1\times$  & $36.7$ & $58.3$ & $39.9$ & $20.9$ & $39.8$ & $47.9$ \\
		HRNetV$2$p-W$18$ & $1\times$ & $36.2$ & $57.3$ & $39.3$ & $20.7$ & $39.0$ & $46.8$ \\
	    ResNet-$50$-FPN & $2\times$  & $37.6$ & $58.7$ & $41.3$ & $21.4$ & ${40.8}$ & ${49.7}$ \\
		HRNetV$2$p-W$18$ & $2\times$ & ${38.0}$ & ${58.9}$ & ${41.5}$ & ${22.6}$ & ${40.8}$ & $49.6$ \\
	    \hline
	    ResNet-$101$-FPN & $1\times$ & $39.2$ & $61.1$ & $43.0$ & $22.3$ & $42.9$ & $50.9$ \\
		HRNetV$2$p-W$32$ & $1\times$ & $39.6$ & $61.0$ & $43.3$ & $23.7$ & $42.5$ & $50.5$ \\
		ResNet-$101$-FPN & $2\times$ & $39.8$ & $61.4$ & $43.4$ & $22.9$ & $43.6$ & $52.4$ \\
	    HRNetV$2$p-W$32$ & $2\times$ & ${40.9}$ & ${61.8}$ & ${44.8}$ & ${24.4}$ & ${43.7}$ & ${53.3}$  \\
	    \hline
	    
	    X-$101$-$64$$\times$$4$d-FPN & $1\times$ & $41.3$ & ${63.4}$ & $45.2$ & $24.5$ & ${45.8}$ & $53.3$ \\
	    HRNetV$2$p-W$48$ & $1\times$ & $41.3$ & $62.8$ & $45.1$ & ${25.1}$ & $44.5$ & $52.9$  \\
	    X-$101$-$64$$\times$$4$d-FPN & $2\times$ & $40.8$ & $62.1$ & $44.6$ & $23.2$ & $44.5$ & $53.7$ \\
	    HRNetV$2$p-W$48$ & $2\times$ & ${41.8}$ & $62.8$ & ${45.9}$ & $25.0$ & $44.7$ & ${54.6}$  \\

	    \hline
	    
	    \multicolumn{8}{c}{Cascade R-CNN~\cite{CaiV19}}\\
	    \hline
	    ResNet-$50$-FPN & $20e$ & $41.1$ & $59.1$ & $44.8$ & $22.5$ & ${44.4}$ & ${54.9}$ \\ 
	    HRNetV$2$p-W$18$ & $20e$ & ${41.3}$ & ${59.2}$ & ${44.9}$ & ${23.7}$ & $44.2$ & $54.1$ \\ 
	    \hline
	    ResNet-$101$-FPN & $20e$ & $42.5$ & $60.7$ & $46.3$ & $23.7$ & $46.1$ & $56.9$ \\ 
	    HRNetV$2$p-W$32$ & $20e$ & ${43.7}$ & ${61.7}$ & ${47.7}$ & ${25.6}$ & ${46.5}$ & ${57.4}$ \\ 
	    \hline
	     X-$101$-$64$$\times$$4$d-FPN & $20e$ & ${44.7}$ & ${63.1}$ & ${49.0}$ & $25.8$ & ${48.3}$ & ${58.8}$ \\
	    HRNetV$2$p-W$48$ & $20e$ & $44.6$ & $62.7$ & $48.7$ & ${26.3}$ & $48.1$ & $58.5$  \\
	    \hline
	\end{tabular} 
	\vspace{-.2cm}
	\end{table}

\renewcommand{\arraystretch}{1.1}
	\begin{table}[t]
	\setlength{\tabcolsep}{4.4pt}
	\scriptsize
	\centering
	\caption{Object detection results on COCO \texttt{val}
	in the FCOS and CenterNet frameworks. 
	The results are obtained using
	the implementations provided by the authors.
	{
	Our approach performs superiorly to ResNet and Hourglass
	for similar parameter and computation complexity.
	Our HRNetV$2$p-W$64$
	performs slightly worse than Hourglass-$104$,
	and
	the reason is that 
	Hourglass-$104$ is much more heavier than
	HRNetV$2$p-W$64$.}
	See Table~\ref{tab:det_GLOPS_Parameter_comparision}
	for \#parameters and GFLOPs.}
	\label{tab:object_detection_anchorfree}
	\vspace{-.3cm}
	\begin{tabular}{l|c|ccc|ccc}
		\hline %
		backbone & LS & AP & AP$_{50}$ & AP$_{75}$ & AP$_S$ & AP$_M$ & AP$_L$\\
		\hline 
		\multicolumn{8}{c}{FCOS~\cite{TianSCH19}}\\
		\hline
		ResNet-$50$-FPN  & $2\times$  & $37.1$ & ${55.9}$ & $39.8$ & $21.3$ & ${41.0}$ & $47.8$ \\
		HRNetV$2$p-W$18$ & $2\times$  & ${37.7}$ & $55.3$ & ${40.2}$ & ${22.0}$ & $40.8$ & ${48.8}$ \\
		\hline
		ResNet-$101$-FPN & $2\times$  & $41.4$ & ${60.3}$ & $44.8$ & $25.0$ & ${45.6}$ & $53.1$ \\
		HRNetV$2$p-W$32$ & $2\times$  & ${41.9}$ & ${60.3}$ & ${45.0}$ & ${25.1}$ & ${45.6}$ & ${53.2}$ \\
		
	    \hline

	    \multicolumn{8}{c}{CenterNet~\cite{DuanBXQHT19}}\\
	    \hline
		Hourglass-$52$ & -  & $41.3$ & $59.2$ & $43.9$ & $23.6$ & $43.8$ & $55.8$ \\
		HRNetV$2$p-W$48$ & -  & ${43.4}$ & ${61.8}$ & ${45.6}$ & ${23.8}$ &${47.1}$ & ${59.3}$ \\
		\hline
		Hourglass-$104$ & -  & ${44.8}$ & $62.4$ & ${48.2}$ & ${25.9}$ & ${48.9}$ & $58.8$ \\
		HRNetV$2$p-W$64$ & -  & $44.0$ & ${62.5}$ & $47.3$ & $23.9$ &$48.2$ & ${60.2}$ \\
		\hline
	\end{tabular} 
	\vspace{-.2cm}
	\end{table}

\renewcommand{\arraystretch}{1.3}
	\begin{table}[t]
	\setlength{\tabcolsep}{2.3pt}
	\centering
	\caption{
	Object detection results on COCO \texttt{val}
	in the Mask R-CNN and its extended frameworks.
	{
	The overall performance 
	of our approach is superior to 
	ResNet except that
	HRNetV$2$p-W$18$ sometimes performs worse than ResNet-$50$.
	Similar to detection (bbox), 
	the improvement for small objects (AP$_S$) in terms of mask
	is also more significant than medium
	(AP$_M$)
	and large objects (AP$_L$).}
	The results are obtained from MMDetection~\cite{mmdetection}.}
	\label{tab:object_detection_maskrcnn}
	\scriptsize
	\vspace{-.3cm}
	\begin{tabular}{l|c|cccc|cccc}
		\hline %
		\multirow{2}{*}{backbone} & 
		\multirow{2}{*}{LS} & \multicolumn{4}{c|}{mask} & \multicolumn{4}{c}{bbox} \\
		\cline{3-10}&  & AP & AP$_S$ & AP$_M$ & AP$_L$ & AP & AP$_S$ & AP$_M$ & AP$_L$ \\
		\hline
		\multicolumn{10}{c}{Mask R-CNN~\cite{HeGDG17}}\\
		\hline

		\hline
	    ResNet-$50$-FPN & $1\times$  & $34.2$ & $15.7$ & $36.8$ & $50.2$ & $37.8$ & $22.1$ & $40.9$ & $49.3$  \\
		HRNetV$2$p-W$18$ & $1\times$ & $33.8$ & $15.6$ & $35.6$ & $49.8$ & $37.1$ & $21.9$ & $39.5$ & $47.9$  \\
	    ResNet-$50$-FPN & $2\times$  & $35.0$ & $16.0$ & ${37.5}$ & ${52.0}$ & $38.6$ & $21.7$ & $41.6$ & $50.9$  \\
		HRNetV$2$p-W$18$ & $2\times$ & ${35.3}$ & ${16.9}$ & ${37.5}$ & $51.8$ & ${39.2}$ & ${23.7}$ & ${41.7}$ & ${51.0}$  \\
	    \hline
		ResNet-$101$-FPN & $1\times$ & $36.1$ & $16.2$ & $39.0$ & $53.0$ & $40.0$ & $22.6$ & $43.4$ & $52.3$  \\
		HRNetV$2$p-W$32$ & $1\times$ & $36.7$ & $17.3$ & $39.0$ & $53.0$ & $40.9$ & $24.5$ & $43.9$ & $52.2$  \\
		ResNet-$101$-FPN & $2\times$ & $36.7$ & $17.0$ & $39.5$ & $54.8$ & $41.0$ & $23.4$ & $44.4$ & $53.9$  \\
		HRNetV$2$p-W$32$ & $2\times$ & ${37.6}$ & ${17.8}$ & ${40.0}$ & ${55.0}$ & ${42.3}$ & ${25.0}$ & ${45.4}$ & ${54.9}$ \\
		\hline
		\multicolumn{10}{c}{Cascade Mask R-CNN~\cite{CaiV19}}\\
		\hline
	    ResNet-$50$-FPN & $20e$  & ${36.6}$ & ${19.0}$ & $37.4$ & $50.7$ & ${42.3}$ & $23.7$ & ${45.7}$ & ${56.4}$  \\
		HRNetV$2$p-W$18$ & $20e$ & $36.4$ & $17.0$ & ${38.6}$ & ${52.9}$ & $41.9$ & ${23.8}$ & $44.9$ & $55.0$  \\
		\hline
	    ResNet-$101$-FPN & $20e$  & $37.6$ & ${19.7}$ & $40.8$ & $52.4$ & $43.3$ & $24.4$ & $46.9$ & $58.0$  \\
	    
		HRNetV$2$p-W$32$ & $20e$ & ${38.5}$ & $18.9$ & ${41.1}$ & ${56.1}$ & ${44.5}$ & ${26.1}$ & ${47.9}$ & ${58.5}$  \\
		\hline
	    X-$101$-$64$$\times$$4$d-FPN & $20e$  & $39.4$ & ${20.8}$ & ${42.7}$ & $54.1$ & $45.7$ & $26.2$ & ${49.6}$ & $60.0$  \\
	    
		HRNetV$2$p-W$48$ & $20e$ & ${39.5}$ & $19.7$ & $41.8$ & ${56.9}$ & ${46.0}$ & ${27.5}$ & $48.9$ & ${60.1}$  \\

		\hline
		\multicolumn{10}{c}{Hybrid Task Cascade~\cite{ChenPWXLSFLSOLL19}}\\
		\hline
		
	    ResNet-$50$-FPN & $20e$  & ${38.1}$ & ${20.3}$ & ${41.1}$ & $52.8$ & ${43.2}$ & $24.9$ & ${46.4}$ & ${57.8}$  \\
	    
		HRNetV$2$p-W$18$ & $20e$ & $37.9$ & $18.8$ & $39.9$ & ${55.2}$ & $43.1$ & ${26.6}$ & $46.0$ & $56.9$  \\
		\hline
	    ResNet-$101$-FPN & $20e$  & $39.4$ & ${21.4}$ & ${42.4}$ & $54.4$ & $44.9$ & $26.4$ & $48.3$ & ${59.9}$  \\
	    
		HRNetV$2$p-W$32$ & $20e$ & ${39.6}$ & $19.1$ & $42.0$ & ${57.9}$ & ${45.3}$ & ${27.0}$ & ${48.4}$ & $59.5$  \\
	    \hline
	    X-$101$-$64$$\times$$4$d-FPN & $20e$ & ${40.8}$ & ${22.7}$ & ${44.2}$ & $56.3$ & ${46.9}$ & ${28.0}$ & ${50.7}$ & ${62.1}$ \\
	    HRNetV$2$p-W$48$ & $20e$ & $40.7$ & $19.7$ & $43.4$ & $59.3$ & $46.8$ & ${28.0}$ & $50.2$ & $61.7$  \\
	    \hline
	    X-$101$-$64$$\times$$4$d-FPN & $28e$ & $40.7$ & $20.0$ & ${44.1}$ & ${59.9}$ & $46.8$ & $27.5$ & ${51.0}$ & $61.7$ \\
	    HRNetV$2$p-W$48$ & $28e$ & ${41.0}$ & ${20.8}$ & $43.9$ & ${59.9}$ & ${47.0}$ & ${28.8}$ & $50.3$ & ${62.2}$  \\
		
		\hline
	\end{tabular}
	\vspace{-0.2cm}
	\end{table}

\vspace{.1cm}
\noindent\textbf{Cityscapes.}
The Cityscapes dataset~\cite{CordtsORREBFRS16} contains $5,000$ high quality pixel-level finely annotated scene images.
The finely-annotated images are divided into $2,975/500/1,525$ images for training, validation and testing. 
There are $30$ classes, and $19$ classes among them are used for evaluation. 
In addition to the mean of class-wise intersection over union (mIoU), 
we report other three scores on the test set: 
IoU category (cat.), iIoU class (cla.) and iIoU category (cat.). 

We follow the same training protocol~\cite{ZhaoSQWJ17, ZhaoZLSLLJ18}.
The data are augmented by random cropping (from $1024\times2048$ to $512\times1024$), random scaling in the range of $[0.5, 2]$, and random horizontal flipping. We use the SGD optimizer with the base learning rate of $0.01$, 
the momentum of $0.9$ and the weight decay of $0.0005$. The poly learning rate policy with the power of $0.9$ is used for dropping the learning rate. All the models are trained for $120K$ iterations with the batch size of $12$ on $4$ GPUs and syncBN.

Table~\ref{tab:cityscapevalresults} provides the comparison with several representative methods 
on the Cityscapes \texttt{val} set
in terms of parameter and computation complexity and mIoU class. 
(i) HRNetV$2$-W$40$ ($40$ indicates the width of the high-resolution convolution), with similar model size to DeepLabv$3$+
and much lower computation complexity, gets better performance:
$4.7$ points gain over UNet++, $1.7$ points gain over DeepLabv3
and about $0.5$ points gain over PSPNet, DeepLabv3+. 
(ii) HRNetV$2$-W$48$, with similar model size to PSPNet and much lower computation complexity, achieves much significant improvement: 
$5.6$ points gain over UNet++, $2.6$ points gain over DeepLabv3
and about $1.4$ points gain over PSPNet, DeepLabv3+. 
In the following comparisons,
we adopt HRNetV$2$-W$48$ 
that is pretrained on ImageNet
and has similar model size 
as most Dilated-ResNet-$101$ based methods. 

Table~\ref{tab:cityscaperesults} provides the comparison of our method 
with state-of-the-art methods on the Cityscapes \texttt{test} set.
All the results are with six scales and flipping.
Two cases w/o using coarse data are evaluated:
One is about the model learned
on the \texttt{train} set,
and the other is about the model
learned on the \texttt{train+val} set.
In both cases, 
HRNetV$2$-W$48$ achieves the superior performance. %

\vspace{.1cm}
\noindent\textbf{PASCAL-Context.}
The PASCAL-Context dataset~\cite{MottaghiCLCLFUY14} 
includes $4,998$ scene images for training and 
$5,105$ images for testing with $59$ semantic labels and $1$ background label. 

The data augmentation and learning rate policy are the same as Cityscapes. 
Following the widely-used training strategy~\cite{0005DSZWTA18, DingJSL018}, 
we resize the images to $480\times480$ and set the initial learning rate to $0.004$ 
and weight decay to $0.0001$. The batch size is $16$ and the number of iterations is $60K$.

We follow the standard testing procedure~\cite{0005DSZWTA18, DingJSL018}. 
The image is resized to $480\times480$ 
and then fed into our network.
The resulting $480\times480$ label maps are then resized to the 
original image size. 
We evaluate the performance of our approach and other approaches
using six scales and flipping.

Table~\ref{tab:pasctxresults}
provides the comparison of our method
with state-of-the-art methods. 
There are two kinds of evaluation schemes:
mIoU over $59$ classes and $60$ classes ($59$ classes + background).
In both cases, 
HRNetV$2$-W$48$ achieves state-of-the-art results
except that the result from~\cite{HeDZWQ} is higher than ours without using the OCR scheme~\cite{YuanCW18}.

\vspace{.1cm}
\noindent\textbf{LIP.}
The LIP dataset \cite{GongLSL17} 
contains $50,462$ elaborately annotated human images,
which are divided into $30,462$ training images,
and $10,000$ validation images. The methods are evaluated on 
$20$ categories ($19$ human part labels and 
$1$ background label). 
Following the standard training and testing settings~\cite{TL18}, the images are resized to $473\times473$ and
the performance is evaluated
on the average of the segmentation maps of the original and flipped images.

The data augmentation and learning rate policy are the same as Cityscapes. 
The training strategy follows the recent setting~\cite{TL18}. 
We set the initial learning rate to $0.007$ and 
the momentum to $0.9$ and the weight decay to $0.0005$.
The batch size is $40$ and the number of iterations is $110$K. 

Table~\ref{tab:lipresults}
provides the comparison of our method
with state-of-the-art methods. 
The overall performance of HRNetV$2$-W$48$
performs the best 
with fewer parameters and lighter computation cost.
We also would like to mention that
our networks do not use extra information such as pose or edge.

\renewcommand{\arraystretch}{1.3}
    \begin{table*}[ht]
	\setlength{\tabcolsep}{12pt}
	\centering
	\caption{Comparison with the state-of-the-art single-model object detectors on COCO \texttt{test-dev} with BN parameters fixed and without mutli-scale training and testing.
	$^*$ means that the result
	is from the original paper~\cite{CaiV18}.
	GFLOPs and \#parameters of the models 
	are given in Table~\ref{tab:det_GLOPS_Parameter_comparision}.
	{
	The observations are similar to those 
	on COCO \texttt{val},
	and show that 
	the HRNet performs better than ResNet and ResNeXt
	under state-of-the-art object detection 
	and instance segmentation frameworks.}}
	\scriptsize
	\label{tab:recent_object_detection_results_single}
	\vspace{-3mm}
	\begin{tabular}{l|lcr|ccc|ccc}%
    	\hline %
    	 & backbone & size & LS & AP & AP$_{50}$ & AP$_{75}$ & AP$_S$ & AP$_M$ & AP$_L$\\

		\hline
		
		\hline
		MLKP \cite{WangWGLZ18} & VGG$16$ & - & - &  $28.6$ & $52.4$ & $31.6$ & $10.8$ & $33.4$ & $45.1$ \\   
		
		STDN \cite{ZhouNGHX18} & DenseNet-$169$ & $513$ & - & $31.8$ & $51.0$ & $33.6$ & $14.4$ & $36.1$ & $43.4$\\
		
		DES \cite{ZhangQX0WY18} & VGG$16$ & $512$  & - & $32.8$ & $53.2$ & $34.6$ & $13.9$ & $36.0$ & $47.6$ \\ 
		CoupleNet \cite{ZhuZWZWL17} & ResNet-$101$ & - & - & $33.1$ & $53.5$ & $35.4$ & $11.6$ & $36.3$ & $50.1$ \\
		DeNet \cite{Tychsen-SmithP17} & ResNet-$101$ & $512$ & - & $33.8$ & $53.4$ & $36.1$ & $12.3$ & $36.1$ & $50.8$ \\
	    RFBNet \cite{LiuHW18} & VGG$16$ & $512$ & -  & $34.4$ & $55.7$ &  $36.4$ & $17.6$ & $37.0$ & $47.6$ \\    
		DFPR \cite{KongSHL18} & ResNet-$101$ & $512$ & $1\times$ & $34.6$ & $54.3$ & $37.3$ & - & - & - \\

	    PFPNet \cite{KimKSKK18} & VGG$16$ & $512$ & - & $35.2$ & $57.6$ &  $37.9$ & $18.7$ & $38.6$ & $45.9$ \\   
	    
	    RefineDet\cite{ZhangWBLL18} & ResNet-$101$ & $512$ & - & $36.4$ & $57.5$ & $39.5$ & $16.6$ & $39.9$ & $51.4$  \\  

	    Relation Net \cite{HuGZDW18} & ResNet-$101$ & $600$ & - & $39.0$ & $58.6$ & $42.9$ & - & - & - \\  
	    C-FRCNN \cite{ChenHT18} & ResNet-$101$ & $800$ & $1\times$ & $39.0$ & $59.7$ & $42.8$ & $19.4$ & $42.4$ & $53.0$ \\
	    RetinaNet \cite{LinGGHD17} & ResNet-$101$-FPN & $800$ & $1.5\times$ & $39.1$ & $59.1$ & $42.3$ & $21.8$ & $42.7$ & $50.2$ \\
        
        Deep Regionlets \cite{XuLWRBC18} & ResNet-$101$ & $800$ & $1.5\times$  & $39.3$ & $59.8$ &  - & $21.7$ & $43.7$ & $50.9$ \\
		
		FitnessNMS \cite{Tychsen-SmithP18} & ResNet-$101$ & $768$ & - & $39.5$ & $58.0$ & $42.6$ & $18.9$ & $43.5$ & $54.1$ \\
        DetNet \cite{LiPYZDS18} & DetNet$59$-FPN & $800$ & $2\times$ & $40.3$ & $62.1$ & $43.8$ & $23.6$ & $42.6$ & $50.0$ \\		
	    CornerNet \cite{LawD18} & Hourglass-$104$ & $511$ & - & $40.5$ & $56.5$ & $43.1$ & $19.4$ & $42.7$ & $53.9$ \\
	    M2Det \cite{M2DETQ} & VGG$16$ & $800$ & $\sim 10\times$ & $41.0$ & $59.7$ & $45.0$ & $22.1$ & ${46.5}$ & $53.8$ \\
	    
	    \hline
        Faster R-CNN \cite{LinDGHHB17} & ResNet-$101$-FPN & $800$ & $1\times$ & $39.3$ & $61.3$ & $42.7$ & $22.1$ & $42.1$ & $49.7$ \\
        Faster R-CNN & HRNetV$2$p-W$32$ & $800$ & $1\times$ & $39.5$ & $61.2$ & $43.0$ & $23.3$ & $41.7$ & $49.1$  \\
        \arrayrulecolor{gray}\hline\arrayrulecolor{black}
		Faster R-CNN \cite{LinDGHHB17} & ResNet-$101$-FPN & $800$ & $2\times$ & $40.3$ & $61.8$ & $43.9$ & $22.6$ & $43.1$ & $51.0$ \\
		Faster R-CNN & HRNetV$2$p-W$32$ & $800$ & $2\times$  & $41.1$ & $62.3$ & $44.9$ & $24.0$ & $43.1$ & $51.4$ \\	
		\arrayrulecolor{gray}\hline\arrayrulecolor{black}
		Faster R-CNN \cite{LinDGHHB17} & ResNet-$152$-FPN & $800$ & $2\times$ & $40.6$ & $62.1$ & $44.3$ & $22.6$ & $43.4$ & $52.0$ \\
		Faster R-CNN & HRNetV$2$p-W$40$ & $800$ & $2\times$  & $42.1$ & $63.2$ & $46.1$ & $24.6$ & $44.5$ & $52.6$ \\
		\arrayrulecolor{gray}\hline\arrayrulecolor{black}
	    Faster R-CNN \cite{mmdetection} & X-$101$-$64$$\times4$d-FPN & $800$ & $2\times$ & $41.1$ & $62.8$ & $44.8$ & $23.5$ & $44.1$ & $52.3$ \\
		Faster R-CNN & HRNetV$2$p-W$48$ &  $800$ & $2\times$ & $42.4$ & ${63.6}$ & $46.4$ & $24.9$ & $44.6$ & $53.0$ \\
		\arrayrulecolor{black}\hline\arrayrulecolor{black}
	    Cascade R-CNN \cite{CaiV18}$^*$ & ResNet-$101$-FPN & $800$ & $\sim 1.6\times$ & $42.8$ & $62.1$ & $46.3$  & $23.7$ & $45.5$ & $55.2$ \\
	    Cascade R-CNN & ResNet-$101$-FPN & $800$ & $\sim 1.6\times$ & $43.1$ & $61.7$ & $46.7$ & $24.1$ & $45.9$ & $55.0$ \\
	    Cascade R-CNN & HRNetV$2$p-W$32$  & $800$ & $\sim 1.6\times$ & $43.7$ & $62.0$ & $47.4$ & $25.5$ & $46.0$ & $55.3$ \\
	    {Cascade R-CNN} & X-$101$-$64\times4$d-FPN  & $800$ & $\sim 1.6\times$ & $44.9$ & $63.7$ & $48.9$ & $25.9$ & $47.7$ & $57.1$ \\
	    {Cascade R-CNN} & HRNetV$2$p-W$48$  & $800$ & $\sim 1.6\times$ & $44.8$ & $63.1$ & $48.6$ & $26.0$ & $47.3$ & $56.3$ \\
	 \arrayrulecolor{black}\hline\arrayrulecolor{black}
	    FCOS~\cite{TianSCH19} & ResNet-$50$-FPN & $800$ & $2\times$ & $37.3$ & $56.4$ & $39.7$ & $20.4$ & $39.6$ & $47.5$ \\
	    FCOS & HRNetV$2$p-W$18$  & $800$ & $2\times$ & $37.8$ & $56.1$ & $40.4$ & $21.6$ & $39.8$ & $47.4$ \\
	    FCOS~\cite{TianSCH19} & ResNet-$101$-FPN & $800$ & $2\times$ & $39.2$ & $58.8$ & $41.6$ & $21.8$ & $41.7$ & $50.0$ \\
	    FCOS & HRNetV$2$p-W$32$  & $800$ & $2\times$ & $40.5$ & $59.3$ & $43.3$ & $23.4$ & $42.6$ & $51.0$ \\
	    \arrayrulecolor{gray}\hline\arrayrulecolor{black}
	    CenterNet\cite{DuanBXQHT19} & Hourglass-$52$ & $511$ & $-$ & $41.6$ & $59.4$ & $44.2$ & $22.5$ & $43.1$ & $54.1$ \\
	    CenterNet & HRNetV$2$-W$48$  & $511$ & $-$ & $43.5$ & $62.1$ & $46.5$ & $22.2$ & ${46.5}$ & ${57.8}$ \\

	    \hline
	    {Cascade Mask R-CNN}~\cite{CaiV19} &  ResNet-$101$-FPN  & $800$ & $\sim 1.6\times$ & $44.0$ & $62.3$ & $47.9$ & $24.3$ & $46.9$ & $56.7$ \\
	    {Cascade Mask R-CNN} & HRNetV$2$p-W$32$  & $800$ & $\sim 1.6\times$ & $44.7$ & $62.5$ & $48.6$ & $25.8$ & $47.1$ & $56.3$ \\	    
	    {Cascade Mask R-CNN}~\cite{CaiV19} &  X-$101$-$64\times4$d-FPN  & $800$ & $\sim 1.6\times$ & $45.9$ & $64.5$ & $50.0$ & $26.6$ & $49.0$ & $58.6$ \\
	    {Cascade Mask R-CNN} & HRNetV$2$p-W$48$  & $800$ & $\sim 1.6\times$ & $46.1$ & $64.0$ & $50.3$ & $27.1$ & $48.6$ & $58.3$ \\	
	    
	    \hline
	    {Hybrid Task Cascade}~\cite{ChenPWXLSFLSOLL19} &  ResNet-$101$-FPN & $800$ & $\sim 1.6\times$ & $45.1$ & $64.3$ & $49.0$ & $25.2$ & $48.0$ & $58.2$ \\
	    {Hybrid Task Cascade} & HRNetV$2$p-W$32$  & $800$ & $\sim 1.6\times$ & $45.6$ & $64.1$ & $49.4$ & $26.7$ & $47.7$ & $58.0$ \\
	    {Hybrid Task Cascade}~\cite{ChenPWXLSFLSOLL19} &  X-$101$-$64\times4$d-FPN & $800$ & $\sim 1.6\times$ & $47.2$ & $66.5$ & $51.4$ & $27.7$ & $50.1$ & $60.3$ \\
	    {Hybrid Task Cascade} & HRNetV$2$p-W$48$  & $800$ & $\sim 1.6\times$ & $47.0$ & $65.8$ & $51.0$ & $27.9$ & $49.4$ & $59.7$ \\
	    {Hybrid Task Cascade}~\cite{ChenPWXLSFLSOLL19} &  X-$101$-$64\times4$d-FPN & $800$ & $\sim 2.3\times$ & $47.2$ & $66.6$ & $51.3$ & $27.5$ & $50.1$ & $60.6$ \\
	    {Hybrid Task Cascade} & HRNetV$2$p-W$48$  & $800$ & $\sim 2.3\times$ & $47.3$ & $65.9$ & $51.2$ & $28.0$ & $49.7$ & $59.8$ \\

		\arrayrulecolor{black}\hline
	\end{tabular}
	\vspace{-2mm}
	\end{table*}

\section{COCO Object Detection}
We perform the evaluation on the MS COCO $2017$ detection dataset,
which contains about $118$k images for training,
$5$k for validation (\texttt{val}) and $\sim20$k testing without provided annotations (\texttt{test-dev}).
The standard COCO-style evaluation is adopted.
Some example results 
by our approach
are given in Figure~\ref{fig:qualitativedetectionresults}.

We apply our multi-level representations (HRNetV$2$p)\footnote{Same as FPN~\cite{LinGGHD17}, 
we also use $5$ levels.},
shown in Figure~\ref{fig:highresolutionneck} (c), 
for object detection.
The data is augmented
by standard horizontal flipping. The input images are resized such that the shorter edge is 800 pixels \cite{LinDGHHB17}.
Inference is performed on a single image scale.

We compare our HRNet with the standard models: 
ResNet~\cite{HeZRS16} and ResNeXt~\cite{XieGDTH17}. 
We evaluate the detection performance
on COCO \texttt{val}.
under two anchor-based frameworks:
Faster R-CNN~\cite{RenHG017} and
Cascade R-CNN~\cite{CaiV18},
and two recently-developed
anchor-free frameworks: FCOS~\cite{TianSCH19}
and CenterNet~\cite{DuanBXQHT19}.
We train the Faster R-CNN 
and Cascade R-CNN models for both our HRNetV$2$p and the ResNet
on the public MMDetection platform~\cite{mmdetection}
with the provided training setup,
except that we use the learning rate schedule suggested in~\cite{DBLP:journals/corr/abs-1811-08883}
for $2\times$,
and FCOS~\cite{TianSCH19}
and CenterNet~\cite{DuanBXQHT19}
from the implementations provided by the authors.
Table~\ref{tab:det_GLOPS_Parameter_comparision}
summarizes \#parameters and GFLOPs.
Table~\ref{tab:object_detection_fpn}
and Table~\ref{tab:object_detection_anchorfree}
report detection scores.

We also evaluate the performance of joint detection and instance segmentation,
under three frameworks: 
Mask R-CNN~\cite{HeGDG17},
Cascade Mask R-CNN~\cite{CaiV19},
and Hybrid Task Cascade~\cite{ChenPWXLSFLSOLL19}.
The results
are 
obtained on the public MMDetection platform~\cite{mmdetection}
and
are in Table~\ref{tab:object_detection_maskrcnn}.

There are several observations.
On the one hand,
as shown in Tables~\ref{tab:object_detection_fpn}
and~\ref{tab:object_detection_anchorfree},
the overall object detection performance
of HRNetV$2$ is
better than ResNet
under similar model size and computation complexity.
In some cases,
for $1\times$,
HRNetV2p-W$18$ performs worse than ResNet-$50$-FPN,
which might come from insufficient optimization iterations.
On the other hand,
as shown in Table~\ref{tab:object_detection_maskrcnn},
the overall object detection and instance segmentation performance
is better than ResNet and ResNeXt.
In particular,
under the Hybrid Task Cascade framework,
the HRNet performs slightly worse than ResNeXt-$101$-$64$$\times$$4$d-FPN
for $20e$,
but better for $28e$.
This implies that our HRNet benefits more 
from longer training.

Table~\ref{tab:recent_object_detection_results_single}
reports the comparison
of our network to state-of-the-art single-model object detectors on COCO \texttt{test-dev} without 
using multi-scale training and 
multi-scale testing that are done in~\cite{LIUQQSJ18, QiLSJ18, LiCYD18, SinghND18,SinghD18,PengXLJZJYS18}.
In the Faster R-CNN framework,
our networks perform better than ResNets with similar 
parameter and computation complexity:
HRNetV$2$p-W$32$ vs. ResNet-$101$-FPN,
HRNetV$2$p-W$40$ vs. ResNet-$152$-FPN,
HRNetV$2$p-W$48$ vs. X-$101$-$64\times4$d-FPN.
In the Cascade R-CNN 
and CenterNet framework,
our HRNetV$2$ also performs better.
In the Cascade Mask R-CNN and Hybrid Task Cascade frameworks,
the HRNet gets the overall better performance.

\section{Ablation Study}

We perform the ablation study
for the components in HRNet
over two tasks:
human pose estimation on COCO validation
and semantic segmentation
on Cityscapes validation. 
We mainly use HRNetV1-W$32$ for human pose estimation,
and HRNetV2-W$48$ for semantic segmentation. 
All results of pose estimation are obtained over the input size $256\times192$. 
We also present the results
for comparing HRNetV$1$ and HRNetV$2$.

\begin{figure}[t]
	\centering
	\includegraphics[width=0.9\linewidth]{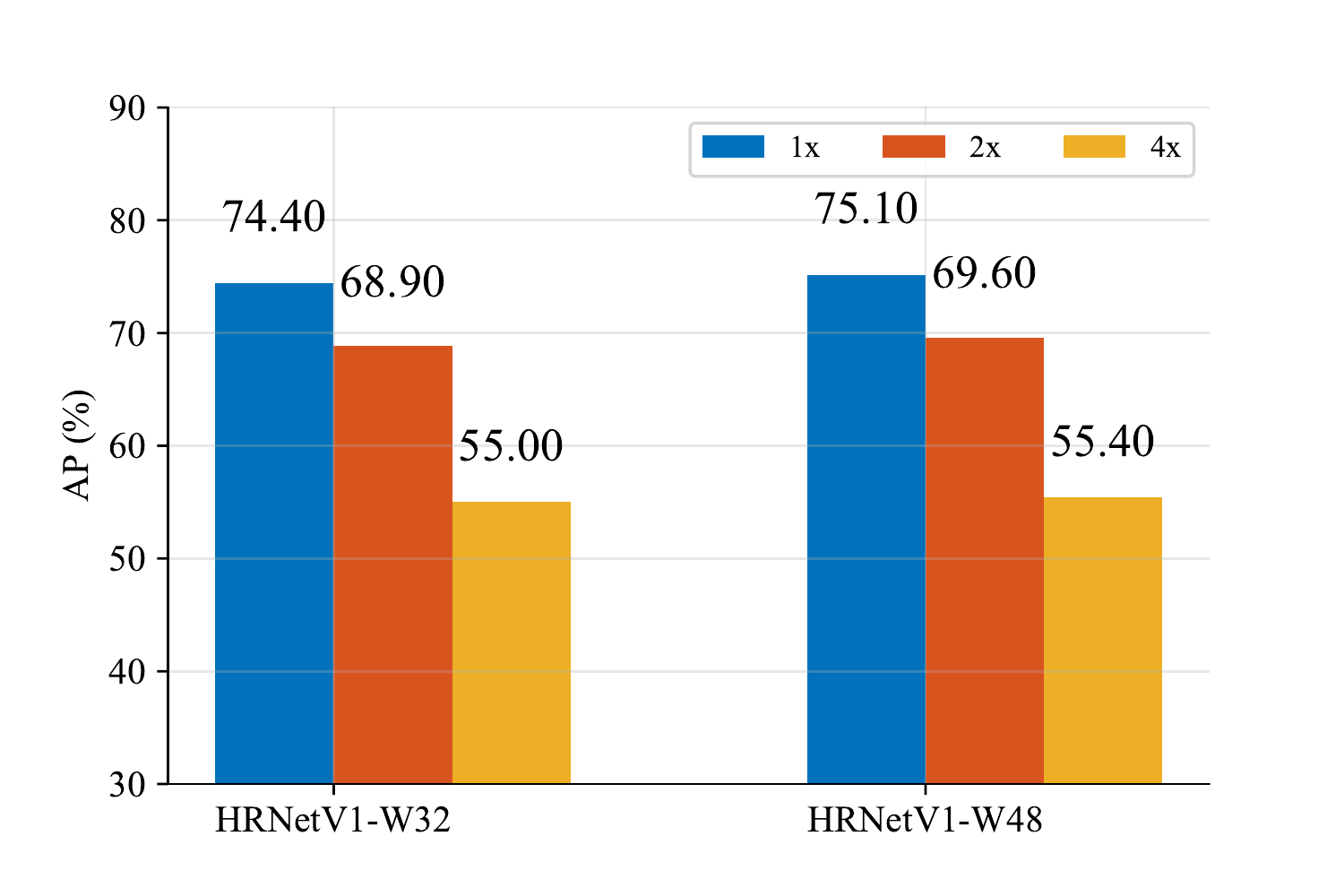}
	\vspace{-3mm}
	\caption{Ablation study about the resolutions of the representations
	for human pose estimation.
			$1\times$, $2\times$, $4\times$ correspond to 
            the representations of 
            the high, medium, low resolutions, respectively. 
            {
            The results imply that 
            higher resolution improves the performance.}
	}
	\label{fig:AblationStudyResolution}
	\vspace{-2mm}
\end{figure}

\vspace{.1cm}
\noindent\textbf{Representations
of different resolutions.}
We study how the representation resolution affects the pose estimation performance 
by checking the quality of the heatmap estimated 
from the feature maps of
each resolution from high to low.

We train two HRNetV$1$ networks initialized 
by the model pretrained for the ImageNet classification.
Our network outputs four response maps from high-to-low {resolutions}.
The quality of heatmap prediction over the lowest-resolution response map
is too low and the AP score is below $10$ points.
The AP scores over the other three maps
are reported in Figure~\ref{fig:AblationStudyResolution}.
The comparison implies that 
the resolution does impact the keypoint prediction quality.

\renewcommand{\arraystretch}{1.3}
\begin{table}[bpt]
\setlength{\tabcolsep}{5pt}
\scriptsize
\caption{Ablation study for multi-resolution fusion units
on COCO \texttt{val} human pose estimation (AP) 
and Cityscapes \texttt{val} semantic segmentation (mIoU).
Final = final fusion immediately 
before representation head,
Across = intermediate fusions across stages,
Within = intermediate fusions within stages.
{
We can see that the three fusions are beneficial
for both human pose estimation 
and semantic segmentation.}}
\label{tab:ablation_exchange_units}
\centering
\vspace{-3mm}
	\begin{tabular}{c|c|c|c|c|c}
	\hline
	Method & Final & Across & Within &  Pose $(\operatorname{AP})$ & Segmentation $ (\operatorname{mIoU})$\\
	\hline
	(a) & \checkmark &  &    & $70.8$ & $74.8$\\
	(b) & \checkmark & \checkmark &    & $71.9$ & $75.4$\\
	(c) & \checkmark & \checkmark & \checkmark    & $73.4$ & $76.4$ \\

	\hline
	\end{tabular}
	\vspace{-2mm}
\end{table}

\vspace{.1cm}
\noindent\textbf{Repeated multi-resolution fusion.}
We empirically analyze 
the effect of the repeated multi-resolution fusion.
We study three variants of our network.
(a) W/o intermediate fusion units ($1$ fusion):
There is no fusion between multi-resolution streams
except the final fusion unit.
(b) W/ across-stage fusion units ($3$ fusions):
There is no fusion between parallel streams
within each stage.
(c) W/ both across-stage and within-stage fusion units 
(totally $8$ fusions):
This is our proposed method.
All the networks
are trained from scratch.
The results on COCO human pose estimation
and Cityscapes semantic segmentation (validation) given in Table~\ref{tab:ablation_exchange_units}
show that the multi-resolution fusion unit is helpful
and more fusions lead to better performance.

{
We also study other possible choices 
for the fusion design:
(i) use bilinear downsample to replace strided convolutions,
and (ii) use the multiplication operation 
to replace the sum operation.
In the former case,
the COCO pose estimation AP score
and the Cityscapes segmentation mIoU score
are reduced to $72.6$ and $74.2$.
The reason is that downsampling reduces the volume size
(width $\times$ height $\times$ \#channels) of the representation maps, and strided convolutions learn better volume size reduction than bilinear downsampling.
In the later case,
the results are much worse:
$54.7$ and $66.0$, respectively.
The possible reason might be that multiplication
increases the training difficulty as pointed in~\cite{WangXLQZZTY17}.
}

\vspace{.1cm}
\noindent\textbf{Resolution maintenance.}
We study the performance of
a variant of the HRNet:
all the four high-to-low resolution streams are added at the beginning
and the depths of the four streams are the same; 
the fusion schemes are the same to ours.
Both the HRNets and the variants
(with similar \#Params and GFLOPs)
are trained from scratch.

The human pose estimation performance
(AP) on COCO \texttt{val}
for the variant is $72.5$, which is lower than $73.4$ for HRNetV$1$-W$32$.
The segmentation performance (mIoU)
on Cityscapes \texttt{val}
for the variant 
is $75.7$, which is lower than $76.4$ for HRNetV$2$-W$48$.
We believe that the reason is that
the low-level features extracted from the early stages
over the low-resolution streams
are less helpful.
In addition, 
another simple variant,
only the high-resolution stream
of similar \#parameters and GFLOPs
without low-resolution parallel streams
shows much lower performance on COCO and Cityscapes.

\vspace{.1cm}
\noindent\textbf{V$1$ vs. V$2$.}
We compare HRNetV$2$ and HRNetV$2$p,
to HRNetV$1$
on pose estimation, semantic segmentation and COCO object detection.
For human pose estimation,
the performance is similar. 
For example,
HRNetV$2$-W$32$ (w/o ImageNet pretraining) 
achieves the AP score $73.6$, 
which is slightly higher than $73.4$ HRNetV$1$-W$32$.

\begin{figure}[t]
    \centering
    \footnotesize
    \subfloat[]{\includegraphics[height=5.6cm]{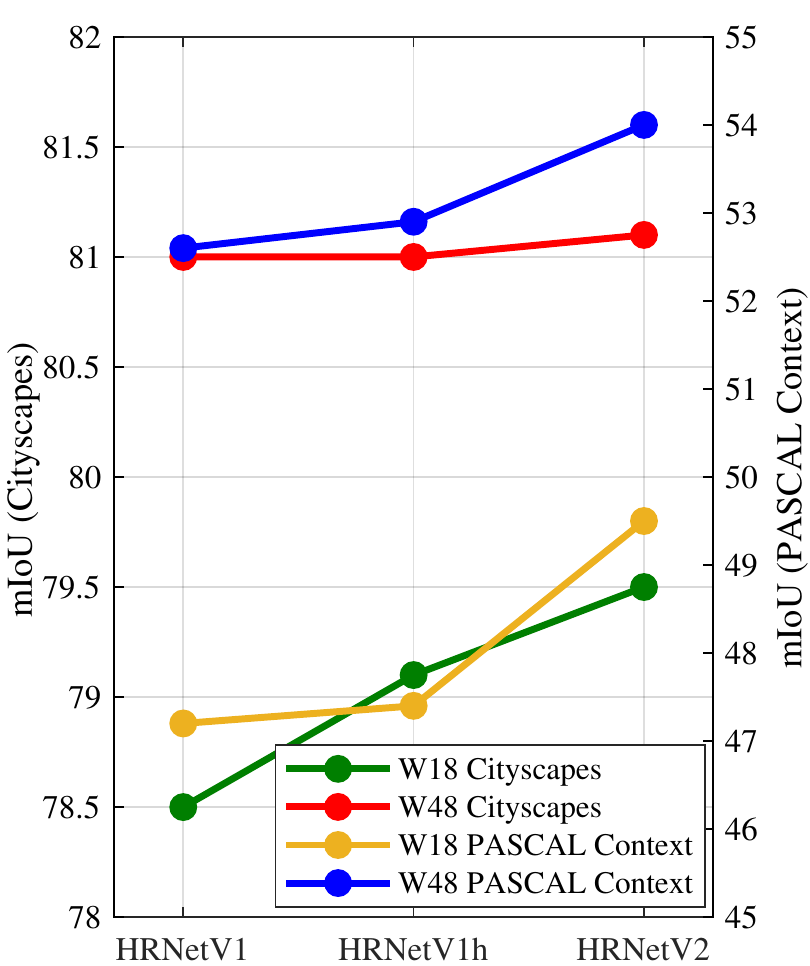}}~~
    \subfloat[]{\includegraphics[height=5.6cm]{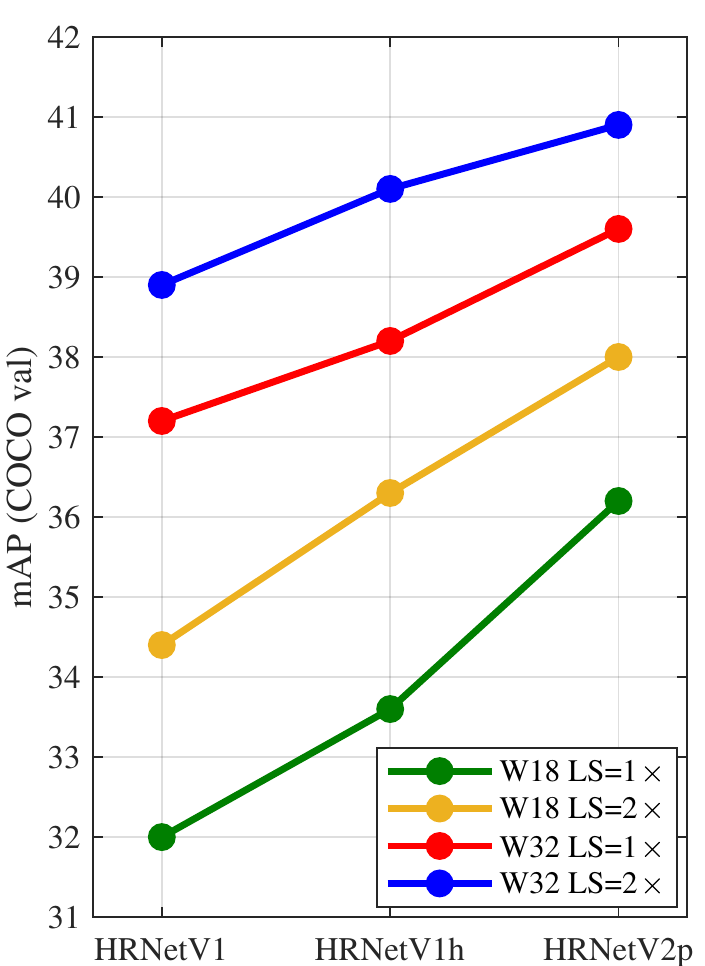}}
    \vspace{-3mm}
    \caption{Comparing HRNetV$1$ and HRNetV$2$.
    (a) Segmentation on Cityscapes \texttt{val} and PASCAL-Context for comparing HRNetV$1$ and its variant HRNetV$1$h, and HRNetV$2$ (single scale and no flipping).
    (b) Object detection on COCO \texttt{val} for comparing HRNetV$1$ and its variant HRNetV$1$h, and HRNetV$2$p (LS = learning schedule).
    {We can see that HRNetV$2$ is superior 
    to HRNetV$1$ for both semantic segmentation and object detection.}}
    \label{fig:empiricalstudy}
    \vspace{-2mm}
\end{figure}

The segmentation and object detection results, given in 
Figure~\ref{fig:empiricalstudy} (a)
and Figure~\ref{fig:empiricalstudy} (b),
imply that HRNetV$2$ outperforms HRNetV$1$ significantly,
except that
the gain is minor
in the large model case ($1\times$)
in segmentation for Cityscapes.
We also test a variant
(denoted by HRNetV$1$h),
which is built by appending a $1\times 1$ convolution to 
align the dimension of the output high-resolution representation
with the dimension of HRNetV$2$.
The results in Figure~\ref{fig:empiricalstudy} (a) and Figure~\ref{fig:empiricalstudy} (b)
show that the variant achieves slight improvement
to HRNetV$1$,
implying that aggregating the representations
from low-resolution parallel convolutions in our HRNetV$2$ 
is essential for improving the capability.

\begin{table*}[t]

    \centering\setlength{\tabcolsep}{2.0pt}
    \scriptsize
    \caption{
    Memory and time cost comparisons
    for pose estimation, semantic segmentation and object detection (under the Faster R-CNN framework) on PyTorch $1.0$
    in terms of training/inference memory and training/inference time. 
    We also report inference time (in $()$) for pose estimation on MXNet $1.5.1$,
    which supports static graph inference
    that mutli-branch convolutions used in 
    the HRNet benefits from.
    The numbers for training are obtained
    on a machine with $4$ V$100$ GPU cards.
    During training,
    the input sizes are $256\times192$, $512\times1024$, and $800\times1333$,
    and the batch sizes
    are $128$, $8$ and $8$ for pose estimation, segmentation and detection respectively.
    The numbers for inference
    are obtained on a single V$100$ GPU card.
    The input sizes are $256\times192$, $1024\times2048$, and $800\times1333$, respectively.
   The score means AP for pose estimation 
   on COCO \texttt{val} (Table~\ref{table:coco_val})
   and detection on COCO \texttt{val}
   (Table~\ref{tab:object_detection_fpn})
   , and mIoU for cityscapes segmentation
   (Table~\ref{tab:cityscapevalresults}). 
    Several observations are highlighted.
    Memory: 
    The HRNet consumes similar memory for both training and inference
    except that it consumes smaller memory for training in human pose estimation.
    Time: The training and inference time cost of the HRNet
    is comparable to previous state-of-the-arts
    except that the inference time of the HRNet for segmentation is much smaller.
   SB-ResNet-$152$ = SimpleBaseline with the backbone of ResNet-$152$.
   PSPNet and DeepLabV3 use dilated ResNet-$101$
   as the backbone (Table~\ref{tab:cityscapevalresults}).}
   \vspace{-2mm}
    \label{tab:memorytime}
    \begin{tabular}{l|cc|ccc|cccc}
        \hline %
		\multirow{2}{*}{} & \multicolumn{2}{c|}{Pose estimation}  & \multicolumn{3}{c|}{Segmentation} & \multicolumn{4}{c}{Detection} \\
		\cline{2-10}
		& SB-ResNet-$152$ & HRNetV$1$-W$48$ & PSPNet & DeepLabV$3$& HRNetV$2$-W$48$ & ResNet-$101$ & ResNeXt-$101$ & HRNetV$2$p-W$32$ & HRNetV$2$p-W$48$ \\
		\hline
        
        \hline
        training memory & $14.8$G & $7.3$G & $14.4$G & $13.3$G & $13.9$G& $5.4$G & $9.5$G & $8.5$G & $11.3$G\\
        inference memory/image & $0.29$G & $0.27$G & $1.60$G& $1.15$G & $1.79$G & $0.62$G & $0.77$G & $0.51$G & $0.79$G \\
        \hline
        training second/iteration & $1.085$ & $1.231$ & $0.837$ & $0.850$ & $0.692$ & $0.550$ & $1.183$ & $0.690$ & $0.965$ \\
        inference second/image  & $0.030~(0.012)$ & $0.058~(0.017)$ & $0.397$ & $0.411$ & $0.150$ & $0.087$ & $0.144$ & $0.101$ & $0.116$ \\
        \hline
        score & $72.0$ & $75.1$ & $79.7$ & $78.5$ & $81.1$ & $39.8$ & $40.8$ & $40.9$ & $41.8$ \\ 
        \hline
    \end{tabular} 
\end{table*}

\section{Conclusions}
In this paper,
we present a high-resolution network 
for visual recognition problems.
There are three fundamental differences from
existing low-resolution classification networks 
and high-resolution representation learning networks:
(\romannum{1}) Connect high and low resolution convolutions
in parallel other than in series;
(\romannum{2}) Maintain high resolution through the whole process instead of recovering high resolution
from low resolution;
and (\romannum{3}) Fuse multi-resolution representations repeatedly,
rendering rich high-resolution representations
with strong position sensitivity.

The superior results on a wide range of visual recognition problems
suggest that our proposed HRNet is a stronger backbone for computer vision problems.
Our research also encourages more research efforts
for designing network architectures directly for specific vision problems 
other than extending, remediating or repairing
representations learned from low-resolution networks (e.g., ResNet or VGGNet).

\vspace{1mm}
\noindent\textbf{Discussions.}
{There is a possible misunderstanding:
the memory cost of the HRNet is larger
as the resolution is higher.
In fact,
the memory cost of the HRNet for all the three applications, human pose estimation,
semantic segmentation and object detection,
is comparable to state-of-the-arts
except that the training memory cost in object detection is a little larger.}

{
In addition,
we summarize the runtime cost comparison
on the PyTorch $1.0$ platform.
The training and inference time cost of the HRNet
    is comparable to previous state-of-the-arts
    except that (1) the inference time of the HRNet for segmentation is much smaller
    and (2) the training time of the HRNet
    for pose estimation is a little larger,
    but the cost on the MXNet $1.5.1$ platform,
which supports static graph inference,
is similar as SimpleBaseline.
We would like to highlight
that for semantic segmentation
the inference cost is significantly smaller
than PSPNet and DeepLabv$3$.
Table~\ref{tab:memorytime}
summarizes memory and time cost comparisons \footnote{The detailed comparisons 
are given in the supplementary file.}.
}

\vspace{1mm}
\noindent\textbf{Future and followup works.}
We will study the combination of the HRNet
with other techniques for semantic segmentation and
instance segmentation. 
Currently, we have results (mIoU),
which are depicted in Tables~\ref{tab:cityscapevalresults}~\ref{tab:cityscaperesults}~\ref{tab:pasctxresults}~\ref{tab:lipresults},
by combining the HRNet with the object-contextual representation
(OCR) scheme~\cite{YuanCW18}
\footnote{We empirically observed
that the HRNet combined with ASPP~\cite{ChenPSA17}
or PPM~\cite{ZhaoSQWJ17}
did not get a performance improvement on Cityscape,
but got a slight improvement on PASCAL-Context and LIP.},
a variant of object context~\cite{YuanW18, HuangYGZCW19}.
{We will
conduct the study by further 
increasing the resolution
of the representation,
e.g., to $\frac{1}{2}$ or even a full resolution.}

The applications of the HRNet
are not limited to 
the above that we have done,
and are suitable to other position-sensitive 
vision applications,
such as 
facial landmark detection\footnote{
We provide the facial landmark detection results
in the supplementary file.},
super-resolution,
optical flow estimation,
depth estimation,
and so on.
There are already followup works,
e.g., image stylization~\cite{LiYL19},
inpainting~\cite{GuoCYCL19},
image enhancement~\cite{Ignatov_2019_CVPR_Workshops},
image dehazing~\cite{Ancuti_2019_CVPR_Workshops},
temporal pose estimation~\cite{BertasiusFTST19},
{and drone object detection~\cite{Zhu_2019_ICCV}}.

{
It is reported in~\cite{ChengCZLHAC19}
that a slightly-modified HRNet
combined with ASPP achieved
the best performance for Mapillary panoptic segmentation
in the single model case.
In the COCO + Mapillary Joint Recognition Challenge Workshop at ICCV 2019,
the COCO DensePose challenge winner
and almost all the COCO keypoint detection challenge participants adopted the HRNet.
The OpenImage instance segmentation challenge winner
(ICCV 2019)
also used the HRNet.}

{\small
\bibliographystyle{ieee}
\bibliography{bib/HRNetV2,bib/HRNetSegmentation,bib/HRNetFaceAlignment,bib/HRNetObjectDetection,bib/HRNetApps}
}

\ifCLASSOPTIONcaptionsoff
  \newpage
\fi

\appendices
\onecolumn
\section{Network Instantiation}
Our current design (except the standard stem and the head,)
contains four stages, as shown in Table~\ref{tab:hrnet-block}.
Each stage consists of modularized blocks,
repeated $1$, $1$, $4$, and $3$ times,
respectively for the four stages.
The modularized block consists of $1$ ($2$, $3$ and $4$) branches
for the $1$st ($2$nd, $3$rd and $4$th) stages.
Each branch corresponds to different resolution,
and is compose of four residual units and one multi-resolution fusion unit (See Figure {\color{red}3} in the main paper).

\newcommand{\blocka}[3]{\multirow{3}{*}{
 $\left[\begin{array}{c}{3\times3, #1}\\[-.1em] {3\times3, #1} \end{array}
\right]
\times#2 
\times#3$
}
}

\newcommand{\blockb}[3]{\multirow{3}{*}{\(\left[\begin{array}{c}{1\times1, #2}\\[-.1em] {3\times3, #2}\\[-.1em] {1\times1, #1}\end{array}\right]\)
$\times$#3}
}

\renewcommand{\arraystretch}{1.3}
\begin{table*}[!htbp]
\setlength{\tabcolsep}{13pt}
\scriptsize
\caption{
\footnotesize 
The architecture of the HRNet (main body). 
There are four stages.
Each stage consists of modularized blocks,
repeated $1$, $1$, $4$, and $3$ times,
respectively for the four stages.
The modularized block consists of $1$ ($2$, $3$ and $4$) branches
for the $1$st ($2$nd, $3$rd and $4$th) stages.
Each branch corresponds to a different resolution,
and is composed of four residual units and one multi-resolution fusion unit.
For clarify,
the fusion unit (after each modulized block) is not depicted in the table,
and could be understood from Figure 3 in the main paper.
In the table,
each cell consists of three components:
the first one ($[~\cdot~]$)
is the residual unit,
the second number is the repetition times
of the residual units,
and the last number is the repetition times
of the modualized blocks.
$C$ in each residual unit is the number of channels.
}
\label{tab:hrnet-block}
\centering
	\begin{tabular}{l|c|c|c|c}
	\hline
	 Resolution & Stage $1$ & Stage $2$ & Stage $3$ & Stage $4$ \\
	\hline
	\multirow{3}{*}{$4\times$} 
	& \blockb{256}{64}{$4\times1$} & \blocka{C}{4}{1}  & \blocka{C}{4}{4} & \blocka{C}{4}{3} \\
	&  &  &  &  \\
	&  &  &  &  \\
	\hline
	\multirow{3}{*}{$8\times$}
	&  & \blocka{2C}{4}{1}  & \blocka{2C}{4}{4} & \blocka{2C}{4}{3}\\
	&  &  &  &  \\
	&  &  &  &  \\
	\hline
	\multirow{3}{*}{$16\times$}
	&  &  & \blocka{4C}{4}{4} & \blocka{4C}{4}{3} \\
	&  &  &  &  \\
	&  &  &  &  \\
	\hline
	\multirow{3}{*}{$32\times$}
	&  &  &  & \blocka{8C}{4}{3}\\
	&  &  &  &  \\
	&  &  &  &  \\
	\hline
	\end{tabular}
\end{table*}

\section{Network Pretraining}
\begin{figure}[!htbp]
\footnotesize
    \centering
    \includegraphics[width=0.5\linewidth]{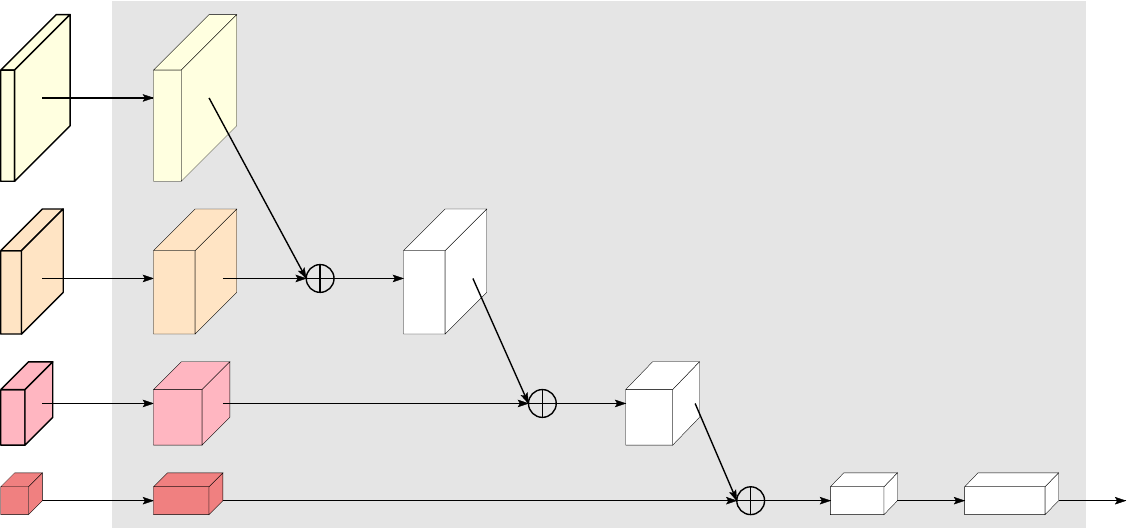}
    \caption{Representation for ImageNet classification.
    The input of the box
    is the representations
    of four resolutions.}
    \label{fig:classificationhead}
\end{figure}

We pretrain our network,
which is augmented by a classification head shown in Figure~\ref{fig:classificationhead},
on ImageNet~\cite{RussakovskyDSKS15}. 
The classification head is described as below.
First, the four-resolution feature maps are fed into 
a bottleneck and the output channels are 
increased 
from 
$C$, $2C$, $4C$, and $8C$
to $128$, $256$, $512$, and $1024$, respectively.
Then, we downsample the high-resolution representation
by a $2$-strided $3 \times 3$ convolution
outputting $256$ channels
and add it to the representation of the second-high-resolution.
This process is repeated two times
to get $1024$ feature channels over the small resolution.
Last, we transform the $1024$ channels to $2048$ channels
through a $1 \times 1$ convolution, followed by a global average pooling operation. 
The output $2048$-dimensional representation is fed into the classifier.

	\begin{table}[ht]
	\setlength{\tabcolsep}{15pt}
	\scriptsize
	\centering
	\caption{ImageNet Classification results of HRNet and ResNets. The proposed method is named HRNet-W$x$-C. $x$ means the width.}
	\begin{tabular}{l|cc|cc}
		\hline
		& \#Params. & GFLOPs & top-1 err. & top-5 err. \\
		\hline
		
		\hline
		\multicolumn{5}{l}{\emph {Residual branch formed by two $3\times 3$ convolutions}}\\
		\hline
		ResNet-$38$ & $28.3$M & $3.80$ & $24.6\%$ & $7.4\%$ \\
		HRNet-W$18$-C& $21.3$M & $3.99$ & $\mathbf{23.1\%}$ & $\mathbf{6.5\%}$ \\
		\hline
		ResNet-$72$ & $48.4$M & $7.46$ & $23.3\%$ & $6.7\%$ \\
		HRNet-W$30$-C& $37.7$M & $7.55$ & $\mathbf{21.9\%}$ & $\mathbf{5.9\%}$ \\
		\hline
		ResNet-$106$& $64.9$M & $11.1$ & $22.7\%$ & $6.4\%$ \\	
		HRNet-W$40$-C& $57.6$M & $11.8$ & $\mathbf{21.1\%}$ & $\mathbf{5.6\%}$ \\
		\hline
		
		\hline
		\multicolumn{5}{l}{\emph {Residual branch formed by a bottleneck}}\\
		\hline
		ResNet-$50$& $25.6$M & $3.82$ & $23.3\%$ & $6.6\%$ \\
		HRNet-W$44$-C& $21.9$M & $3.90$ & $\mathbf{23.0\%}$ & $\mathbf{6.5\%}$ \\
		\hline
		ResNet-$101$& $44.6$M & $7.30$ & $21.6\%$ & $\mathbf{5.8\%}$ \\
		HRNet-W$76$-C& $40.8$M & $7.30$ & $\mathbf{21.5\%}$ & $\mathbf{5.8\%}$ \\
		\hline
		ResNet-$152$& $60.2$M & $10.7$ & $21.2\%$ & $5.7\%$ \\
		HRNet-W$96$-C& $57.5$M & $10.2$ & $\mathbf{21.0\%}$ & $\mathbf{5.6\%}$ \\
		\hline
	\end{tabular}
	\label{tab:ImageNetClassificationComparison}
	\end{table}

We adopt the same data augmentation scheme for training images as in \cite{HeZRS16}, 
and train our models for $100$ epochs with a batch size of $256$. 
The initial learning rate is set to $0.1$ 
and is reduced by $10$ times at epoch $30$, $60$ and $90$.
We use SGD with a weight decay of $0.0001$ and a Nesterov momentum of $0.9$. We adopt standard single-crop testing, so that $224\times 224$ pixels are cropped from each image. The top-$1$ 
and top-$5$ error are reported on the validation set.

Table~\ref{tab:ImageNetClassificationComparison}
shows our ImageNet classification results.
As a comparison, we also report the results of ResNets. 
We consider two types of residual units:
One is formed by a bottleneck,
and the other is formed by two $3 \times3 $ convolutions. We follow the 
PyTorch implementation of ResNets and replace the $7 \times 7$ 
convolution in the input stem with two 
$2$-strided $3 \times 3$ convolutions decreasing the resolution to $1/4$ 
as in our networks. When the residual units are formed by two $3 \times3 $ convolutions,
an extra bottleneck is used to increase the dimension of output feature maps from $512$ to $2048$.
One can see that under similar \#parameters
and GFLOPs, our results are comparable to and slightly better
than ResNets.

In addition, we look at the results 
of two alternative schemes:
(i) the feature maps on each resolution go through a global pooling separately and then are concatenated 
together to output a $15C$-dimensional representation vector,
named HRNet-W$x$-Ci;
(ii) the feature maps on each resolution are fed into several $2$-strided residual units (bottleneck, each dimension is increased to the double)
to increase the dimension to $512$,
and concatenate and average-pool them together to reach a $2048$-dimensional representation vector,
named HRNet-W$x$-Cii,
which is used in~\cite{SunXLW19}.
Table~\ref{tab:ImageNetClassificationAblationStudy}
shows such an ablation study. 
One can see that the proposed manner is superior to the two alternatives.

	\begin{table}[ht]
	\setlength{\tabcolsep}{15pt}
	\tiny
	\scriptsize
	\centering
	\caption{Ablation study on ImageNet classification
	by comparing our approach (abbreviated as HRNet-W$x$-C) with two alternatives: HRNet-W$x$-Ci and HRNet-W$x$-Cii (residual branch formed by two $3\times 3$ convolutions).}
	\begin{tabular}{l|cc|cc}
		\hline
		& \#Params. & GFLOPs & top-1 err. & top-5 err.\\
		\hline 
	    HRNet-W$27$-Ci& $21.4$M & $5.55$ & $26.0\%$ & $7.7\%$ \\
        HRNet-W$25$-Cii& $21.7$M & $5.04$ & $24.1\%$ & $7.1\%$ \\
	    HRNet-W$18$-C& $21.3$M & $3.99$ & $\mathbf{23.1}\%$ & $\mathbf{6.5\%}$ \\
        \hline
        HRNet-W$36$-Ci& $37.5$M & $9.00$ & $24.3\%$ & $7.3\%$ \\
        HRNet-W$34$-Cii& $36.7$M & $8.29$ & $22.8\%$ & $6.3\%$ \\
	    HRNet-W$30$-C& $37.7$M & $7.55$ & $\mathbf{21.9\%}$ & $\mathbf{5.9\%}$ \\
        \hline
        HRNet-W$45$-Ci& $58.2$M & $13.4$ & $23.6\%$ & $7.0\%$ \\
        HRNet-W$43$-Cii& $56.3$ M & $12.5$ & $22.2\%$ & $6.1\%$ \\
	    HRNet-W$40$-C& $57.6$M & $11.8$ & $\mathbf{21.1\%}$ & $\mathbf{5.6\%}$ \\
	    \hline
	\end{tabular}
	\label{tab:ImageNetClassificationAblationStudy}
	\end{table}

\clearpage
\onecolumn
\section{Training/Inference Cost}
Tables~\ref{tab:pytroch_pose_comparison}, \ref{tab:pytorch_seg_comparison} and \ref{tab:pytorch_det_comparison} provide GPU memory comparisons between HRNets and other standard networks
for both training and inference
in the PyTorch platform.
Compared to state-of-the-arts for human pose estimation,
the training and inference memory costs of the HRNet are similar or lower
for similar parameter complexity (Table~\ref{tab:pytroch_pose_comparison}).
Compared to state-of-the-arts for semantic segmentation,
the training and inference memory costs are similar 
(Table~\ref{tab:pytorch_seg_comparison})
for similar parameter complexity.
Compared to state-of-the-arts for object detection for similar parameter complexity,
the training and inference memory costs 
are similar or slightly higher
(Table~\ref{tab:pytorch_det_comparison}).

In addition,
we provide the runtime cost comparison.
(1) For semantic segmentation,
the time cost of the HRNet for training is slightly smaller
and for inference significantly smaller
than PSPNet and DeepLabv3 (Table~\ref{tab:pytorch_seg_comparison}).
(2) For object detection,
the time cost of the HRNet for training is larger than ResNet based networks and smaller than ResNext based networks, and for inference the HRNet 
is smaller for similar GFLOPs
(Table~\ref{tab:pytorch_det_comparison}).
(3) For human pose estimation,
the time cost of the HRNet for training 
is similar
and for inference
larger;
and the time cost of the HRNet for training and inference
in the MXNet platform 
is similar as SimpleBaseline (Table~\ref{tab:pytroch_pose_comparison}).

\begin{table*}[!h]
    \centering\setlength{\tabcolsep}{2pt}
    \scriptsize
    \caption{
    \footnotesize Human pose estimation complexities
    on PyTorch $1.0$
    in terms of \#parameters,
    GFLOPs, training/inference memory. We also report training/inference time on Pytorch $1.0$ and MXNet $1.5.1$, shown as $time$(Pytorch)$/time$(MXNet). The reason that the runtime cost is smaller than PyTorch is that MXNet supports dynamic graph
    which the HRNet benefits from.
    We compare the HRNet and the previous state-of-the-art, Simplebaseline. 
    Training: $4$ V$100$ GPU cards, input size $256\times192$,
    and batch size $128$.
    Inference: a single V$100$ GPU card,
    input size $256\times192$,
    and batch size $1$, $4$, $8$ and $16$.
    Two observations are highlighted here:
    (1) The HRNet consumes smaller memory for both training and inference
    for similar \#parameters,
    and for similar AP scores;
    (2) The HRNet takes slightly higher training runtime cost
    and a little higher inference runtime cost on Pytorch and similar on MXNet
    for similar GFLOPs,
    and the inference efficiency of HRNet for is improved
    for larger batch size.
    sec.= seconds, iter. = iteration, mem. = memory, bs = batchsize.}
    \label{tab:pytroch_pose_comparison}
    \begin{tabular}{l|cc|cc|cccc|cccc|c}
        \hline %
		\multirow{2}{*}{backbone} & \multirow{2}{*}{GFLOPs}  & \multirow{2}{*}{\#params} & \multirow{2}{*}{train sec./iter.} &\multirow{2}{*}{train mem.}& \multicolumn{4}{c|}{infer sec./batch} & \multicolumn{4}{c|}{infer mem./batch} & \multirow{2}{*}{$\operatorname{AP}$}  \\
		\cline{6-13}
		& & & & & $bs$ = 1 & $bs$ = 4 & $bs$ = 8 & $bs$ = 16 & $bs$ = 1 & $bs$ = 4 & $bs$ = 8 & $bs$ = 16 & \\
		\hline
        
        \hline
        SB-Res-$50$   & $8.90$ & $34.0$M & $0.946/0.211$ & $11.9$G & $0.012/0.005$ & $0.013/0.010$& $0.015/0.017$& $0.024/0.027$ & $0.16$G & $0.17$G & $0.21$G & $0.30$G &${70.4}$ \\
        SB-Res-$101$  & $12.4$ & $53.0$M & $1.008/0.320$ & $13.2$G & $0.020/0.009$ & $0.021/0.014$ & $0.024/0.023$ & $0.035/0.038$ & $0.23$G & $0.24$G & $0.28$G & $0.37$G &${71.4}$ \\
        SB-Res-$152$  & $15.7$ & $68.6$M & $1.085/0.415$ & $14.8$G & $0.030/0.012$ & $0.033/0.019$ & $0.035/0.031$ & $0.048/0.051$ & $0.29$G & $0.31$G & $0.35$G & $0.43$G &${72.0}$ \\
        \hline
        HRNetV$1$-W$32$  & $7.10$ & $28.5$M & $1.153/0.389$ & $5.7$G & $0.057/0.015$ & $0.059/0.017$ & $0.061/0.020$ & $0.062/0.031$ & $0.13$G & $0.15$G& $0.19$G & $0.28$G & $74.4$ \\
        HRNetV$1$-W$48$  & $14.6$ & $63.6$M & $1.231/0.507$ & $7.3$G & $0.058/0.017$ & $0.060/0.021$ & $0.062/0.033$& $0.066/0.051$ & $0.27$G & $0.30$G & $0.32$G & $0.44$G
        & $75.1$ \\
        \hline
    \end{tabular} 
\end{table*}

\begin{table*}[h!]
    \centering\setlength{\tabcolsep}{7pt}
    \scriptsize
    \caption{
    \footnotesize Semantic segmentation complexities
    on PyTorch $1.0$.
    Training: $4$ V$100$ GPU cards, input size $512\times1024$,
    and batch size $8$.
    Inference: a single V$100$ GPU card,
    input size $1024\times2048$,
    and batch size $1$, $4$, and $8$.
    Several observations are highlighted:
    (1) The training memory costs are similar,
    and the inference memory costs are similar
    but larger for our approach with larger batch size;
    (2) The training and inference time costs of our approach
    are much smaller.
    The results are obtained on Cityscapes \texttt{val}.
    }
    \label{tab:pytorch_seg_comparison}
        \begin{tabular}{l|cc|cc|ccc|ccc|c}
        \hline %
		\multirow{2}{*}{backbone} & \multirow{2}{*}{GFLOPs}  & \multirow{2}{*}{\#params} & \multirow{2}{*}{train sec./iter.} &\multirow{2}{*}{train mem.}& \multicolumn{3}{c|}{infer sec./batch} & \multicolumn{3}{c|}{infer mem./batch} & \multirow{2}{*}{mIoU} \\
		\cline{6-11}
		& & & & & $bs$ = 1 & $bs$ = 4 & $bs$ = 8 & $bs$ = 1 & $bs$ = 4 & $bs$ = 8 & \\
        \hline

        \hline
        Dilated-ResNet & $1661.6$ & $52.1$M & $0.6611$ & $12.4$G & $0.3351$ & $1.2882$ & $2.7039$ & $1.13$G & $3.92$G & $7.64$G & $75.7$\\
        PSPNet & $2017.6$ & $65.9$M & $0.8368$ & $14.4$G & $0.3972$ & $1.5296$ & $3.2003$ & $1.60$G & $5.04$G & $9.81$G & $79.7$\\
        DeepLabv$3$ & $1778.7$ & $58.0$M & $0.8502$ & $13.3$G & $0.4113$ & $1.5307$ & $3.2000$ & $1.15$G & $3.95$G & $7.67$G & $78.5$ \\
        \hline
        HRNetV$2$-W$48$ & $696.2$  & $65.9$M & $0.6920$ & $13.9$G & $0.1502$ & $0.05421$ & $1.1032$ & $1.79$G & $6.37$G & $12.5$G & $81.1$ \\
        \hline
    \end{tabular}
\end{table*}

\begin{table*}[!htbp]
    \centering\setlength{\tabcolsep}{3pt}
    \scriptsize
    \caption{
    \footnotesize COCO object detection complexities
    on PyTorch $1.0$.
    Training: $4$ V$100$ GPU cards, input size $800 \times 1333$,
    and batch size $8$.
    Inference: a single V$100$ GPU card,
    input size $800\times1333$,
    and batch size $1$, $4$, and $8$.
    The performance are reported on the COCO $2017$val for each model with the learning schedule of $2\times$.
    Several observations are as follows:
    (1) The training memory for the HRNet is a little larger for similar \#params,
    but the inference memory are similar with higher AP scores;
    (2) The training runtime costs of the HRNet
    are a little larger than ResNet based networks and smaller than ResNext based networks, and the inference runtime costs is smaller for similar GFLOPs.}
    \vspace{-.2cm}
    \label{tab:pytorch_det_comparison}
    \begin{tabular}{l|cc|cc|ccc|ccc|cccccc}
        \hline %
		\multirow{2}{*}{backbone} & \multirow{2}{*}{GFLOPs}  & \multirow{2}{*}{\#params} & \multirow{2}{*}{sec./iter.} &\multirow{2}{*}{train mem.}& \multicolumn{3}{c|}{infer sec./batch} & 
		\multicolumn{3}{c|}{infer mem./batch} & 
		\multirow{2}{*}{$\operatorname{AP}$} & 
		\multirow{2}{*}{$\operatorname{AP}_{50}$} & \multirow{2}{*}{$\operatorname{AP}_{75}$} & \multirow{2}{*}{$\operatorname{AP}_{S}$} & 
		\multirow{2}{*}{$\operatorname{AP}_{M}$} &
		\multirow{2}{*}{$\operatorname{AP}_{L}$} \\
		\cline{6-11}
		& & & & & $bs$ = 1 & $bs$ = 4 & $bs$ = 8 & $bs$ = 1 & $bs$ = 4 & $bs$ = 8 & & & & & & \\
		\hline

        \hline
        ResNet-$50$   & $172.34$ & $39.77$M & $0.4167$ & $3.4$G & $0.0739$ & $0.2495$ & $0.4921$ & $0.55$G & $1.68$G & $2.93$G & $37.6$ & $58.7$ & $41.3$ & $21.4$ & $40.8$ & $49.7$\\
        ResNet-$101$  & $239.37$ & $57.83$M & $0.5502$ & $5.4$G & $0.0870$ & $0.3018$ & $0.6089$ & $0.62$G & $1.76$G & $3.25$G & $39.8$ & $61.4$ & $43.4$ & $22.9$ & $43.6$ & $52.4$\\
        X-$101$-$64\times4$d & $381.83$ & $94.85$M & $1.1828$ & $9.5$G & $0.1438$ & $0.4832$ & $0.9764$ & $0.77$G & $1.92$G & $3.41$G & $40.8$ & $62.1$ & $44.6$ & $23.2$ & $44.5$ & $53.7$\\
        \hline
        HRNetV$2$p-W$18$  & $159.06$ & $26.18$M & $0.6166$ & $6.2$G & $0.0968$ & $0.2320$ & $0.4471$ & $0.33$G & $1.01$G & $1.91$G & $38.0$ & $58.9$ & $41.5$ & $22.6$ & $40.8$ & $49.6$\\
        HRNetV$2$p-W$32$  & $245.33$ & $45.04$M & $0.6901$ & $8.5$G & $0.1014$ & $0.2738$ & $0.5546$ & $0.51$G & $1.51$G & $2.83$G & $40.9$ & $61.8$ & $44.8$ & $24.4$ & $43.7$ & $53.3$ \\
        HRNetV$2$p-W$48$  & $399.12$ & $79.42$M & $0.9648$ & $11.3$G & $0.1162$& $0.3762$ & $0.7296$ & $0.79$G  & $2.16$G & $3.99$G & $41.8$ & $62.8$ & $45.9$ & $25.0$ & $44.7$ & $54.6$\\
        \hline
    \end{tabular} 
\end{table*}

\clearpage

\section{Facial Landmark Detection}
Facial landmark detection
a.k.a. face alignment
is a problem of detecting the keypoints
from a face image.
We perform the evaluation over 
four standard datasets:
WFLW~\cite{Wu0YWC018},
AFLW~\cite{KostingerWRB11},
COFW~\cite{Burgos-ArtizzuPD13},
and $300$W~\cite{SagonasTZP13}. 
We mainly use the normalized mean error (NME) for evaluation.
We use the inter-ocular distance as normalization
for WFLW, COFW, and $300$W, and the face bounding box as normalization
for AFLW.
We also report area-under-the-curve scores (AUC) and failure rates. 

We follow the standard scheme~\cite{Wu0YWC018} for training.
All the faces are cropped by the provided boxes according 
to the center location and resized to $256 \times 256$. 
We augment the data by $\pm30$ degrees in-plane rotation,
$0.75-1.25$ scaling, and
randomly flipping. 
The base learning rate is $0.0001$ and is dropped to
$0.00001$ and $0.000001$ at the $30$th and $50$th epochs. The models 
are trained for $60$ epochs with the batch size of $16$ on
one GPU. Different from semantic segmentation,
the heatmaps are not upsampled from $1/4$ to the input size,
and the loss function is optimized over
the $1/4$ maps.

At testing,
each keypoint location is predicted
by transforming the highest heatvalue location from $1/4$
to the original image space
and adjusting it with a quarter offset
in the direction from the highest response
to the second highest response~\cite{ChenSWLY17}.

We adopt HRNetV$2$-W$18$ for face landmark detection
whose parameter and computation cost
are similar to or smaller than 
models with widely-used backbones: ResNet-$50$ and Hourglass~\cite{NewellYD16}.
HRNetV$2$-W$18$: \#parameters $=9.3$M, GFLOPs $=4.3$G;
ResNet-$50$: \#parameters $=25.0$M, GFLOPs $=3.8$G;
Hourglass: \#parameters $=25.1$M, GFLOPs $=19.1$G.
The numbers are obtained on the input size $256 \times 256$.
It should be noted that
the facial landmark detection methods adopting ResNet-$50$ and Hourglass as backbones
introduce extra parameter and computation overhead.

\vspace{.1cm}
\noindent\textbf{WFLW.} 
The WFLW dataset \cite{Wu0YWC018} is a recently-built  
dataset based on the WIDER Face \cite{YangLLT16}. 
There are $7,500$ training and $2,500$ testing images 
with $98$ manual annotated landmarks.
We report the results on the test set and several subsets: 
large pose ($326$ images), expression ($314$ images),
illumination ($698$ images), make-up ($206$ images), occlusion ($736$ images) and blur ($773$
images).

Table \ref{table:comparison_wflw_testset} provides the comparison of our method with state-of-the-art methods. 
Our approach is significantly better than other methods on the test set and all the subsets,
including LAB that exploits extra boundary information~\cite{Wu0YWC018} and PDB that uses stronger data augmentation~\cite{FengKA0W18}.

\vspace{.1cm}
\noindent\textbf{AFLW.}
The AFLW \cite{KostingerWRB11} dataset is
a widely used benchmark dataset,
where each image has $19$ facial landmarks. Following \cite{ZhuLLT15, Wu0YWC018}, we train our models on $20,000$ training images, and report the results on the AFLW-Full set ($4,386$ testing images) and the AFLW-Frontal set ($1314$ testing images selected from $4386$ testing images).

Table \ref{table:comparison_aflw_testset} provides the comparison of our method with state-of-the-art methods. 
Our approach achieves the best performance among methods without extra information and stronger data augmentation and even outperforms DCFE with extra $3$D information. 
Our approach performs slightly worse than
LAB that uses extra boundary information~\cite{Wu0YWC018}
and PDB~\cite{FengKA0W18}
that uses stronger data augmentation.

\vspace{.1cm}
\noindent\textbf{COFW.}
The COFW dataset \cite{Burgos-ArtizzuPD13} 
consists of $1,345$ training
and $507$ testing faces
with occlusions,
where each image has $29$ facial landmarks.

Table~\ref{table:comparison_cofw_testset} provides the comparison of our method with state-of-the-art methods. 
HRNetV$2$ outperforms other methods by a large margin.
In particular, it achieves the better performance than LAB with
extra boundary information and PDB with stronger data augmentation.

\vspace{.1cm}
\noindent\textbf{$300$W.}
The dataset~\cite{SagonasTZP13} is a combination
of HELEN~\cite{LeBLBH12}, LFPW ~\cite{BelhumeurJKK13}, AFW~\cite{ZhuR12}, XM2VTS~ and IBUG datasets,
where each face has $68$ landmarks. 
Following~\cite{RenCWS16}, we use the $3,148$ training images, which
contains the training subsets of HELEN and LFPW and the full set of AFW. 
We evaluate the performance
using two protocols, full set and test set. 
The full set contains $689$ images and is further divided
into a common subset ($554$ images) from HELEN and LFPW, and a challenging subset
($135$ images) from IBUG. 
The official test set, used for competition,  contains $600$ images ($300$ indoor and $300$ outdoor images).

Table \ref{table:comparison_300w_fullset} 
provides the results 
on the full set, and its two subsets:
common and challenging. 
Table \ref{table:comparison_300w_testset} provides the results
on the test set. 
In comparison to Chen et al. \cite{ChenSWLY17}
that uses Hourglass with large parameter and computation complexity
as the backbone,
our scores are better except the AUC$_{0.08}$ scores.
Our HRNetV$2$ gets the overall 
best performance among methods without extra information and stronger data augmentation, and is even better than LAB with extra boundary information and DCFE~\cite{ValleBVB18} that explores extra $3$D information.

\renewcommand{\arraystretch}{1.3}
\begin{table}[!htbp]
\scriptsize
\setlength{\tabcolsep}{13pt}
\centering
\caption{Facial landmark detection results 
(NME) on WFLW \texttt{test} and $6$ subsets: 
pose,
expression (expr.), 
illumination (illu.), 
make-up (mu.), 
occlusion (occu.)
and blur.
LAB~\cite{Wu0YWC018} is trained with extra boundary information (B).
PDB~\cite{FengKA0W18} adopts stronger data augmentation (DA).
Lower is better.}
\label{table:comparison_wflw_testset}
\begin{tabular}{l|l|r|r|r|r|r|r|r }
\hline \noalign{\smallskip}
   & backbone & test & pose & expr. & illu. & mu & occu. & blur\\
\hline

\hline
 ESR \cite{CaoWWS12}& - & $11.13$ & $25.88$ & $11.47$ & $10.49$ & $11.05$ & $13.75$ & $12.20$\\
SDM \cite{XiongT13}& - &$10.29$ & $24.10$ & $11.45$ & $9.32$ & $9.38$ & $13.03$ & $11.28$\\
CFSS \cite{ZhuLLT15}& - &$9.07$ & $21.36$ & $10.09$ & $8.30$ & $8.74$ & $11.76$ & $9.96$\\
DVLN \cite{WuY17} & VGG-16&$6.08$ & $11.54$ & $6.78$ & $5.73$ & $5.98$ & $7.33$ & $6.88$\\
\hline
Our approach & HRNetV$2$-W$18$ & $\mathbf{4.60}$ & $\mathbf{7.94}$ & $\mathbf{4.85}$ & $\mathbf{4.55}$ & $\mathbf{4.29}$ & $\mathbf{5.44}$ & $\mathbf{5.42}$\\
\hline

\hline
\multicolumn{3}{l}{
\emph {Model trained with \texttt{extra} info.}}\\
\hline
LAB (w/ B)~\cite{Wu0YWC018}& Hourglass & $5.27$ & $10.24$ & $5.51$ & $5.23$ & $5.15$ & $6.79$ & $6.32$\\
PDB (w/ DA)~\cite{FengKA0W18}& ResNet-$50$ & $5.11$ & $8.75$ & $5.36$ & $4.93$ & $5.41$ & $6.37$ & $5.81$\\
\hline
\end{tabular} 
\end{table}

\renewcommand{\arraystretch}{1.3}
\begin{table}[!htbp]
\setlength{\tabcolsep}{13pt}
\scriptsize
\centering
\caption{Facial landmark detection results (NME) on AFLW.
DCFE~\cite{ValleBVB18} uses extra $3$D information ($3$D). Lower is better.}
\label{table:comparison_aflw_testset}
\begin{tabular}{l|l|c|c }
\hline \noalign{\smallskip}
 & backbone &  full &  frontal \\
\hline

\hline
RCN \cite{HonariYVP16} &- & $5.60$ & $5.36$\\
CDM \cite{YuHZYM13} & -& $5.43$ & $3.77$\\
ERT \cite{KazemiS14}&- & $4.35$ & $2.75$\\
LBF \cite{RenCW014} & -& $4.25$ & $2.74$\\
SDM \cite{XiongT13} &- & $4.05$ & $2.94$\\
CFSS \cite{ZhuLLT15}&- & $3.92$ & $2.68$ \\
RCPR \cite{Burgos-ArtizzuPD13}&- &$3.73$ & $2.87$\\
CCL \cite{ZhuLLT16}&- &$2.72$ & $2.17$\\
DAC-CSR \cite{FengKC0W17}& &$2.27$ & $1.81$\\
TSR \cite{LvSXCZ17}&VGG-S &$2.17$ & - \\
CPM + SBR \cite{DongYWW0S18}& CPM &$2.14$&-\\
SAN \cite{DongYO018}& ResNet-$152$ &$1.91$ & $1.85$\\
DSRN \cite{MiaoZLDAH18}& - &$1.86$ & -\\
LAB (w/o B) \cite{Wu0YWC018}& Hourglass & $1.85$ & $1.62$\\
\hline
Our approach & HRNetV2-W$18$ & $\mathbf{1.57}$ & $\mathbf{1.46}$ \\
\hline

\hline
\multicolumn{3}{l}{
\emph {Model trained with \texttt{extra} info.}}\\
\hline
DCFE (w/ $3$D)~\cite{ValleBVB18}& - &$2.17$ & - \\
PDB (w/ DA)~\cite{FengKA0W18}& ResNet-$50$&$1.47$ & -\\
LAB (w/ B)~\cite{Wu0YWC018}& Hourglass &$1.25$ & $1.14$\\
\hline
\end{tabular} 
\end{table}

\renewcommand{\arraystretch}{1.3}
\begin{table}[!htbp]
\setlength{\tabcolsep}{13pt}
\scriptsize
\centering
\caption{Facial landmark detection results on COFW \texttt{test}. 
The failure rate is calculated at the threshold $0.1$. 
Lower is better for NME and FR$_{0.1}$.
}
\begin{tabular}{l|l|cc }
\hline\noalign{\smallskip}
 & backbone & NME & FR$_{0.1}$\\
\hline

\hline
Human & - & $5.60$ & -\\
ESR \cite{CaoWWS12} &-&$11.20$&$36.00$\\
RCPR \cite{Burgos-ArtizzuPD13}& - &$8.50$ & $20.00$\\
HPM \cite{GhiasiF14}& - &$7.50$ & $13.00$ \\
CCR \cite{FengHKCW15}& -& $7.03$ & $10.90$ \\
DRDA \cite{ZhangKSC16}& - &$6.46$ & $6.00$ \\
RAR \cite{XiaoFXLYK16}& - &$6.03$ & $4.14$ \\
DAC-CSR \cite{FengKC0W17}& - &$6.03$ & $4.73$ \\
LAB (w/o B) \cite{Wu0YWC018}& Hourglass & $5.58$ & $2.76$\\
\hline
Our approach & HRNetV$2$-W$18$ & $\mathbf{3.45}$ & $\mathbf{0.19}$ \\
\hline

\hline
\multicolumn{3}{l}{
\emph {Model trained with \texttt{extra} info.}}\\
\hline
PDB (w/ DA)~\cite{FengKA0W18}& ResNet-$50$ & $5.07$ & $3.16$\\
LAB (w/ B)~\cite{Wu0YWC018}& Hourglass &$3.92$ & $0.39$ \\
\hline
\end{tabular}
\label{table:comparison_cofw_testset}
\end{table}

\renewcommand{\arraystretch}{1.3}
\begin{table}[t]
\setlength{\tabcolsep}{13pt}
\centering
\scriptsize
\caption{Facial landmark detection results (NME) 
on $300$W:
common, challenging and full.
Lower is better.}

\label{table:comparison_300w_fullset}
\begin{tabular}{ l|l|ccc  }
\hline\noalign{\smallskip}
  & backbone &common & challenging & full \\
\hline

\hline
RCN \cite{HonariYVP16} &-& $4.67$ & $8.44$ & $5.41$\\
DSRN \cite{MiaoZLDAH18} &-& $4.12$ & $9.68$ & $5.21$ \\
PCD-CNN \cite{KumarC18} &-& $3.67$ & $7.62$ & $4.44$ \\
CPM + SBR \cite{DongYWW0S18} & CPM & $3.28$ & $7.58$ & $4.10$ \\
SAN \cite{DongYO018} &ResNet-152& $3.34$ & $6.60$ & $3.98$ \\
DAN \cite{KowalskiNT17} &-& $3.19$ & $5.24$ & $3.59$ \\
\hline
Our approach & HRNetV$2$-W$18$ &$\mathbf{2.87}$ & $\mathbf{5.15}$ & $\mathbf{3.32}$\\
\hline

\hline
\multicolumn{3}{l}{
\emph {Model trained with \texttt{extra} info.}}\\
\hline
LAB (w/ B) \cite{Wu0YWC018}& Hourglass & $2.98$ & $5.19$ & $3.49$ \\
DCFE (w/ $3$D) \cite{ValleBVB18}& - &$2.76$ & $5.22$ & $3.24$ \\
\hline
\end{tabular} 
\end{table}

\renewcommand{\arraystretch}{1.3}
\begin{table}[t]
\setlength{\tabcolsep}{13pt}
\scriptsize
\centering
\caption{Facial landmark detection results on $300$W \texttt{test}.
DCFE~\cite{ValleBVB18} uses extra 3D information (3D).
LAB~\cite{Wu0YWC018} is trained with extra boundary information (B).
Lower is better for NME, FR$_{0.08}$ and FR$_{0.1}$, 
and higher is better for AUC$_{0.08}$ and AUC$_{0.1}$.}
\label{table:comparison_300w_testset}
\begin{tabular}{l|l|c|c|c|c|c}
\hline\noalign{\smallskip}
 & backbone &NME & AUC$_{0.08}$ & AUC$_{0.1}$ & FR$_{0.08}$ & FR$_{0.1}$\\
\hline

\hline
Balt. et al. \cite{Baltrusaitis0M13} & - & - &$19.55$ &-& $38.83$ & -\\
ESR \cite{CaoWWS12}& - &$8.47$ & $26.09$ & - & $30.50$ & - \\
ERT \cite{KazemiS14}& - &$8.41$ & $27.01$ & - & $28.83$ & - \\
LBF \cite{RenCW014}& - &$8.57$ & $25.27$ & - & $33.67$ & - \\
Face++ \cite{ZhouFCJY13} & - & - &$32.81$ & - & $13.00$ & - \\
SDM \cite{XiongT13}& - &$5.83$ & $36.27$ & - & $13.00$ & - \\
CFAN \cite{ZhangSKC14}& - &$5.78$ & $34.78$ & - & $14.00$ & - \\
Yan et al. \cite{YanLYL13} & -&- &$34.97$&-&$12.67$&-\\
CFSS \cite{ZhuLLT15}& - &$5.74$ & $36.58$ & - & $12.33$ & - \\
MDM \cite{TrigeorgisSNAZ16}& - &$4.78$ & $45.32$ & - & $6.80$ & - \\
DAN \cite{KowalskiNT17}& - &$4.30$ & $47.00$ & - & $2.67$ & - \\
Chen et al. \cite{ChenSWLY17}& Hourglass &$3.96$ & $\mathbf{53.64}$ & - & $2.50$ & - \\
Deng et al. \cite{DengLYT16}& - & - & - & $47.52$ & - & $5.50$ \\
Fan et al. \cite{FanZ16}& - &- & - & $48.02$ & - & $14.83$ \\
DReg + MDM \cite{GulerTASZK17}& ResNet101 &- & - & $52.19$ & - & $3.67$ \\
JMFA \cite{DengTZZ17}&Hourglass &- & - & $54.85$ & - & $1.00$ \\
\hline
Our approach & HRNetV$2$-W$18$& $\mathbf{3.85}$  & 52.09  & $\mathbf{61.55}$ & $\mathbf{1.00}$ & $\mathbf{0.33}$  \\
\hline

\hline
\multicolumn{3}{l}{
\emph {Model trained with \texttt{extra} info.}}\\
\hline
LAB (w/ B) \cite{Wu0YWC018}& Hourglass &- & -& $58.85$ & - & $0.83$ \\
DCFE (w/ $3$D) \cite{ValleBVB18}& - &$3.88$ & $52.42$ & - & $1.83$ & - \\
\hline
\end{tabular} 
\end{table}

\clearpage
\onecolumn
\section{More object detection and instance
results on COCO \texttt{val2017}}
\begin{center}
\small
\renewcommand{\arraystretch}{1.2}
\setlength{\LTcapwidth}{\textwidth}
\addtolength{\tabcolsep}{3.1pt}
\begin{longtable}{l|c|cccccc|cccccc}
	\caption[c]{More object detection and 
	instance segmentation results on COCO \texttt{val}.
	$\operatorname{AP}^{b}$ and $\operatorname{AP}^{m}$ denote box mAP and mask mAP respectively. Most results are taken from~\cite{mmdetection}
	except that the results using HRNet
	are obtained by running the code at~\url{https://github.com/open-mmlab/mmdetection}.}
	\label{tab:detailed-results}\\
	\hline
	 Backbone & LS & $\operatorname{AP}^{b}$ & $\operatorname{AP}^{b}_{50}$ & $\operatorname{AP}^{b}_{75}$ & $\operatorname{AP}^{b}_{S}$ & $\operatorname{AP}^{b}_{M}$ & $\operatorname{AP}^{b}_{L}$ & $\operatorname{AP}^{m}$ & $\operatorname{AP}^{m}_{50}$ & $\operatorname{AP}^{m}_{75}$ & $\operatorname{AP}^{m}_{S}$ & $\operatorname{AP}^{m}_{M}$ & $\operatorname{AP}^{m}_{L}$ \\
	\hline
	\endfirsthead
	\multicolumn{14}{c}%
	{\tablename\ \thetable\ -- \textit{Continued from previous page}} \\
	\hline
	Backbone & Lr Schd & $\text{AP}^{b}$ & $\text{AP}^{b}_{50}$ & $\text{AP}^{b}_{75}$ & $\text{AP}^{b}_{S}$ & $\text{AP}^{b}_{M}$ & $\text{AP}^{b}_{L}$ & $\text{AP}^{m}$ & $\text{AP}^{m}_{50}$ & $\text{AP}^{m}_{75}$ & $\text{AP}^{m}_{S}$ & $\text{AP}^{m}_{M}$ & $\text{AP}^{m}_{L}$ \\
	\hline
	\endhead
	\hline \multicolumn{14}{r}{\textit{Continued on next page}} \\
	\endfoot
	\hline
	\endlastfoot

	\multicolumn{14}{c}{FCOS}\\
	\hline
	R-50 (c)    & 1x & 36.7 & 55.8 & 39.2 & 21.0 & 40.7 & 48.4 & -    & -    & -    & -    & -    & -   \\
	R-101 (c)   & 1x & 39.1 & 58.5 & 41.8 & 22.0 & 43.5 & 51.1 & -    & -    & -    & -    & -    & -   \\
	R-50 (c)    & 2x & 36.9 & 55.8 & 39.1 & 20.4 & 40.1 & 49.2 & -    & -    & -    & -    & -    & -    \\
	R-101 (c)   & 2x & 39.1 & 58.6 & 41.7 & 22.1 & 42.4 & 52.5 & -    & -    & -    & -    & -    & -    \\
    {HRNetV2-W18} & 1x & 35.2 & 52.9 & 37.3 & 20.4 & 37.8 & 46.1 & -    & -    & -    & -    & -    & -   \\
    {HRNetV2-W32} & 1x & 38.2 & 56.2 & 40.9 & 22.2 & 41.8 & 50.0 & -    & -    & -    & -    & -    & -   \\
    {HRNetV2-W18} & 2x & 37.7 & 55.9 & 40.1 & 22.0 & 40.8 & 48.5 & -    & -    & -    & -    & -    & -   \\
    {HRNetV2-W32} & 2x & 40.3 & 58.7 & 43.3 & 23.6 & 43.4 & 52.9 & -    & -    & -    & -    & -    & -   \\
	\hline
	\multicolumn{14}{c}{FCOS (mstrain)}\\
	\hline
	R-50 (c)    & 2x & 38.7 & 58.0 & 41.4 & 23.4 & 42.8 & 49.0 & -    & -    & -    & -    & -    & -   \\
	R-101 (c)   & 2x & 40.8 & 60.1 & 43.8 & 24.5 & 44.5 & 52.8 & -    & -    & -    & -    & -    & -   \\
	{HRNetV2-W18} & 2x & 38.1 & 56.3 & 40.6 & 22.9 & 41.1 & 48.6 & -    & -    & -    & -    & -    & -   \\
	{HRNetV2-W32} & 2x & 41.4 & 60.3 & 44.2 & 25.2 & 44.8 & 52.3 & -    & -    & -    & -    & -    & -   \\
	{HRNetV2-W48} & 2x & 42.9 & 61.9 & 45.9 & 26.4 & 46.7 & 54.6 & -    & -   & -    & -    & -    & -   \\
	X-101-64x4d & 2x & 42.8 & 62.6 & 45.7 & 26.5 & 46.9 & 54.5 & -    & -    & -    & -    & -    & -   \\
	\hline
	\multicolumn{14}{c}{Mask R-CNN} \\
	\hline
	R-50 (c)    & 1x & 37.4 & 58.9 & 40.4 & 21.7 & 41.0 & 49.1 & 34.3 & 55.8 & 36.4 & 18.0 & 37.6 & 47.3 \\
	R-101 (c)   & 1x & 39.9 & 61.5 & 43.6 & 23.9 & 44.0 & 51.8 & 36.1 & 57.9 & 38.7 & 19.8 & 39.8 & 49.5 \\
	R-50        & 1x & 37.3 & 59.0 & 40.2 & 21.9 & 40.9 & 48.1 & 34.2 & 55.9 & 36.2 & 18.2 & 37.5 & 46.3 \\
	R-101       & 1x & 39.4 & 60.9 & 43.3 & 23.0 & 43.7 & 51.4 & 35.9 & 57.7 & 38.4 & 19.2 & 39.7 & 49.7 \\
	{HRNetV2-W18} & 1x & 37.3 & 58.2 & 40.7 & 22.1 & 40.2 & 47.6 & 34.2 & 55.0 & 36.2 & 18.4 & 36.7 & 46.0 \\
    {HRNetV2-W32} & 1x & 40.7 & 61.9 & 44.6 & 25.1 & 44.4 & 51.8 & 36.8 & 58.7 & 39.5 & 20.9 & 40.0 & 49.3 \\
	X-101-32x4d & 1x & 41.1 & 62.8 & 45.0 & 24.0 & 45.4 & 52.6 & 37.1 & 59.4 & 39.8 & 19.7 & 41.1 & 50.1 \\
	X-101-64x4d & 1x & 42.1 & 63.8 & 46.3 & 24.4 & 46.6 & 55.3 & 38.0 & 60.6 & 40.9 & 20.2 & 42.1 & 52.4 \\
	R-50        & 2x & 38.5 & 59.9 & 41.8 & 22.6 & 42.0 & 50.5 & 35.1 & 56.8 & 37.0 & 18.9 & 38.0 & 48.3 \\
	R-101       & 2x & 40.3 & 61.5 & 44.1 & 22.2 & 44.8 & 52.9 & 36.5 & 58.1 & 39.1 & 18.4 & 40.2 & 50.4 \\
	{HRNetV2-W18} & 2x & 39.2 & 60.1 & 42.9 & 24.2 & 42.1 & 50.8 & 35.7 & 57.3 & 38.1 & 17.6 & 37.8 & 52.3 \\
    {HRNetV2-W32} & 2x & 42.3 & 62.7 & 46.1 & 26.1 & 45.5 & 54.7 & 37.6 & 59.7 & 40.3 & 21.4 & 40.5 & 51.2 \\
	X-101-32x4d & 2x & 41.4 & 62.5 & 45.4 & 24.0 & 45.4 & 54.5 & 37.1 & 59.4 & 39.5 & 19.9 & 40.6 & 51.3 \\
	X-101-64x4d & 2x & 42.0 & 63.1 & 46.1 & 23.9 & 45.8 & 55.6 & 37.7 & 59.9 & 40.4 & 19.6 & 41.3 & 52.5 \\
	\hline

	\multicolumn{14}{c}{Cascade Mask R-CNN} \\
	\hline
	R-50        & 1x  & 41.2 & 59.1 & 45.1 & 23.3 & 44.5 & 54.5 & 35.7 & 56.3 & 38.6 & 18.5 & 38.6 & 49.2 \\
	R-101       & 1x  & 42.6 & 60.7 & 46.7 & 23.8 & 46.4 & 56.9 & 37.0 & 58.0 & 39.9 & 19.1 & 40.5 & 51.4 \\
	X-101-32x4d & 1x  & 44.4 & 62.6 & 48.6 & 25.4 & 48.1 & 58.7 & 38.2 & 59.6 & 41.2 & 20.3 & 41.9 & 52.4 \\
	X-101-64x4d & 1x  & 45.4 & 63.7 & 49.7 & 25.8 & 49.2 & 60.6 & 39.1 & 61.0 & 42.1 & 20.5 & 42.6 & 54.1 \\
	R-50        & 20e & 42.3 & 60.5 & 46.0 & 23.7 & 45.7 & 56.4 & 36.6 & 57.6 & 39.5 & 19.0 & 39.4 & 50.7 \\
	R-101       & 20e & 43.3 & 61.3 & 47.0 & 24.4 & 46.9 & 58.0 & 37.6 & 58.5 & 40.6 & 19.7 & 40.8 & 52.4 \\
    {HRNetV2-W18} & 20e & 41.9 & 59.6 & 45.7 & 23.8 & 44.9 & 55.0 & 36.4 & 56.8 & 39.3 & 17.0 & 38.6 & 52.9 \\
    {HRNetV2-W32} & 20e & 44.5 & 62.3 & 48.6 & 26.1 & 47.9 & 58.5 & 38.5 & 59.6 & 41.9 & 18.9 & 41.1 & 56.1 \\
    {HRNetV2-W48} & 20e & 46.0 & 63.7 & 50.3 & 27.5 & 48.9 & 60.1 & 39.5 & 61.1 & 42.8 & 19.7 & 41.8 & 56.9 \\
	X-101-32x4d & 20e & 44.7 & 63.0 & 48.9 & 25.9 & 48.7 & 58.9 & 38.6 & 60.2 & 41.7 & 20.9 & 42.1 & 52.7 \\
	X-101-64x4d & 20e & 45.7 & 64.1 & 50.0 & 26.2 & 49.6 & 60.0 & 39.4 & 61.3 & 42.9 & 20.8 & 42.7 & 54.1 \\
	\hline
	\multicolumn{14}{c}{Hybrid Task Cascade} \\
	\hline
	R-50        & 1x  & 42.1 & 60.8 & 45.9 & 23.9 & 45.5 & 56.2 & 37.3 & 58.2 & 40.2 & 19.5 & 40.6 & 51.7 \\
	R-50        & 20e & 43.2 & 62.1 & 46.8 & 24.9 & 46.4 & 57.8 & 38.1 & 59.4 & 41.0 & 20.3 & 41.1 & 52.8 \\
	R-101       & 20e & 44.9 & 63.8 & 48.7 & 26.4 & 48.3 & 59.9 & 39.4 & 60.9 & 42.4 & 21.4 & 42.4 & 54.4 \\
	{HRNetV2-W18} & 20e & 43.1 & 61.5 & 46.8 & 26.6 & 46.0 & 56.9 & 37.9 & 59.0 & 40.6 & 18.8 & 39.9 & 55.2 \\
    {HRNetV2-W32} & 20e & 45.3 & 63.6 & 49.1 & 27.0 & 48.4 & 59.5 & 39.6 & 61.2 & 43.0 & 19.1 & 42.0 & 57.9 \\
    {HRNetV2-W48} & 20e & 46.8 & 65.3 & 51.1 & 28.0 & 50.2 & 61.7 & 40.7 & 62.6 & 44.2 & 19.7 & 43.4 & 59.3 \\
    {HRNetV2-W48} & {28e} & 47.0 & 65.5 & 51.0 & 28.8 & 50.3 & 62.2 & 41.0 & 63.0 & 44.7 & 20.8 & 43.9 & 59.9 \\
	X-101-32x4d & 20e & 46.1 & 65.1 & 50.2 & 27.5 & 49.8 & 61.2 & 40.3 & 62.2 & 43.5 & 22.3 & 43.7 & 55.5 \\
	X-101-64x4d & 20e & 46.9 & 66.0 & 51.2 & 28.0 & 50.7 & 62.1 & 40.8 & 63.3 & 44.1 & 22.7 & 44.2 & 56.3 \\
	{X-101-64x4d} & {28e}  & 46.8 & 65.6 & 50.9 & 27.5 & 51.0 & 61.7 & 40.7 & 63.1 & 43.9 & 20.0 & 44.1 & 59.9 \\
\end{longtable}
\end{center}

\end{document}